\documentclass[a4paper,twocolumn]{article} 

\ifx\pdfoutput\undefined
   \usepackage[dvips]{graphicx}
   \DeclareGraphicsExtensions{.eps} 
 \else
   \usepackage{graphicx}
   \DeclareGraphicsExtensions{.pdf,.jpg,.png,.mps} 
\fi
\graphicspath{{./../Figures/},{./../Proposal_PhD_Chang_FFPD/Figures/}, {./../sn_version/}}

\usepackage{amsmath,amssymb}   
\usepackage[ansinew]{inputenc} 
\usepackage[english]{babel}    
\usepackage[round,authoryear]{natbib}  
\usepackage[font=footnotesize,labelfont=bf]{caption}
\usepackage{algorithm}
\usepackage{algpseudocode}
\usepackage{titlesec}

\titleformat*{\section}{\Large\bfseries}
\titleformat*{\subsection}{\large\bfseries}
\titleformat*{\subsubsection}{\normalsize\bfseries}
\titleformat*{\paragraph}{\normalsize\bfseries}

\usepackage{hyperref}
\usepackage{tabularx}
\newcolumntype{L}[1]{>{\raggedright\let\newline\\\arraybackslash\hspace{0pt}}p{#1}}
\newcolumntype{C}[1]{>{\centering\let\newline\\\arraybackslash\hspace{0pt}}p{#1}}
\newcolumntype{R}[1]{>{\raggedleft\let\newline\\\arraybackslash\hspace{0pt}}b{#1}}
\usepackage{multirow}
\newcolumntype{P}[1]{>{\raggedright\arraybackslash}m{#1}}
\usepackage[a4paper, total={6.5in, 10in}]{geometry}

\usepackage{cuted}

\usepackage{xstring} 

\providecommand{\sci}[1]{\protect\ensuremath{\times 10^{\StrSubstitute[0]{#1}{e}{}}}}

\usepackage[colorinlistoftodos,prependcaption,textsize=tiny]{todonotes}
\usepackage{cleveref}

\title{{\LARGE \textbf{Skin feature tracking using deep feature encodings}}\\
\vspace*{0.3cm}
{\large Jose Ramon Chang, MSc \textsuperscript{1} and Torbj{\"o}rn E. M. Nordling, MSc, PhD \textsuperscript{1,2}\footnote{Corresponding author}}\\
\vspace*{0.2cm}
{\footnotesize \begin{tabular}{C{6cm} C{0.1cm} C{7cm}}
\textsuperscript{1}Department of Mechanical Engineering& &\textsuperscript{2}Department of Applied Physics and Electronics\\
National Cheng Kung University& &Ume\aa{} University\\
Tainan 701, Taiwan & &90187 Ume\aa, Sweden \\
\end{tabular}}\\
\vspace*{10pt}
{\footnotesize \texttt{\{jose.chang, torbj{\"o}rn.nordling\}@nordlinglab.org}}
}
\date{\vspace*{-1.6cm} } 

\renewenvironment{abstract}
{
  \centerline{\large \bfseries \scshape Abstract}
  \vspace*{0.2cm} \footnotesize 
}
{

}
\begin{document}               
\maketitle
\abstract{\itshape
Facial feature tracking is a key component of imaging ballistocardiography (BCG) where accurate quantification of the displacement of facial keypoints is needed for good heart rate estimation. 
Skin feature tracking enables video-based quantification of motor degradation in Parkinson's disease.
Traditional computer vision algorithms include Scale Invariant Feature Transform (SIFT), Speeded-Up Robust Features (SURF), and Lucas-Kanade method (LK). 
These have long represented the state-of-the-art in efficiency and accuracy but fail when common deformations, like affine local transformations or illumination changes, are present.

Over the past five years, deep convolutional neural networks have outperformed traditional methods for most computer vision tasks. 
We propose a pipeline for feature tracking, that applies a convolutional stacked autoencoder to identify the most similar crop in an image to a reference crop containing the feature of interest.
The autoencoder learns to represent image crops into deep feature encodings specific to the object category it is trained on.

We train the autoencoder on facial images and validate its ability to track skin features in general using manually labeled face and hand videos.
The tracking errors of distinctive skin features (moles) are so small that we cannot exclude that they stem from the manual labelling based on a $\chi^2$-test. 
With a mean error of 0.6-4.2 pixels, our method outperformed the other methods in all but one scenario.
More importantly, our method was the only one to not diverge.

We conclude that our method creates better feature descriptors for feature tracking, feature matching, and image registration than the traditional algorithms.
}


\section{Introduction}\label{sec:introduction}
\label{sec:intro}

Facial feature points are distinctive points typically located around the eyes, nose, chin, and mouth that carry the most relevant information for both discriminative and generative purposes \citep{Wang2018}.
With increasingly larger and broadly available facial imagery datasets, researchers have used deep learning to teach computers to perform interesting tasks, such as using these feature points to determine the level of drowsiness/sleepiness of a driver \citep{garcia2012vision}, recognize emotions \citep{Zhao2016}, quantify fatigue \citep{Uchida2018}, estimate heart rate \citep{Hassan2017}, and lip-reading \citep{chung2017lip}. 

Feature tracking can be achieved through iterative feature matching across sequential frames in a video.
Current feature matching and tracking methods are not robust enough, failing sometimes even from small photometric/geometric changes, motion deformations, or occlusions present in the data as we demonstrate.
We define feature matching as the task of, based on information about the feature in one image, finding the same feature in another deformed image.
A feature, in this case, is any characteristic that is a part of the main entity/object portrayed in the image.
While an object constitutes a whole on its own with a boundary  separating it from other objects and distinguishing it from its environment, $e.g.$ a car on the street or a bear in a forest, a feature is a part of an object and does not have a boundary.
A feature is defined by a local pattern of some sort, $e.g.$ a wrinkle on skin, and it is typically hard to even say where it ends.
This make it significantly harder to define, detect, match, and track features compared to objects.
Nonetheless, feature matching is sometimes treated as an object detection/recognition task.
In general, the real world performance of tracking methods decreases with the number of assumptions made in the method.
These assumptions include but are not limited to; small motion, specific types of geometric deformations, constant lighting, corner points, and no occlusion of points.
Tracking of the position of a skin region of interest is particularly challenging.
In an image, skin typically appears homogenous with only some irregularly spaced locally distinct features. 
The pattern of these locally distinct features may be heavily deformed due to change in facial expression or lightning.
Some skin features are though unique and easily recognisable by the human eye, $e.g.$ a mole or birth mark.
To track these feature points, here we follow the established engineering practice of applying the simplest method proven to work in other applications. 
Therefore, we use deep learning, more precisely an autoencoder, to track crops containing these distinct skin feature points, aiming  to push the limits of current computer vision methods by creating a more robust, and accurate pipeline to track skin feature points.

While SIFT \citep{Lowe2004} and SURF \citep{Bay2006} have been invented almost 20 years ago, they are still considered by many to be state-of-the-art feature matching techniques.
One research group found SIFT to be the second-best descriptor in terms of feature matching and recognition accuracy of the same object or scene \citep{Mikolajczyk2005}.
Gradient Location and Orientation Histogram (GLOH), their proposed extension of SIFT, marginally outperformed SIFT for most tasks. 
This shows the robustness and distinctiveness of SIFT descriptors.
A review found that SIFT and SURF converged similarly to the lowest errors for monocular visual-odometry tasks \citep{chien2016use}.
Another group concluded that SIFT is still the most accurate feature descriptor for image recognition and feature matching applications \citep{khan2015better}.
They also found that general-purpose feature descriptors perform fairly well for feature matching but are not good for image recognition.
Some researchers compared SIFT descriptors with descriptors obtained using a pre-trained Convolutional Neural Network (CNN) on the Imagenet dataset \citep{zheng2017sift}.
They found that while CNN-based methods have advantages in nearly all their benchmarking datasets, SIFT is still better in the cases where images are gray-scale, there are intense color changes, and when there are severe occlusions.
One review concluded that SIFT is the most accurate feature-detector-descriptor for overall scale, rotations, and affine deformations \citep{Tareen2018}.
We think this is due to the CNN not having these types of instances in its training distribution. 
In the past 2 years, SIFT and SURF have been used in many applications, like underwater image recognition for autonomous underwater vehicles \citep{ansari2019review}, image forgery detection \citep{alberry2018fast, chen2019rotational}, and mosaicking of unmanned aerial vehicle images for crop growth monitoring \citep{zhao2019rapid}.
In 2017, SIFT was even used for currency recognition \citep{doush2017currency}.

By far, the most commonly used method for sparse optical flow is LK \citep{lucas1981iterative}.
This method is mathematically well defined and has clear assumptions for proper functioning like no photometric deformations, small displacement (or the images are separated by a very small time increment), and the point is a corner.
Since this method is computationally cheap, it can be applied in many real-time feature tracking applications.
Within the last 5 years applications of LK include image alignment \citep{chang2017clkn}, image registration \citep{douini2017image}, feature tracking \citep{ahmine2019adaptive, douini2017solving}, visual odometry \citep{wong2017uncertainty}, and motion breath recognition \citep{tran2018pyramidal}.
In 2015, this method was used for facial expression recognition \citep{pu2015facial}.

\subsection{Deep learning for skin feature tracking}
Deep learning has been extensively used in medical applications involving skin. 
Recent reviews include deep learning for remote photoplethysmography (rPPG) \citep{ni2021review, cheng2021deep}, gaze estimation \citep{cheng2021appearance}, speech recognition \citep{lee2021biosignal}, skin lesion segmentation \citep{stofa2021skin}, and driver fatigue detection \citep{sikander2018driver}.
Although all of these implicitly require skin feature tracking, there has been little original research done on quantifying the matching error of skin features explicitly.
The most similar research that we could find to ours is of a group that tried to match keypoints on the back of patients using hyperspectral images \citep{manni2020hyperspectral}.
Their hyperspectral camera, running at 16 FPS, acquired 41 equally distributed spectral bands in the range of 450-950 nm in one snapshot of size $2048 \times 1080$.
A color image is used as a reference to register all the hyperspectral data, ensuring proper data fusion.
This yields a tensor of size $1080 \times 2048 \times 41$ to which they then later applied the Saliency-Band-Based Selection algorithm (SSBS) \citep{su2018saliency} to reduce its dimensions.
They aim to find the correspondences of points on the back of 17 patients with images taken when they breathe in and out.
They applied SURF and DEep Local Feature (DELF) to create feature descriptors of keypoints in the images.
Their results indicate that DELF had a localization error of 0.25 mm and that it outperformed the SURF descriptors with respect to the ground truth based on optical markers.
As explained in detail later, DELF implicitly select features and cannot be used to track a selected feature like in our case.

\subsection{Autoencoders for feature matching}
The most popular learning paradigm in computer vision, and all fields in general, is supervised learning \cite{chum2019beyond, khan2021machine}.
However, a common bottleneck factor when training a supervised learning model is the amount of good-quality sample-label pairs.
The same complexity that allows these models to learn very complex functions is also responsible for them requiring vast amounts of data to tune this extensive amount of parameters.

An autoencoder is a type of neural network that aims to learn the identity function.
As such, it can be trained by minimizing the discrepancy or distance between the original data and its reconstruction.
This means that it only requires samples for its training and removes the need for labels.
The identity function may seem a trivial, meaningless function to learn.
However, if we constrain the latent space to have fewer dimensions than our input, then we force the network to learn the most salient, distinctive features of our data.
Their use for dimensionality reduction was popularized in 2006 \citep{Hinton2006}.
They described the algorithm as a nonlinear generalization of Principal Component Analysis (PCA).
Similarly to PCA, a key feature of autoencoders is that they are able to encode the input data into latent state representations that provide interesting insight into the samples of the data.
It is described as a nonlinear generalization of PCA because a neural network typically passes the output of each neuron through a nonlinear function, $e.g.$ sigmoid, tanh, Rectified Linear Units (ReLU), or all of the ReLU variants.

Autoencoders have been used for feature matching in the past.
A siamese network was used by researchers to match the latent space representation of tops and bottoms from the FashionVC and MbFashion \citep{gao2019fashion}.
They used the outfits the fashionistas published on different social networks as their ground truths.
Another team used a convolutional autoencoder to match image patches between two infrared images for 3D reconstruction \citep{knyaz2017deep}.
Their results indicate that their autoencoder was better for matching the features of the test part of their Multi-View Stereo InfraRed (MVSIR) imagery dataset compared to SIFT \citep{Lowe2004}, deep convolutional feature point descriptors \citep{simo2015discriminative}, and stereo matching convolutional neural network \citep{zbontar2016stereo}.
They mentioned that if the convolutional autoencoder is able to obtain good restoration quality then it has learned to extract the most salient information from the data.
Thus the hidden representations can be used for sparse image matching.
Another group used a convolutional autoencoder to learn features from images of finger veins and then used a support vector machine (SVM) to classify them \citep{hou2019convolutional}.
They credited the convolutional autoencoder for effective learning of finger vein features, preserving the main information of the image, reducing redundant information, and improving the recognition efficiency with the help of their SVM classifier.
We train our autoencoder to match skin features in general.

\subsection{Deep learning for feature tracking}
\subsubsection{Object detection based tracking}
Most deep learning feature tracking research is built upon the simple task of object detection.
Object detection is one of the most elementary problems in computer vision where one seeks to locate object instances from a set of predefined categories in images \citep{liu2020deep}.
An object detection network takes the image as input and outputs a set of bounding boxes.
Each of these bounding boxes is typically defined by its $x$ and $y$ coordinates, box width and height, confidence that there is an object inside, and the probability of that object being from each one of the predefined classes.

Many deep learning approaches for Multiple Object Tracking (MOT), also known as Multi-Target Tracking, have been published in the last four years \citep{Ciaparrone2020}.
Similar to object detection, MOT assigns bounding boxes to the detected objects but in addition, each bounding box is associated with a target ID.
This target ID serves to differentiate among intra-class objects.

The standard approach of most MOT algorithms is referred to as ``tracking-by-detection", $i.e.$ the tracking is done by iteratively detecting objects successively across the frames of a video.
The bounding boxes of two successive frames are usually associated together to assign the same target ID to the bounding boxes that contain the same object. 
Recently, many methods that have been developed can detect objects with very good accuracy and precision, $e.g.$ Faster-Region based Convolutional Neural Networks (RCNN) \citep{Ren2015}, Single Shot Detector \citep{Liu2016ssd}, You Only Look Once (YOLO) \citep{Redmon2016}, YOLO v2 \citep{Redmon2017}, YOLO v3 \citep{Redmon2018}, YOLO v4 \citep{Bochkovskiy2020}, Mask RCNN \citep{He2017}, and Region-based Fully Connected Networks (R-FCN) \citep{Dai2016}.
Since these methods can already provide very good detections, most of these algorithms focus on developing better associating algorithms to track objects with the same target ID.
These algorithms are developed for large objects, $e.g.$ a pedestrian, while we aim to track small objects with low resolutions, $e.i.$ skin features.
This makes our problem inherently more difficult as we have less local information and less distinctive features (patterns/ structures) to identify our target in the images; leading to a possible increase in false positive cases as there could be many parts of the image with a similar appearance.

The most common deep learning algorithms for detection are variants of Faster-RCNN \citep{Ren2015} or Single-Shot Detection (SSD) \citep{Liu2016ssd}.
While these algorithms have shown to be very accurate at tracking objects, most of these deep learning methods require large amounts of labeled data to train the detector.
Labeling data is usually strenuous labor, as accurate training labels require the use of a consensus label of many labeling attempts.
This data restraint is a hallmark of supervised learning.
The parameters of the model are updated based on the gradients of the error between the prediction and ground truth.
In addition, they are still slow and do not meet real-time requirements \citep{Bi2019}.
Here we use an autoencoder to avoid the labelling issue.

Centernet \citep{zhou2019objects} is another object detection algorithm that represents objects as the center of their bounding box.
Object size, dimension, 3D extent, orientation, and pose are then regressed directly from image features at these center locations.
Their best model, which is based on Hourglass-104 \citep{law2018detecting}, is trained to produce heatmaps where peaks correspond to object centers. 
It achieves 42.2\% Average Precision (AP) in 7.8 FPS in the Microsoft (MS) Common Objects in COntext (COCO) \citep{lin2014microsoft}.
While this is comparable to other state-of-the-art object detectors like FasterRCNN \citep{Ren2015}, YoloV3 \citep{Redmon2018}, and RetinaNet \citep{lin2017focal}, it differs significantly from our applications as its more similar to object detection than feature matching.
An object recognition algorithm aims to identify the main object in the image.
It does so by predicting a probability distribution of all the object categories it has been trained to recognize.
An object detection algorithm aims to, in addition to recognizing which are the objects in the image, locate all the main objects by estimating their bounding box locations and dimensions.

\subsubsection{Other deep learning based tracking}
Deepflow \citep{weinzaepfel2013deepflow} is a variational approach that uses a CNN to calculate the dense optical flow between two images.
Variational approaches had been widely used ever since their invention in 1981 \citep{horn1981determining}.
They use a 6-layered convolutional architecture to create descriptors between images and retrieve quasi-dense correspondences based on these feature descriptors.
Their claim for being able to deal with large motions comes from their performance on the Middlebury \citep{baker2011database} and MPI-Sintel \citep{butler2012naturalistic} datasets.
These datasets are well-known in the optical flow community.
However, the Middlebury dataset has less than 3\% of its pixels with displacements over 20 pixels and none have displacements over 25 pixels.
The MPI-Sintel dataset contains approximately 10\% of its pixels with a displacement of over 40 pixels.
These displacements are still small for our applications.
In addition, while they use convolutional layers and max pooling, there is no learning done by the network and all of its kernel parameters are predetermined.

FlowNet \citep{dosovitskiy2015flownet} was one of the first attempts to train a deep CNN to estimate optical flow in an end-to-end manner. 
They developed FlowNetSimple and FlowNetCorr.
FlowNetSimple takes as input both images stacked together and tries to predict the optical flow map.
FlowNetCorr takes each image as a separate input, processes them, and then uses a correlation layer to identify correspondences between the processed maps.
Both networks aim to minimize the endpoint error, which is the Euclidean distance between the predicted flow vector and the ground truth.
They mentioned that the amount of frames with ground truth in the Middlebury, Sintel, and Karlsruhe Institute of Technology and Toyota Technological Institute (KITTI) \citep{geiger2013vision} datasets is too small for training a deep CNN.
They resorted to creating a dataset called Flying Chairs consisting of 22,872 frame pairs with their respective ground truth.
They show that both FlowNet models outperform other architectures in the Flying Chairs dataset but struggle in other datasets.
The authors believe that FlowNetCorr can learn the features better and can outperform all other methods if trained on a realistically large dataset.

DELF \citep{noh2017large} is a method to calculate feature descriptors of image patches and keypoint selection.
It uses the convolutional layers of ResNet-50 to reduce the input image into feature maps of 1024 channels.
These are then assigned attention scores to select the spatially most distinctive features of the image.
The added attention layer was trained using images of the Google-Landmarks \citep{zheng2009tour} dataset by minimizing the cross entropy loss of the predicted class and its true label.
The dimensionality of the selected keypoints was later reduced to 40 dimensions using Principal Component Analysis (PCA).
They tested this algorithm on image retrieval of this dataset.
Its area under the precision-recall curve outperformed other state-of-the-art feature descriptors like Deep Image Retrieval (DIR) \citep{gordo2016deep} and CONGAS \citep{neven2008image}.
One of the main drawbacks of this method is that the granularity of positions of keypoints is effectively determined by the feature map before the attention layer. 
In other words, one can not calculate the descriptors of an exact location ($x$,$y$) of a keypoint in the original image. 

Detector-Free Local Feature Matching with Transformers (LoFTR) \citep{sun2021loftr} is an algorithm inspired by SuperGlue \citep{sarlin2020superglue}.
Like Scross-attentiones self and cross attention to find correspondences between keypoints.
The main difference between these two is that LoFTR does not require a detector and does pixel-wise dense matching at a coarse level and refining the good matches at a fine level. 
Similarly, they require a method to obtain the descriptors of keypoints.
In their experiments, they used Feature Pyramid Networks \citep{lin2017feature} (FPN) to obtain feature maps at multiple levels from the images.
FPN tries to build feature pyramids whose layers at all levels are equally semantically strong.
It does this by nearest neighbor upsampling from the lowest level of the pyramid of a traditional CNN.
This approach has been shown to improve the AP performance of the Faster R-CNN \citep{Ren2015} over its original backbone on the COCO dataset.
However, both LoFTR and FPN address the feature matching problem from a different perspective than ours.
FPN creates feature maps from an entire image which could contain multiple objects of different sizes and classes.
We aim to create an algorithm that, given a crop of an image with a feature/keypoint at its center, creates the most distinctive representation from all other possible crops.
It is worth noting that, to the extent of our knowledge, none of the feature constructors aim to build representations of features representing a part of an object.
Since an object is a dominant element in an image, it typically has textures, color changes, gradients, and shapes that make it significantly more distinctive that its surroundings.
In our case, we would like to distinguish between features of the same domain, $e.g.$ a wrinkle edge, a hair, a mole, from other skin features in an image whose object is a face (imaging BCG) or a hand (tremor quantification of PD patients).
This makes our problem inherently more difficult.
LoFTR is a matching algorithm based on these feature pyramids, which is an entirely different part of the feature matching pipeline.

SuperGlue \citep{sarlin2020superglue} is a graph neural network that can find matches between two sets of features by jointly finding correspondences and rejecting non-matchable points.
It is a transformer \citep{vaswani2017attention} based graph neural network that uses intra-image (self) and inter-image (cross) attention to understand the appearance of keypoints and the spatial relationships between them; respectively.
This algorithm takes as input the position and detection confidence of the keypoints in two images along with their respective descriptors and tries to find associations between them. 
While efficient and precise matching of keypoint pairs is essential in a feature-matching algorithm, our application focuses on the learning and formulation of feature representations.

\section{Materials and Methods}
\label{sec:methods}

\begin{figure*}[tb!]
\centering
\includegraphics[width=\textwidth]{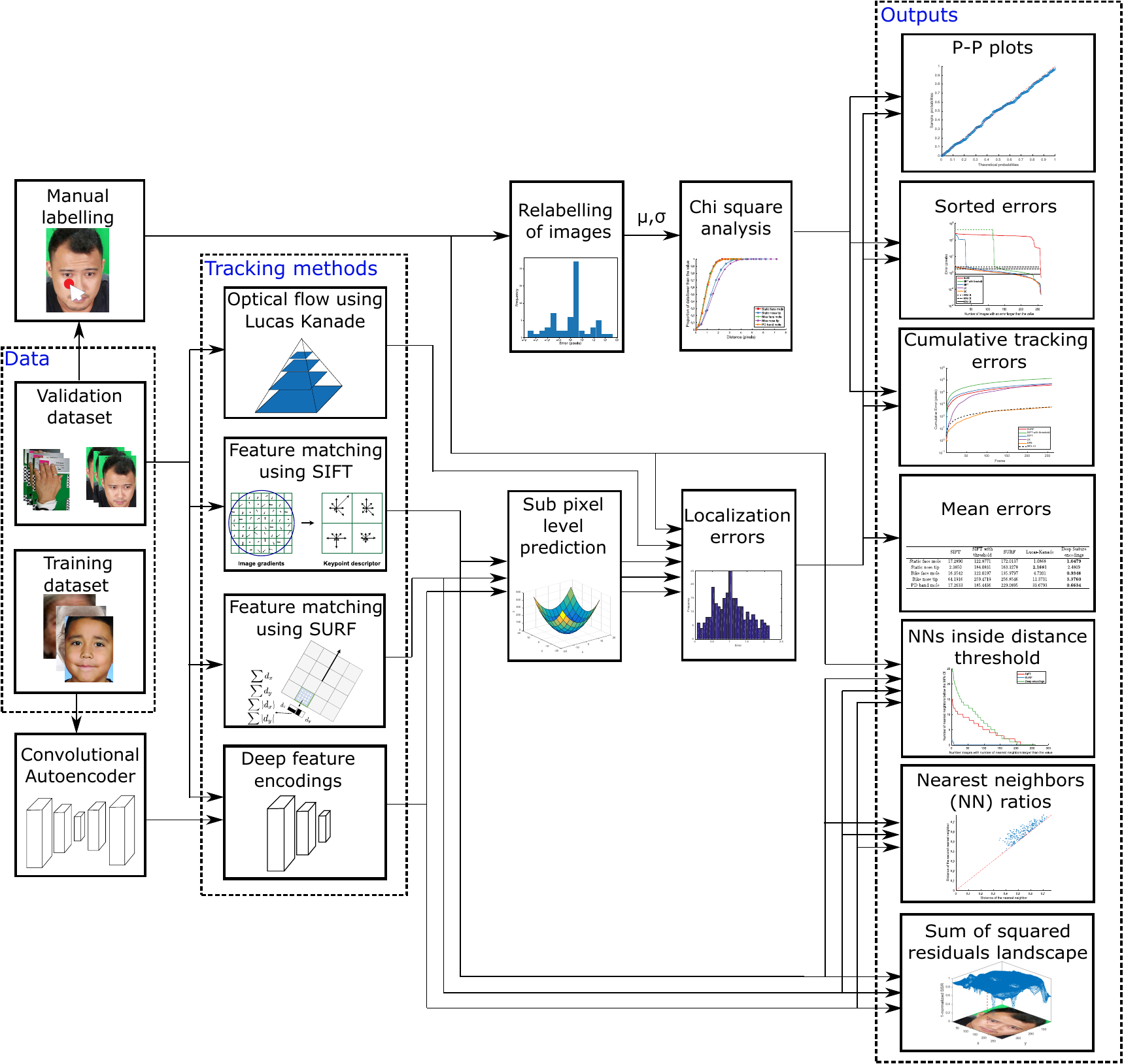}
\caption[Schematic workflow of our analyses and experiments.]{Schematic workflow of our analyses and experiments. The UTKface (training) dataset was used to train the autoencoder used in our DFE method. Our validation dataset was manually labeled to obtain the ground truth of the location of the features. It was relabeled several times and the mean and standard deviations of these localization errors were used to simulate a Chi-square distribution of the labelling errors. The feature tracking methods were used to predict the localization of the skin features in the validation dataset. For SIFT, SURF, and DFE the optimal match is determined to obtain subpixel level predictions before calculating the error relative to the manual labelling. Based on those errors the P-P plots, sorted errors, cumulative tracking errors, and mean errors are reported. The nearest neighbors inside the distance threshold, the nearest neighbors ratios, and the SSR landscape take the high-dimensional representations of the points as input and visualise them.}
\label{fig:ffpd_flowchart} 
\end{figure*}

\begin{figure*}[tb!]
\centering
\includegraphics[width=\textwidth]{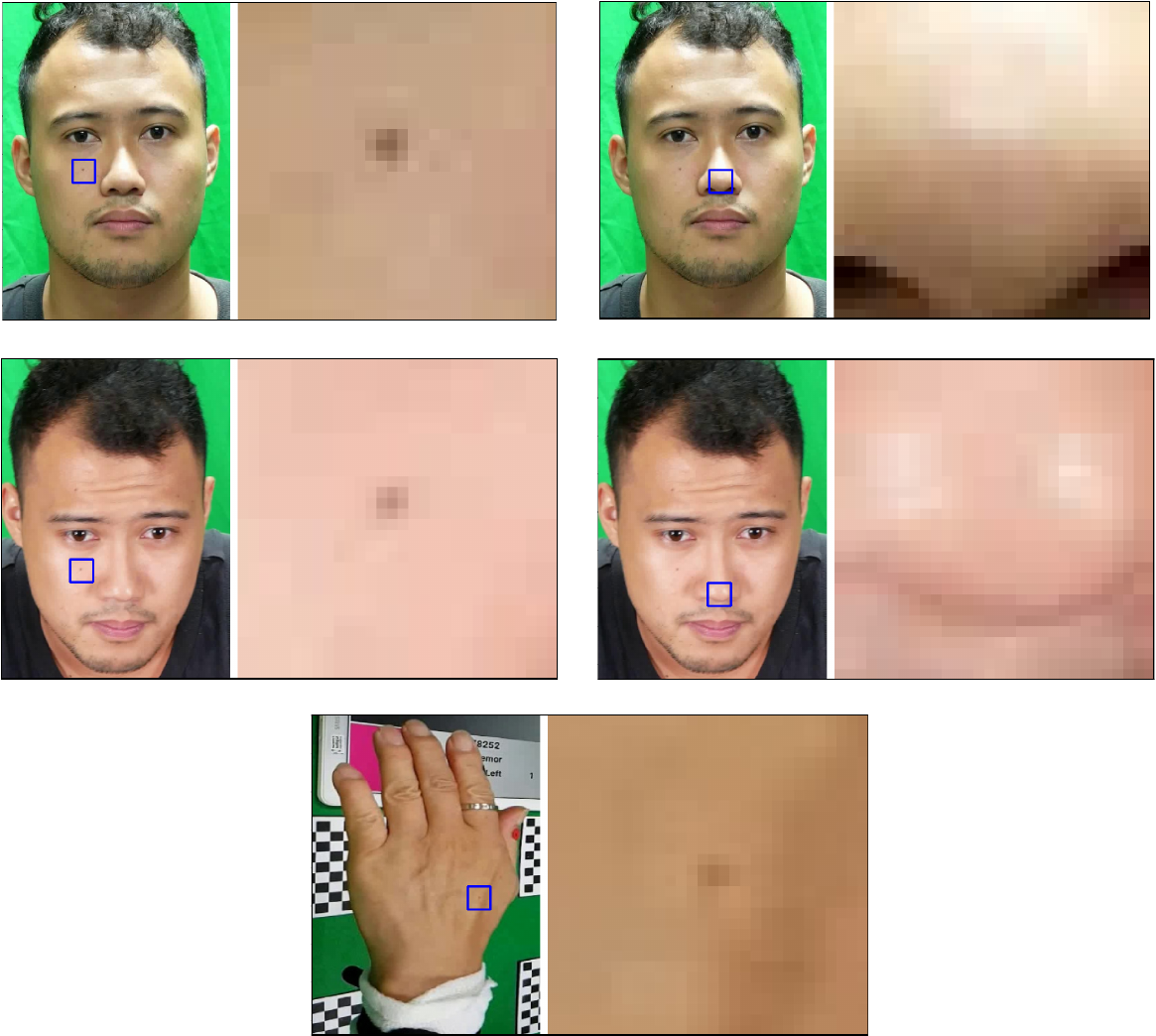}
\caption[Reference crops for static and bike videos.]{Reference crops of the face mole for static conditions (top-left), nose tip for static conditions (top-right), face mole for bike conditions (middle-left), nose tip for bike conditions (middle-right), and the hand mole of the PD patient marked (bottom)by the blue square of size $31 \times 31$ pixels and their magnifications for representative frames with resolution $420 \times 300$ pixels.}
\label{fig:reference_crops_marked_zoomed} 
\end{figure*}

Feature tracking can be simplified into doing feature matching successively for all frames in a video.
We want to compare the feature description ability and matching of different algorithms: SIFT, SURF, LK, and our Deep Feature Encodings (DFE).
For a full schematic of the flow of our experiments see Figure~\ref{fig:ffpd_flowchart}.
We decided to track two face features, under two different motion conditions, and one hand feature of a patient with Parkinson's disease (PD) performing a postural tremor test.
The facial features we decided to track are a face mole and a nose tip and the hand feature is a mole, see Figure~\ref{fig:reference_crops_marked_zoomed}.
These are visible in every frame in our validation dataset.

We used the default implementation of SIFT in Open CV, $i.e.$ the $128$-feature descriptor.
We expanded SURF to use rich descriptors ($128$-feature descriptor).
For LK we used the pyramidal implementation using a $10 \times 10$ window and a $4$-leveled pyramid.
For the DFE method, we constrained the autoencoder to a compression factor of $0.0444$ (128 latent variables) using $31 \times 31 \times 3$ crops as input.
Skin is rather homogeneous with distinct features in some places.
We, therefore, want to be able to select an area of distinct features to track, which is why it makes sense to work with patches.

\begin{figure}[tb!]
\centering
\includegraphics[width=\linewidth]{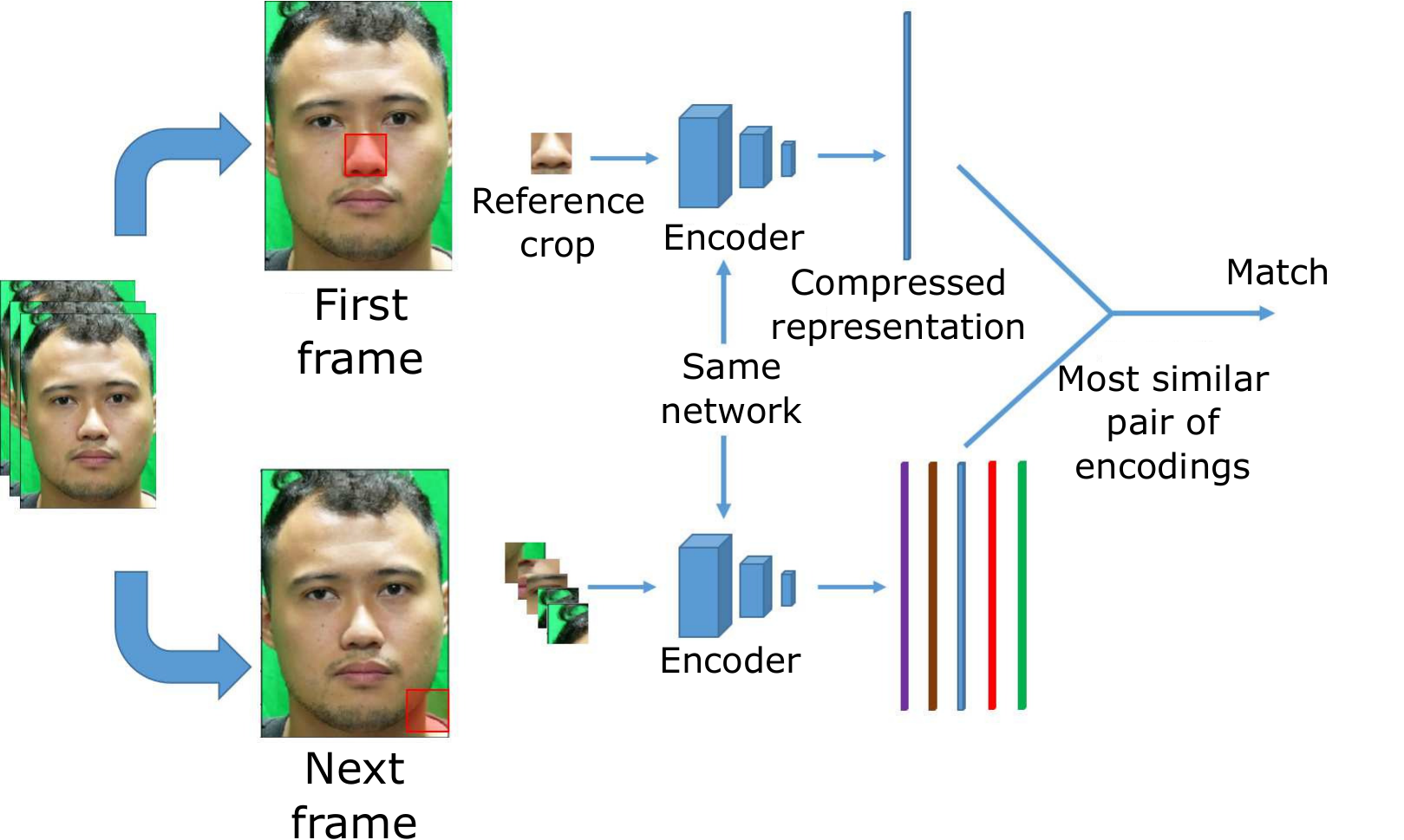}
\caption[Flowchart of the algorithm for using an autoencoder for matching face features.]{Flowchart of the algorithm for using an autoencoder for matching facial features.}
\label{fig:Autoencoder_matching_algorithm_flowchart}
\end{figure}

The task is to match a reference point from the first frame of the video (either face mole, nose tip, or hand mole) in every subsequent video frame.
For SIFT, SURF, and DFE; a match is made between the pair of descriptors $h$ that minimize the Sum of Squared Residuals (SSR),
\begin{align*}
SSR=\sum (h_{\mathrm{predicted}}-h_{\mathrm{reference}})^2,
\end{align*}
where $h_{\mathrm{predicted}}$ is the descriptors of a potential match in the second image and $h_{\mathrm{reference}}$ is the descriptor of the feature we want to track.
Both 
must be produced using the same method.
The matching strategy of DFE is shown in Figure~\ref{fig:Autoencoder_matching_algorithm_flowchart}.
For LK, the match is simply the point that minimises, in a least-squares sense, the optical flow equations.
Only DFE uses color information, coded in the CIELAB color space.
SIFT, SURF, and LK all use images in grayscale.

To evaluate the matching error of the methods, the images were manually labeled with pixel-level accuracy.
A Chi-square analysis was performed on the differences of the $x$ and $y$ coordinates in a series of relabeling attempts for a subset of the validation images to estimate the labelling error and assess the errors of the methods.

\subsection{Training dataset}
The training data for the autoencoder of our DFE method is the University of Tennessee, Knoxville Face (UTKFace) dataset.
UTKFace is a large-scale face dataset with a long age span (range from $0$ to $116$ years old) \citep{Zhang2017age}.
It consists of $24, 108$ face images in the wild of humans of varying sex, age, and ethnicity, see Figure \ref{fig:Resnet10_cropped_faces_UTKwild_validation_marked}.

All the wild face images are RGB color-coded but were converted into the CIELAB color space \citep{mclaren1976xiii}.
The CIELAB color space aims to be perceptually pseudo uniform.
Large differences in the color representations represent large color changes as perceived by humans.
The transformation from RGB to CIELAB is described in the section CIELAB color space (Appendix \ref{sec:cielab}).

The images have different backgrounds, lighting, postures, and resolutions.
Since the images are collected from the internet, some images seem to be black and white and some include some type of watermark.
There is a lot of variety in the face crops as the dataset includes faces with makeup, facial expressions, glasses, facial hair, and other head garments, $e.g.$ headbands, hats, bandanas, bindi, and earrings.

To train an autoencoder that is invariant to image scale and specialises in skin features, we have resorted to crops from the original images of the faces to obtain images of different resolutions.
We use the OpenCV ResNet-10 based model to detect the faces across all the $24,108$ pictures in the UTKface wild faces dataset and select the face with the highest confidence.
The results of the face cropping using the ResNet-10 based model are images of aligned cropped faces with the majority of the image corresponding to face information.

We subsequently, cropped image patches of size $31 \times 31 \times 3$ from all the detected faces with a stride of $30$.
These cropped patches serve as input for our deep convolutional autoencoder during the training phase.
With this partition, we have approximately $1.9$ million training crops.

\subsection{Validation dataset}

The videos we used to test our algorithms are all collected in our lab for two different projects.
The data can be divided into the following sets:
\begin{enumerate}
	\item \textbf{Remote ballistocardiography:} This dataset was obtained with the goal of estimating the heart rate remotely by tracking the subtle head movement produced by the blood flowing through the carotid artery.
The face videos are recorded in full HD resolution at $50$ FPS ($1920 \times 1080$ pixels) using a Panasonic GX85 with Olympus M.ZUKIO DIGITAL ED 14-150mm F4-5.6II lens \citep{Wang2020noncontact}.
These videos were recorded under two motion conditions:
	\begin{itemize}
		\item \textbf{Small motion--}The subject is instructed to simply look at the camera for a couple of minutes without moving, referred to as static condition.
		\item \textbf{Large motion--}The subject is filmed while riding an exercise bicycle. 
The motion is larger than in the static case and periodic. 
This is referred to as the bike condition.
	\end{itemize}
Both videos show the same subject performing the experiments.
The distance-to-pixel ratio was calculated from the physical distance between pupils (mm) and the estimated distance of pupils (pixels) from the first frame in each of the experiments.
To estimate the distance of pupils in the images, the two pupils were selected manually and the square root of the L2 norm of their residuals was calculated.
The subject has approximately 63 mm of the distance between his pupils.
In the static conditions, there are 219.23 pixels between pupils; equating to 0.287 mm/pixel.
In the bike conditions, there are 227.06 pixels between pupils; equating to 0.277 mm/pixel.

For the static conditions, the distance between pupils is 

Face cropping was done by first detecting all faces in the video frames using the Open CV ResNet10 model.
The cropping was based on the average bounding box center with width and height scaled at $1.5$ times the average width and height of all detected faces across the video frames.
This cropping scheme allows us to obtain images of the same size where the face dominates while allowing us to observe the motion of the head in the video.

	\item \textbf{Parkinson's disease postural tremor test:}
This data set of hands of PD patients has been recorded as a step toward digitalising the Unified Parkinson's Disease Rating Scale (UPDRS) motor exams.  
We use this video to evaluate the generalizability of our method to track skin features outside the face.
For more information on how the examination was administered see our protocol \citep{ashyani2022digitization}.
The subject was diagnosed with a Movement Disorder Society-Unified Parkinson's Disease Rating Scale (MDS-UPDRS) score for postural tremor on the left hand of one.
From the three available views (front, right, left) we took the left view of the left hand.
The hand video is recorded at $240$ FPS in HD resolution ($1280 \times 720$ pixels) using the back camera of a Samsung S7 mobile phone.
The area corresponding to the hand was manually cropped using a similar static window approach as for the face dataset, this video is referred to as PD.
\end{enumerate}

Both video datasets were coded in RGB color space but transformed to the CIELAB color space using the algorithm described in section CIELAB color space (Appendix \ref{sec:cielab}).
The videos were also subsampled to $2$ FPS to increase the displacement from frame to frame, $i.e.$ tracking challenge, and reduce the computation time.
The final frame count for the face videos after temporal subsampling is $261$ for the static condition and $170$ for the bike condition.
For the hand video, the final frame count was $40$ frames.
The images were reduced to $420 \times 300$ pixels to better match the images extracted from the UTKface dataset and to make the evaluation process computationally cheaper.

\subsection{Hardware used}

The training of the autoencoder was done in an Ubuntu server with an AMD Ryzen Threadripper 1950X 16-core processor with 32 threads and 64 GB of RAM.
The system also has three ASUS 1080 Ti GPUs to speed up the parallel computations in the training of the deep neural network.

\subsection{Chi-square analysis}
\label{par:chi_square}
A Chi-square analysis was used to establish a threshold for determining which errors were likely a result of human labelling.
Errors that are not due to human labelling errors are, by default, attributed to the flaws of the tracking method.

For each body part and motion condition, fifteen images were selected for each skin feature and relabeled five times.
These images were equispaced temporally and presented cyclically to the labeller.
This was done to prevent the labeller from seeing the same image multiple times in a row.
The relabelling was done roughly three months after the original labelling of the images and the labeller was not allowed to take breaks while relabeling each set of 15 images five times.
The original labelling was included to obtain a set of 90 samples for each body part and motion condition.

For each relabeling attempt, we computed the error against the mean of all the labeling attempts of the image separately for each image, body part, and motion condition.
This yielded a distribution of the errors for the $x$ and $y$ coordinates that is normal, see section Normality tests for relabeling errors in x and y directions (Appendix \ref{sec:normality_of_errors}).
Thus the sum of the standardized squared errors is Chi-square distributed. 
The standard deviations of $\delta x$ and $\delta y$ are different for each body part and motion condition, as shown in Table~\ref{tab:error_std}.
The standard deviations of the errors for the nose tips are larger than for the moles.
This is a result of the nose tip being a harder point to locate compared to a mole as it lacks the distinctive features necessary for a human to accurately pinpoint it from frame to frame.
Thus we call the moles distinctive skin features.

\begin{table}
\centering
\caption[Standard deviations of the error distributions for the manual relabeling attempts.]{Standard deviations of the error distributions for the manual relabeling attempts.}\label{tab:error_std}
\begin{tabular}{c c c }
\hline
 &$\sigma_x$ &$\sigma_y$\\
\hline
Static face mole& 0.773& 1.010\\
Static nose tip&  1.165& 1.256 \\
Bike face mole& 1.206& 1.179\\
Bike nose tip& 1.337& 1.319\\
PD hand mole& 1.162& 0.915\\
\hline
\end{tabular}
\end{table}

To assess whether the errors are due to human mistakes in the relabeling or proper to the tracking method, we used the following metric.
The test statistic, $\hat{\chi}^2$, is the sum of the standardized squared spatial errors 
\begin{align*}
\hat{\chi}^2  = \sum_{i=1}^{n}\frac{\delta x(i)^{2}}{\sigma_{x}^2}+\frac{\delta y(i)^{2}}{\sigma_{y}^2},
\end{align*}
where $n$ is the number of frames and $\sigma_x$ and $\sigma_y$ are the standard deviations of the errors of the manual labelling for each condition (Table~\ref{tab:error_std}) for the $x$ and $y$ directions, respectively.
We evaluated our test statistic on a Chi-square distribution with $2n$ degrees of freedom, $\chi^{2}(2n)$, to test our null hypothesis $H_0$ of the errors coming from human labelling.
Note that significance values smaller than $5 \times10^{-324}$ can not be calculated in $numpy$ and $math$ packages in Python due to the finite precision of doubles.

\subsection{Probability plot for Chi-square distributed errors}
A Probability--Probability plot (P-P plot) is a graph of the percentiles of one distribution versus the percentiles of another \citep{holmgren1995pp}.
For two distributions with CDFs $F_1$ and $F_2$ the P-P plot can be represented in functional form as $F_1(F_2^{-1}(p))$ where $p$ ranges between 0 and 1.
It is used to determine how closely the distributions of two datasets agree.
If two distributions are identical, then the values plotted in the P-P plot will lie on the $x=y$ line.
Deviations from this straight line mean that one distribution stochastically dominates the other.

We plot the standardized squared spatial errors
\begin{align*}
e_{std} =\frac{\delta x^2}{\sigma_x^2}+\frac{\delta y^2}{\sigma_y^2}
\end{align*} of DFE against a Chi-square distribution with $2$ degrees of freedom in Figure ~\ref{fig:chi2_pp_plots}.
The errors were standardized by the variances in Table~\ref{tab:error_std}.

\subsection{Optical flow using Lucas-Kanade method}

LK \citep{lucas1981iterative} uses spatial intensity gradients to direct the search for the best match.
It is still a widely used method for sparse optical flow given that it is computationally fast and efficient, making it suitable for many real-time applications.
It is assumed that the points in both images are in the same neighborhood and therefore have approximate registration, the brightness is relatively constant, and that distortions are linear.

For a small $m \times m$ window centered at point, $p$ the local image flow velocity vector $v$ must satisfy
\begin{align*}
\nabla I(q_i) \cdot v= -I_t(q_i),
\end{align*}
where $I_t(q_i)$ and $\nabla I(q_i)$ are the temporal and spatial gradients respectively for each pixel $q_i$ inside the $m \times m$ window.
Assuming the window contains $n$ pixels, denoted $q_1, q_2,...,q_n$, we can then write the equation on matrix form 
\begin{align*}
\nabla I v = -I_t.
\end{align*}
Here, the flow vector $v$ represents the original position of the point of interest, denoted $p = [p_x, p_y]$ plus the  image velocity vector $d=[d_x,  d_y]$ at $p$, such that
\begin{align*}
	v=\begin{bmatrix}
v_x\\
v_y
\end{bmatrix}=
\begin{bmatrix}
p_x+d_x\\
p_y+d_y
\end{bmatrix}.\
\end{align*}

To deal with cases where the motion is large, $i.e.$ more than one pixel, a typical approach is to apply LK on a pyramid of reduced-resolution images \citep{Bouguet2001}. 
The idea is to successively double the length of a pixel by halving the resolution and reducing fine image details that act as noise for large motion.
A pyramid of images has its raw image $I^0$ as the lowest level of the pyramid and every subsequent image $I^1, I^2, ..., I^{L_m}$ being the image from the level below sub-sampled to be half the width and height.
For the overall pyramidal tracking the algorithm first calculates the optical flow at the deepest level of the pyramid $L_m$ with the lowest resolution and uses that result as an initial guess for the next level $L_{m-1}$, the optical flow at this level is then used at $L_{m-2}$, and so on until level $0$ (the original image).
In our experiments, we used the implementation from OpenCV 3.4.2.17 with a pyramid of 5 levels, $L_m=4$, with a $10 \times 10$ observation window at each level.

\subsection{Feature matching using SIFT}
SIFT \citep{Lowe2004} is a method to create feature descriptors that are invariant to scale and rotation and enable robust feature matching across changes in viewpoint and illumination in the presence of noise and affine distortions. 
The idea is that the same keypoint in two images taken from, $e.g.$, different angles should have the same local information, i.e. gradients.
The SIFT algorithm for extracting and describing distinctive invariant features, $i.e.$ keypoints, can be divided into four parts: (i) construction of the scale space, (ii) keypoint localisation, (iii) orientation assignment, and (iv) generation of the keypoint descriptor. 
The keypoint descriptor of an image feature of interest is then compared to a database of keypoint descriptors for a second image to find the location of the same feature in it.
We next describe the four parts, excluding some algorithm details, to foster understanding.

The scale space is constructed by repeatedly halving the image size and blurring at multiple different scales.
To each image Gaussian blur is added to remove fine details and noise through convolution of the image with Gaussians, $G(x,y,\sigma)$, with different scales $\sigma$. 
This is implicit in the Difference of Gaussian (DOG)
\begin{align*}
	D(x,y,\sigma) = (G(x,y,k\sigma)-G(x,y,\sigma))*I(x,y),
\end{align*}
separated by a scaling factor $k$, which is used to find blobs.
The DOG pyramid is constructed one octave at a time by repeatedly taking every second pixel in each row and column of the previous image and calculating DOGs at different scales.

Keypoint localisation is based on finding the local scale space extrema in the constructed DOG pyramid of images and removing low contrast ones. 
Every extremum in its neighborhood of 26 surrounding pixels in the stack of images with different scales in each octave is considered a candidate keypoint.
These are rather poorly localized in spatial and scale space. 
To obtain the location with subpixel accuracy and filter out low contrast keypoints, a second-order Taylor expansion of $D(x,y,\sigma)$ is computed at the extremum point, $z_0 = (x_0,y_0,\sigma_0)$, by approximating the derivative and Hessian using differences of neighbouring sample points
\begin{align*}
	D(z_0 + z) = D(z_0) + D_z(z_0)^T z + 0.5 z^T D_{zz}(z_0) z.
\end{align*}
The location of the extremum, $\hat{z}$ is determined by taking the derivative and setting it to zero.
The value of the DOG at this localized extremum is  
\begin{align*}
	D(\hat{x}, \hat{y}, \hat{\sigma}) = D(z_0 + \hat{z}) = D(z_0) + 0.5 D_{zz}(z_0)^T \hat{z}.
\end{align*}
Keypoint candidates with $ \lvert D(\hat{x}, \hat{y}, \hat{\sigma}) \rvert$ less than a contrast threshold and poorly localized keypoints along an edge are discarded. 
Edge keypoints have a large principal curvature along the edge but a small principal curvature in the perpendicular direction, $i.e.$ large ratio of the eigenvalues of the Hessian which in the spatial space $(x,y)$ is defined as,
\begin{align*}
H=\begin{bmatrix}
D_{xx}& D_{xy}\\
D_{xy}&D_{yy}
\end{bmatrix}.
\end{align*}
Any keypoint with an eigenvalue ratio above the edge threshold is discarded.

An orientation is assigned to each keypoint so its descriptor can be made rotation invariant.
Based on the scale of the localized keypoint, $\hat{\sigma}$, the Gaussian smoothed image $L(x,y) = G(x,y,\sigma))*I(x,y)$ closest in scale is selected. 
Then the gradient magnitude $m(x,y)$ and orientation $\theta(x,y)$ are computed using pixel differences:
\begin{align*}
m(x,y)=\sqrt{(L(x+1,y)-L(x-1,y))^2}\\
\overline{+(L(x,y+1)-L(x,y-1))^2}\\
\theta(x,y)=\frac{tan^{-1}((L(x,y+1)-L(x,y-1))}{(L(x+1,y)-L(x-1,y)))}.
\end{align*}
A 36-binned orientation histogram with 10 degrees bin range is created from these for sample points in a region around the keypoint.
The peaks in the histogram represent the dominant direction of the local gradients.
The gradient magnitudes of the histogram are weighted by a Gaussian-weighted circular window with a $\sigma$ that is 1.5 times that of the scale of the keypoint.
Thus, the height of the bins represents the weighted sum of the gradient magnitudes in that area.   
Peaks correspond to dominant directions and the highest one determines the keypoint orientation.

Finally, the keypoint descriptor is generated.
For each keypoint, after the dominant orientation is found, its coordinates and gradient orientations are rotated relative to this orientation to achieve orientation invariance.
The magnitudes and orientations are then smoothed using a $16 \times 16$ weighted Gaussian window.
A new orientation histogram is then calculated from the orientations of the smoothed $4 \times 4$ subregions which contain the sum of the gradients and magnitudes within that space. 
The histogram divides the orientations into 8 bins and the height of each bin is the magnitude of that orientation.
This gives us $4 \times 4$ different histograms, each with $8$ bins.
The resulting feature descriptor consists of these histograms concatenated into a $4 \times 4 \times 8 = 128$ feature space.
The feature descriptor vector is then normalized to unit length to achieve illumination normalization.
We used the default implementation of SIFT in OpenCV 3.4.2.17.
The parameters used were \texttt{nfeatures = 0}, \texttt{nOctaveLayers = 3}, \texttt{contrastThreshold = 0.04}, \texttt{edgeThreshold = 10}, and $\sigma = 1.6$.

After obtaining the keypoint descriptors, SIFT matches a set of keypoints from one image to a set of keypoints in another image by comparing the L2 norm of their residuals.
In the original publication, SIFT used the Best Bin First \citep{Beis1997} algorithm to find approximate nearest neighbors and speed up the matching process.
This process does not always return the nearest neighbor.
Since we are interested in the ability of the methods to produce feature descriptors, we decided to use the global minimum of the L2 norm of the residuals instead.
 

We evaluated SIFT under two conditions.
First, forcing it to select the point minimising the SSR.
Second, allowing it to filter out points with the nearest neighbor to second nearest neighbor distance ratio greater than $0.8$ to remove false positive matches for the high-dimensional descriptors, as described in the original publication \citep{Lowe2004}.
In the second case, for the frames where the distance between the nearest neighbors exceeded the threshold, SIFT failed to find a matching feature.
We, therefore, assigned the largest possible error to such frames, $i.e.$ the length of the diagonal of the image 516.14 pixels.

\subsection{Feature matching using SURF}
SURF \citep{Bay2006} is an algorithm largely based on SIFT.
The main goal was to produce a faster feature detector-descriptor by reducing the computation time through several techniques to approximate the solutions used in SIFT.
In addition, it also uses by default a simpler, $64$-dimensional feature descriptor as opposed to the $128$-dimensional feature descriptor used in SIFT. 

The main differences between SIFT and SURF are the following: SIFT used the Difference of Gaussians to approximate the Laplacian of Gaussians, SURF pushes the approximation one step further and approximates it with box filters.
Instead of the feature vectors representing the magnitude and orientation of the gradients, the feature descriptors of SURF are the sum of the Haar responses (and the absolute value of the responses) for the parallel $x$ and normal $y$ directions relative to the keypoint orientation.
The Haar wavelet response is obtained by comparing how similar image patches are with square filters.
The keypoint orientation is defined by the longest vector (largest magnitude) of the sum of Haar wavelet responses in a 2D plane which is defined by the horizontal response in the abscissa and the vertical response in the ordinate.

The extended version of SURF (named SURF-$128$) computes the sums of the Haar responses in the parallel direction relative to the dominant keypoint orientation $d_x$ and $\lvert d_x\rvert$ separately for the Haar responses in the normal direction relative to the dominant keypoint orientation $d_y \geq 0$  and $d_y <0$.
Similarly, the sums of $d_y$ and $\lvert d_y\rvert$ are computed separately for $d_x \geq 0$  and $d_x <0$.
This yields a descriptor vector $v$ for each of the $4\times 4$ regions that is $8$ dimensional, as opposed to the original $4$ dimensional vector.
Consequently, the length of the concatenated descriptors is doubled to $128$ dimensions. 
We have used the extended version of SURF as implemented in OpenCV 3.4.2.17.
The parameter values are \texttt{hessianThreshold = 100}, \texttt{nOctaves = 4}, \texttt{nOctaveLayers = 3}, \texttt{extended = true}, and \texttt{upright = false}.

\subsection{Deep feature encodings}

DFE aims to use unsupervised deep learning to learn skin feature representations that are better to track.
We use an autoencoder that consists of an encoder subnetwork and a decoder subnetwork. 
The encoding subnetwork compresses the original data into a latent space that has typically fewer dimensions than the original input while the decoding subnetwork reconstructs the input from the compressed representation created by the encoder.
Formally, we can describe the network as a composition of an encoder function $f$ that creates a representation $h$ of our input data $x$, and a decoder function $g$ that reconstructs the input from the encoded data. 
We define the output reconstruction $y$ in terms of the input $x$ as
\begin{align*}
y(x)=g(f(x)).
\end{align*}
Both $f$ and $g$ are functions constructed using 2D convolutional \citep{lecun2010convolutional} and transpose convolutional \citep{zeiler2010deconvolutional} layers.
We used these as networks that have shown superhuman level performance for vision tasks all have this technology as their foundation \citep{Bochkovskiy2020,Ren2015,Liu2016ssd,he2017mask,Szegedy2015,He2016deep}.
For our skin feature tracking application, these encoded representations $h$ are the skin feature descriptors of image crops $x$ that we compare to find matches of keypoints between face images.

For our experiments, we want to constrain the autoencoder to have a latent space of $128$ features to make it a fair comparison with SIFT and SURF with rich descriptors.
This architecture yields a compression factor of $0.0444$ compared to the original input crop of size $31 \times 31 \times 3$.
The loss function used to train the autoencoder was the Mean Squared Error (MSE) between input $x$ encoded in the CIELAB space and its reconstructed output $y$ which is also encoded in the same color space.

\begin{figure*}[tb!]
\centering
\includegraphics[width=\textwidth]{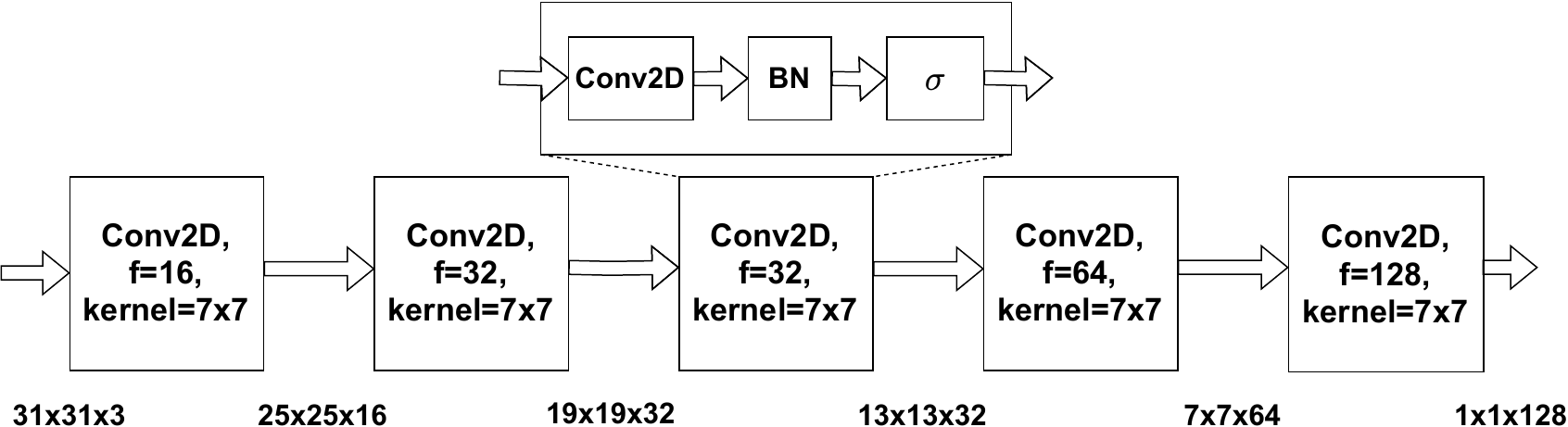}
\caption[Encoder architecture of the autoencoder used in our experiments.]{Encoder architecture of the autoencoder used in our experiments. The convolutional blocks are composed of three sequential operations: two-dimensional convolution (Conv2D), batch normalization (BN), and activation ($\sigma$). All convolutional layers are valid convolutions so the operation reduces the spatial dimensions of the input. The number of filters (f) and kernel size (kernel) in each block is given.}
\label{fig:experiment_encoder_architecture_flowchart}
\end{figure*}

To encode the $31 \times 31 \times 3$ image crops into a $128$-dimensional vector we use a symmetrical convolutional autoencoder architecture, see Figure~\ref{fig:experiment_encoder_architecture_flowchart}.
We downsample the spatial dimensions of the inputs by applying valid convolution operations and iteratively increase the number of channels by increasing the number of filters in our layers.
To avoid overfitting and improve generalization, we apply Batch Normalization just before applying the activation function and right after the convolution operation.
This allows the activation layer to receive unit Gaussian ($\mu=0$ and $\sigma=1$) inputs and for the network to control the diffusion of the outputs of the activation layer.
In addition, Batch Normalization can become the identity function if the network deems that no normalization is needed.
The nonlinearity used at each hidden layer is the Rectified Linear Unit (ReLU), which constrains all the activations to non-negative values while preserving the magnitude of all positive inputs. 

The network is symmetrical, so the decoder network is similar to the encoder network with the convolutional layers replaced by transposed convolutional layers to upsample the spatial dimensions of the input.
The last transposed convolutional layer only has three filters so that the shape of the output is identical to the input and the loss function can be calculated from the feature map-channel pairs. 
The activation function in the last layer of the network is a Sigmoid to constrain the output values between $0$ and $1$.
The optimizer Adamax is used, $i.e.$ a variant of Adaptive Moment estimation (Adam) \citep{Kingma2014}.
This network architecture contains $1,160,847$ trainable parameters.
It took approximately 138.9 hours to train it for 1000 epochs to a training MSE of $9.4 \times 10^{-5}$ on our hardware.

\subsection{Subpixel level prediction}
\label{sec:subpixel_level_prediction}
By default LK estimate the location with subpixel level accuracy.
Since we select the pixel whose feature descriptor minimizes the SSR with the reference descriptor, SIFT, SURF, and our DFE method only achieve pixel-level accuracy
To do a fair comparison of the methods, a subpixel level accuracy of the prediction from SIFT, SURF, and our DFE methods is needed.
We perform subpixel level prediction by fitting a surface to the SSR between the reference point and all points of the second image and interpolating locally within a neighborhood of the pixel with minimum SSR \citep{hajirassouliha2018subpixel}.

Let the SSR be represented as a function $z$, which is dependent on the $x$ and $y$ coordinates of points in the image, such that
\begin{align*}
z=f(x,y).\
\end{align*}
First, a pixel-level prediction is done by forcing the methods (SIFT, SURF, and DFE) to select the pixel with the minimum SSR through a global search for the minimum in the image space.
Second, to obtain subpixel level accuracy a surface of order $2$ is fitted to a $3 \times 3$ neighborhood of points around the global minimum by solving for $c$ in the equation
\begin{align*}
z=Ac,
\end{align*}
where $A=[1,x,y,xy,x^2,y^2]$ and $z$ contains the SSR of the nine points in the $3\times 3$ window.
To prevent the matrix $A$ from becoming ill-conditioned, we center the image coordinates around the $3 \times 3$ window by subtracting the coordinates of the middle pixel from all the pixels in the observation window.
The surface equation is given by
\begin{align*}
z=c_6y^2+c_5x^2+c_4xy+c_3y+c_2x+c_1.\
\end{align*}

It is possible that for a single image there are multiple points with the same minimum SSR.
In this case, we will select the point with the maximum curvature with a positive 2nd derivative $z_{xx}$.
This ensures that our prediction is a local minimum. 
The curvature $D$ is
\begin{align*}
D=z_{xx}z_{yy}-z_{xy}^2.
\end{align*}
This is identical to finding the determinant of a $2\times 2$ Hessian matrix 
\begin{align*}
Hz(x,y)=\begin{bmatrix}
z_{xx}& z_{xy}\\
z_{xy}&z_{yy}
\end{bmatrix}.
\end{align*}

If $D>0$ and $z_{xx}>0$, then according to Sylvester's criterion the Hessian $Hz(x,y)$ is positive definite and the surface has a local minimum.
Similarly, if $D>0$ and $z_{xx}<0$, then the surface has a local maximum.
If $D<0$, then the eigenvalues of $Hz(x,y)$ are of opposite sign, meaning the surface has a saddle point.
If $D=0$, then at least one of the eigenvalues of $Hz(x,y)$ is zero, yielding the test inconclusive \citep{abramowitz1988handbook, thomas1961calculus}. 
That means that there is no curvature in at least one direction, $i.e.$ L-shaped curve or flat surface.
Any point that is not a local minimum will be discarded. 
Also Lowe discard a point if it does not meet the curvature criterion \citep{Lowe2004}.

Once the point with the lowest SSR, $D>0$, and $z_{xx}>0$ is found, the local minimum of $z$ is the point where the gradients are zero in both directions, $i.e.$ $z_x=0$ and $z_y=0$.
We find this theoretical minimum of the surface by solving $\nabla f(x,y) = 0$, which yields
\begin{align*}
x&=\frac{c_3c_4-2c_2c_6}{4c_5c_6-c_4^2}\\
y&=\frac{c_2c_4-2c_3c_5}{4c_5c_6-c_4^2}.\
\end{align*}

\subsection{Tracking}
\label{sec:Tracking}

We evaluate the tracking of features using two extreme cases.
The first uses only the information from the first frame, $i.e.$ no update of the reference feature.
A divergent error is less common in this tracking scheme as the method is always searching for the same reference feature in all frames.
Even if the method matches the feature poorly, the error can be corrected due to the reference feature always remaining constant, but it is sensitive to change in the feature with time.
This tracking scheme is just repeated feature matching frame by frame.

The second is using only the information from the previous frame, allowing the algorithm to replace the reference feature.
This makes the tracking insensitive to gradual change in the feature with time.
The reference feature is the representation in the 128-dimensional space of the point we want to locate from the previous frame.
A divergent error is common as locating and associating features across many frames yields errors that can propagate when the method matches the feature poorly.
For the case where we evaluate SIFT by its nearest neighbor to nearest neighbor ratio, we track the last prediction for which the ratio was less than the threshold even if it was several frames ago.
That is to prevent SIFT from completely deviating.
If SIFT can not assign a prediction, we assign the squared maximum possible distance to that image.

The errors are squared and standardized by the variances in $x$ and $y$ directions of each body part and motion condition.
These are presented in a cumulative sum for each frame.
A 99\% CI line is plotted by evaluating the inverse of the $\chi^2$ CDF evaluated at a probability of 0.99 with $2 n_{\mathrm{frame}}$ degrees of freedom, where $n_{\mathrm{frame}}$ is the frame number.


\subsection{Analysis of feature matching}

\begin{figure}[tb!]
\centering
\includegraphics[width=\linewidth]{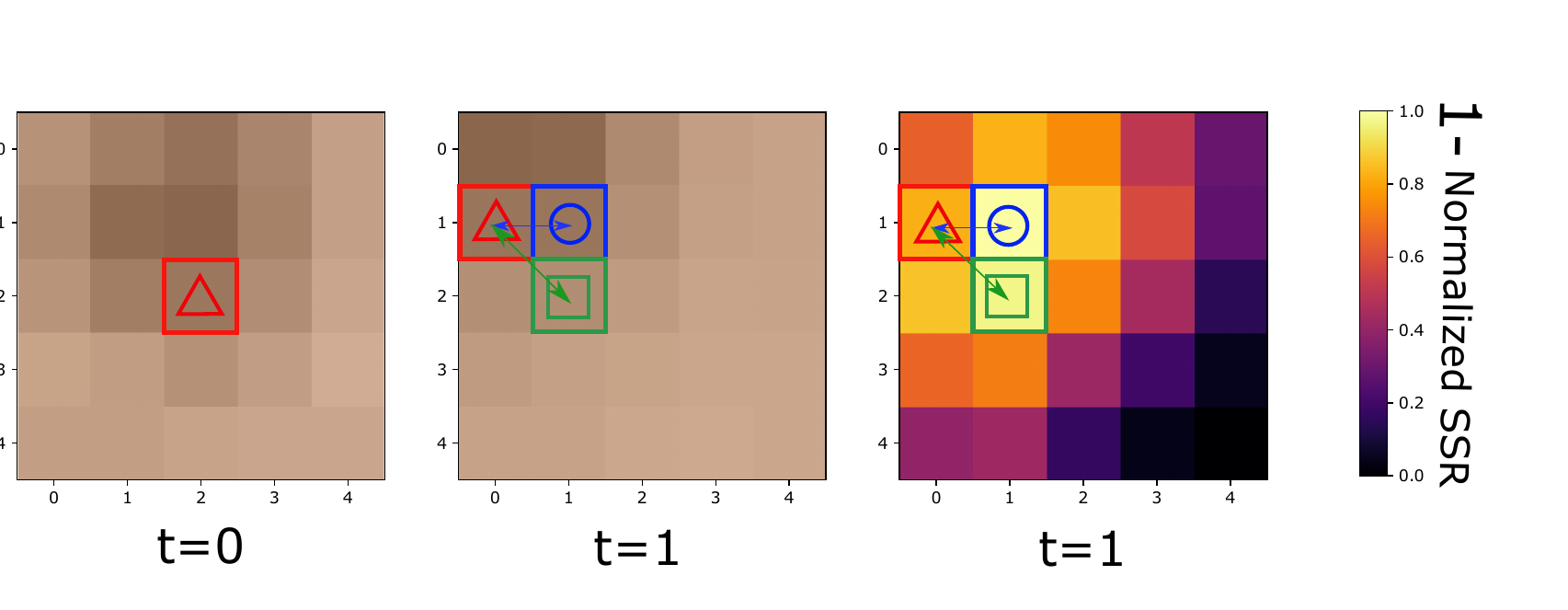}
\caption[Descriptor distance and spatial distance of nearest neighbors.]{Descriptor distance and spatial distance of nearest neighbors. The nearest neighbor (marked by blue circle) is the point with minimum SSR compared to the reference feature. By extension, the second nearest neighbor (marked by green square) is the point with the second-lowest SSR. In this case, the spatial distances to the ground truth (marked by red triangle) of the nearest and second nearest neighbor (marked by the blue and green arrows) are 1 and 1.44 pixels, respectively. The left image shows a $5 \times 5$ enlargement of the hand mole on the original image, the middle image shows the same feature in a second image with a displacement of 2 pixels to the left and 1 pixel up, the right image shows the SSR values of the displaced image. }
\label{fig:nn_descriptor_distance_spatial_distance}
\end{figure}

The aim of feature matching is to find the location of the feature of interest in another image based on similarity of the feature descriptors, which we measure using the sum of squared residuals.
The point with the smallest descriptor distance, $i.e.$ minimum SSR, is called the nearest neighbor.
The point with the second smallest SSR, the second nearest neighbor and so forth.
For feature matching to correctly locate the feature of interest the point with smallest descriptor distance should also have the smallest spatial distance to the true location of the feature of interest.
The relation between the spatial and descriptor distance is illustrated for the hand mole in Figure~\ref{fig:nn_descriptor_distance_spatial_distance}. 
The spatial distance refers to the Euclidean distance, $i.e.$ difference in position between two points in an image.

When calculating the distance between two high-dimensional vectors, certain variables contribute to the distance more than others.
SIFT employed a check for the nearest neighbor to second nearest neighbor ratio to avoid the high number of false positives that arise due to this phenomenon.
We investigated the ratio of the nearest neighbor to the second nearest neighbor based on the descriptor distance to the reference point for our DFE method (Appendix \ref{app:nearestneighorsratios}).

\section{Results}\label{sec:results}
\subsection{Feature matching accuracy}
\label{sec:feature_matching_accuracy}

Deep feature encoding alone located the skin features in every frame.
It also had the smallest mean error in all cases except the static nose tip, see Table~\ref{tab:mean_errors}.
For the static nose tip case, its performance is still comparable to SIFT and LK, with the latter being the best method in terms of the mean error.
SIFT with threshold and SURF did not perform well under any body part-motion combination.
The matching of all methods was made with sub-pixel level accuracy.
The frame count for the static, bike, and PD conditions was 260, 169, and 39 frames; respectively.
It is worth noting that the maximum possible distance in all cases is 516.14 pixels, equaling the length of the diagonal of the images of size $420 \times 300$ pixels.

\begin{table*}[tb!]
\footnotesize
\centering
\caption[Mean errors (in pixels).]{Mean errors (in pixels) for feature matching using the same reference feature. The best results are highlighted in bold. SIFT* stands for SIFT with the nearest neighbors threshold.}
\label{tab:mean_errors}
\begin{tabular}{C{2.5cm} C{1cm} C{1cm} C{1cm} C{1 cm} C{1cm} }
\hline
 &SIFT & SIFT* &SURF & LK& DFE  \\
\hline
Static face mole& 17.28 & 230.09& 93.84 & 1.08 & \textbf{1.04}\\
Static nose tip& 2.37 & 385.69& 48.13 & \textbf{2.16}& 2.47\\
Bike face mole& 16.33 & 216.25 & 74.93  & 4.68& \textbf{0.90} \\
Bike nose tip& 64.12 & 421.99 & 116.63 &11.31&  \textbf{3.31}\\
PD hand mole&17.29  & 227.19  &85.27  & 33.67  &\textbf{0.64}\\
\hline
\end{tabular}
\end{table*}

\begin{figure*}[tb!]
 \centering
 \includegraphics[width=\textwidth]{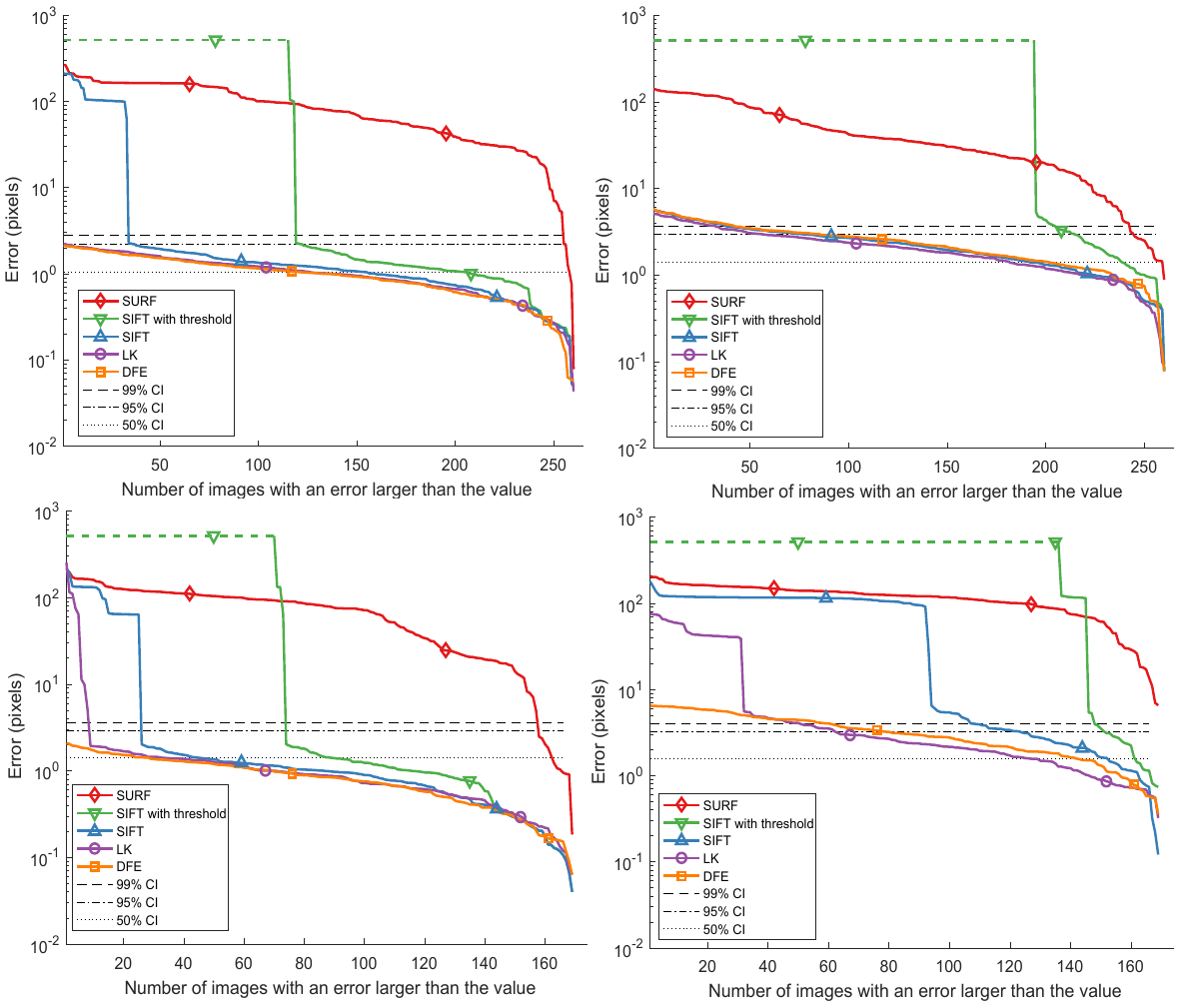}
\caption[Sorted errors for matching features across frames.]{Sorted errors for matching the face mole under static conditions (top-left), nose tip under static conditions (top-right), face mole under bike conditions (bottom-left), and nose tip under bike conditions (bottom-right). The black lines represent the 50\%, 95\%, and 99\% CI from the simulated Chi-square distribution. The green dashed line for SIFT with threshold stands for the frames where the nearest neighbor distance threshold was larger than $0.8$.}
\label{fig:sorted_matching_errors}
\end{figure*}

The distance errors (in pixels) of all methods sorted in descending order  for each body part and motion condition is shown in Figure~\ref{fig:sorted_matching_errors}.
We note that LK is comparable to DFE for the static conditions.
The best performance for SIFT was while tracking the nose tip under static conditions.
SURF and SIFT with the recommended nearest neighbor to second nearest neighbor threshold did not perform well in any of the scenarios we tested.
Our Deep feature encodings method was the only method to perform well in every condition.
Even while tracking the nose tip under bike conditions the error was at most $6.5$ pixels.
For the distinctive features, i.e. moles, the error was at most $2.1$ pixels, while it exceeded $219$ for the other methods in some conditions. 

\subsection{Tracking}
\label{sec:tracking}
\begin{figure*}[tb!]
 \centering
  \includegraphics[width=\textwidth]{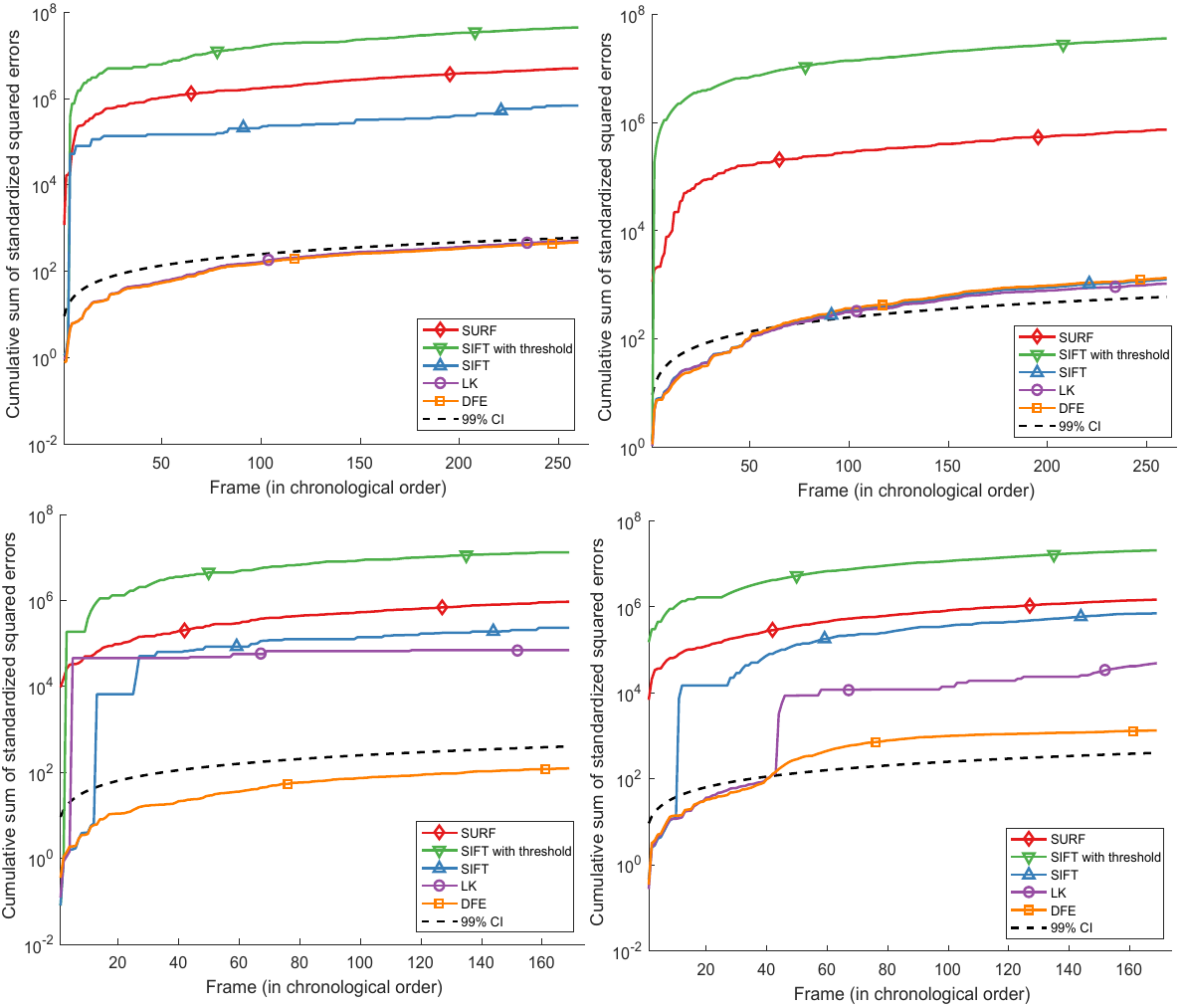}
\caption[Cumulative sum of standardized squared errors for tracking features using the reference feature from the original image.]{Cumulative sum of standardized squared errors using the reference feature from the original image for tracking the face mole under static conditions (top-left), nose tip under static conditions (top-right), face mole under bike conditions (bottom-left), and nose tip under bike conditions (bottom-right). The black dotted line marks the 99\% CI line.}
\label{fig:tracking_errors_ofeat}
\end{figure*}

\begin{figure*}[tb!]
 \centering
  \includegraphics[width=\textwidth]{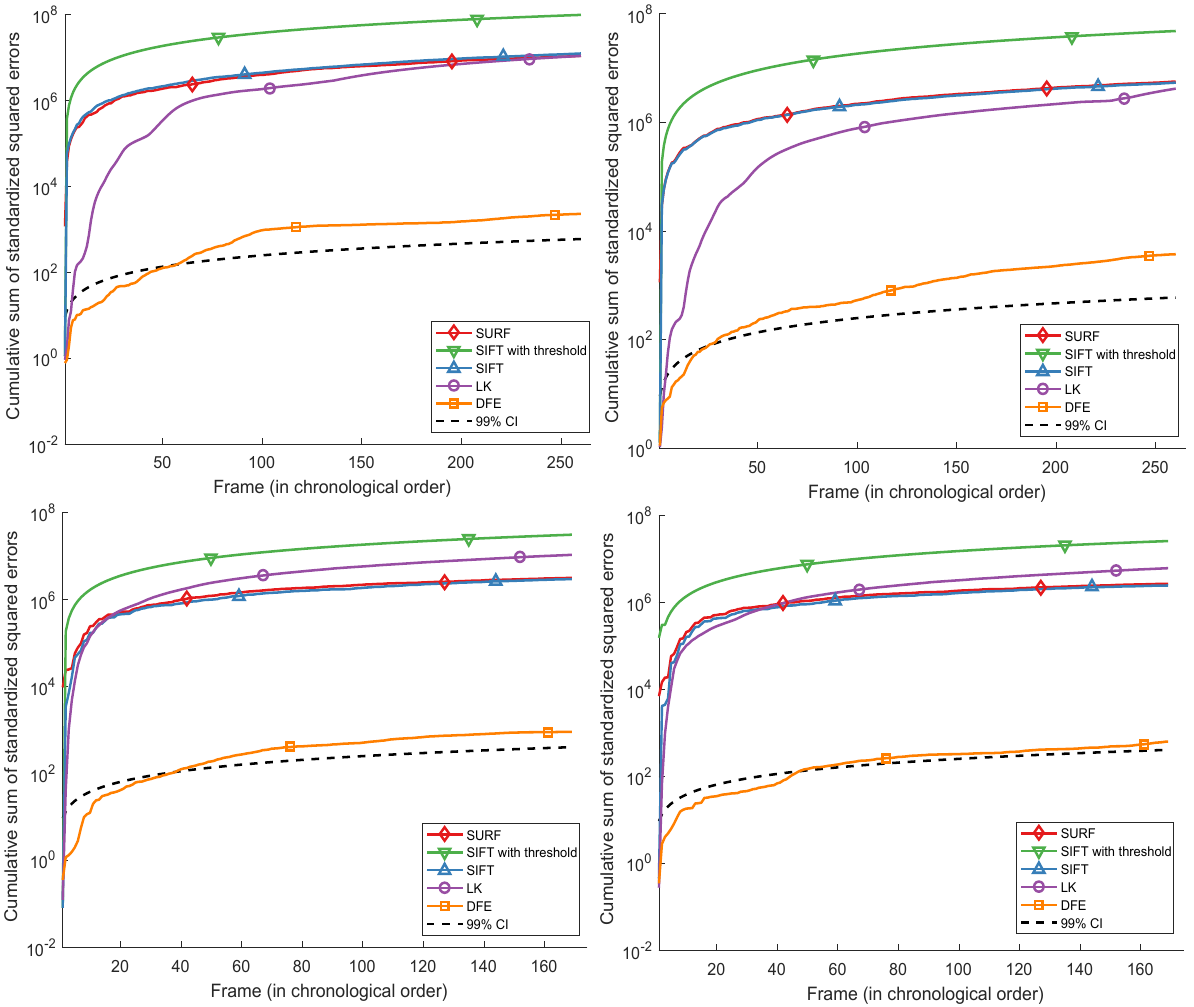}
\caption[Cumulative sum of standardized squared errors for tracking features using the prediction of the previous image as reference feature.]{Cumulative sum of standardized squared errors using the prediction from the previous image as reference feature for tracking the face mole under static conditions (top-left), nose tip under static conditions (top-right), face mole under bike conditions (bottom-left), and nose tip under bike conditions (bottom-right). The black dotted line marks the 99\% CI line.}
\label{fig:tracking_errors_pfeat}
\end{figure*}

The results for tracking only the reference feature from the original image is shown in Figure~\ref{fig:tracking_errors_ofeat} and replacing the reference feature with the prediction from the previous frame in Figure~\ref{fig:tracking_errors_pfeat}.
The average errors using the reference feature from the original image are shown in Table~\ref{tab:mean_errors} and replacing the reference feature with the prediction from the previous frame in Table~\ref{tab:mean_errors_previous}.
In general, DFE is clearly the best method for tracking, achieving the lowest mean error for all conditions except the nose tip under static conditions.
In the former case, the cumulative sum of standardized squared errors of DFE even remain within the 99\% confidence interval for the distinctive skin mole. 
In the latter case, it exceeds it, but all other methods have mean errors greater than $100$ pixels.
The large mean errors are due to the divergence problem discussed in section \nameref{sec:methods}.
In this tracking scheme, LK also diverges in all conditions, while it in the former scheme did not for the static condition.
This renders it not useful for tracking.
In contrast, DFE remains useful as it did not diverge; even when all other methods did.
The ratios of nearest to second nearest neighbors for SIFT with threshold consistently exceeded the $0.8$ as time increased.
This is a result of the method diverging to the point where it is no longer tracking features that are distinctive enough for it.

\begin{table*}[tb!]
\centering
\footnotesize
\caption[Mean errors (in pixels) for tracking using the prediction of the previous image as a reference feature.]{Mean errors (in pixels) for tracking using the prediction of the previous image as a reference feature. The best results are highlighted in bold. SIFT* stands for SIFT with the nearest neighbors threshold.}
\label{tab:mean_errors_previous}
\begin{tabular}{C{2.5cm} C{1cm} C{1cm} C{1cm} C{1 cm} C{1cm} }
\hline
 &SIFT & SIFT* &SURF & LK& DFE  \\
\hline
Static face mole& 191.81 & 514.16& 178.66 & 179.99 & \textbf{2.15}\\
Static nose tip&70.62 & 514.16& 173.83 & 135.94& \textbf{4.24}\\
Bike face mole& 144.41 & 513.09 & 149.95  & 293.76& \textbf{2.56} \\
Bike nose tip& 142.41 & 513.09 & 150.49 & 244.70 &\textbf{2.30}\\
PD hand mole&142.22 & 502.92  &150.16  & 162.07  & \textbf{1.30}\\

\hline
\end{tabular}
\end{table*}

\subsection{Probability plot for Chi-square distributed errors}

\begin{figure*}[tb!]
 \centering
 \includegraphics[width=\textwidth]{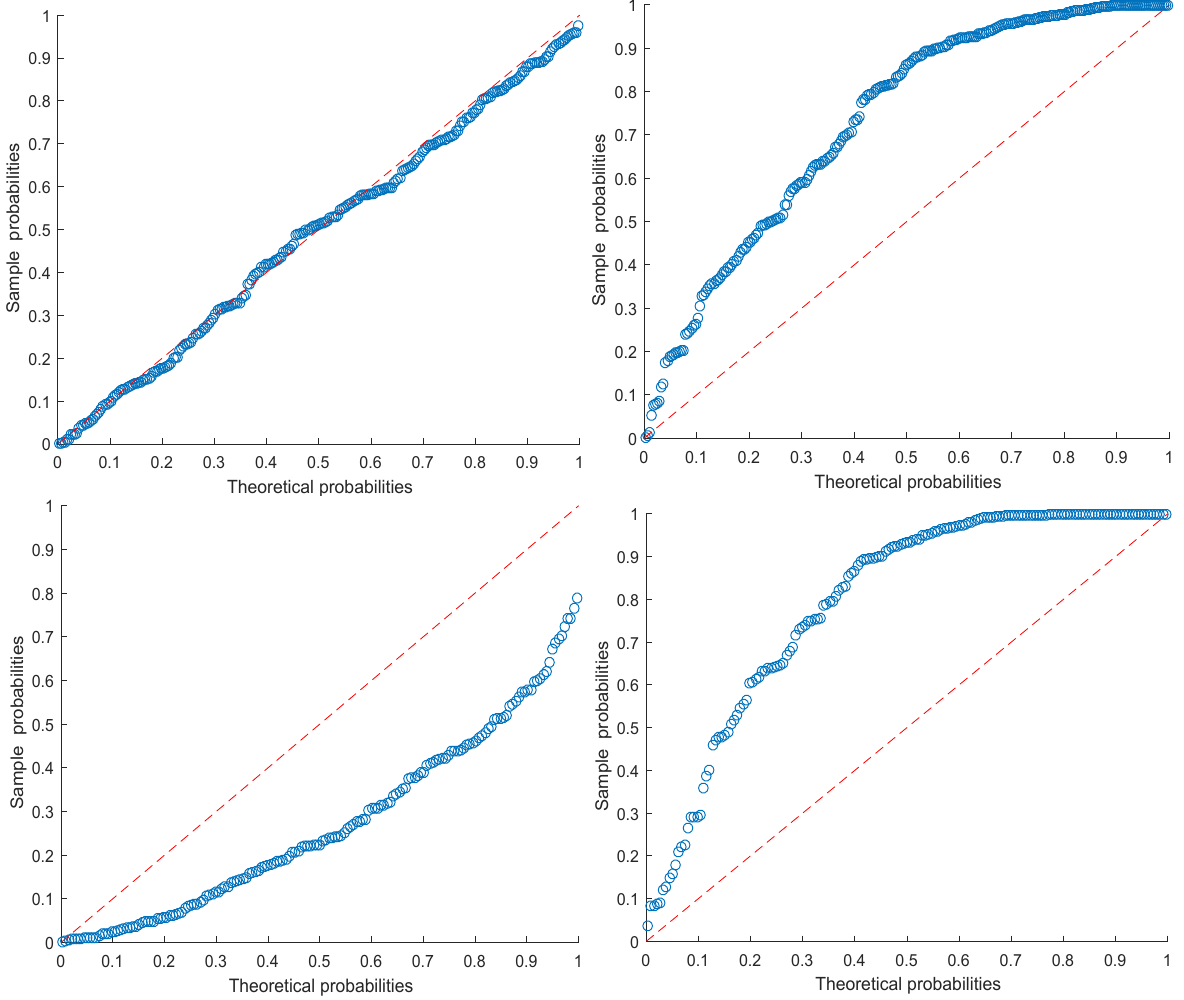}
\caption[P-P plot for the standardized squared spatial errors for DFE.]{P-P plot for the standardized squared spatial errors for tracking the face mole under static conditions (top-left), nose tip under static conditions (top-right), face mole under bike conditions (bottom-left), and nose tip under bike conditions (bottom-right) for DFE. The red dashed line marks the $x=y$ line and deviations from this line represent deviations from the theoretical Chi-square distribution.}
\label{fig:chi2_pp_plots}
\end{figure*}

For tracking the moles the empirical data falls on or below the $x=y$ line of the P-P plot in Figure~\ref{fig:chi2_pp_plots} and~\ref{fig:results_pd}, while it for the nose tip falls above the $x=y$ line.
This means that for the face mole, the errors of DFE are likely due to human labelling.
On the other hand, the errors for the nose tips are likely in part due to the method; as is the case for all other methods, see Table~\ref{tab:sig_level_ho_human_errors}.
This further emphasizes our assumption that the mole is a more distinctive feature that is easier to track.

The significance level at which we can reject the null hypothesis of the errors coming from human labelling for all conditions, previously described in section \nameref{par:chi_square}, can be seen in Table~\ref{tab:sig_level_ho_human_errors}.
This hypothesis cannot be rejected for DFE when tracking the moles, nor for LK when tracking the face mole under static condition. 
The $\hat{\chi}^2$-test statistic is so small that the significance level is near one for DFE.
The significance level of SURF for all conditions was smaller than the minimum precision of doubles of $5.0 \times 10^{-324}$.

\begin{table*}[tb!]
\centering
\footnotesize
\caption[Significance level at which the null hypothesis of the errors coming from human labelling is rejected.]{Significance level at which the null hypothesis of the errors coming from human labelling is rejected. 
Values larger than 0.01, representing the threshold to reject the null hypothesis with 99\% confidence, are highlighted in bold. $\epsilon$ stands for a value smaller than the smallest increment of doubles, $i.e.$ $5.0 \times 10^{-324}$.}
\label{tab:sig_level_ho_human_errors}
\begin{tabular}{C{2.5cm} C{1.6cm} C{1.6cm} C{1.6cm} C{1.6cm} }
\hline
 &SIFT &SURF & LK & DFE\\
\hline
Static face mole&  $\epsilon$ &  $\epsilon$ &  \textbf{0.64}&  \textbf{0.95}\\
Static nose tip&  2.6\sci{e-62} &   $\epsilon$  &  3.7\sci{e-38} &  7.9\sci{e-73} \\
Bike face mole&  $\epsilon$ &  $\epsilon$ & $\epsilon$ &  \textbf{1.0}\\
Bike nose tip&  $\epsilon$ &   $\epsilon$ &  $\epsilon$ & 4.4\sci{e-118}\\
PD hand mole& $\epsilon$   &  $\epsilon$   & $\epsilon$  &  \textbf{1.0}\\
\hline
\end{tabular}
\end{table*}

\subsection{SSR as a function of spatial coordinates}
\begin{figure*}[tb!]
 \centering
 \includegraphics[width=\textwidth]{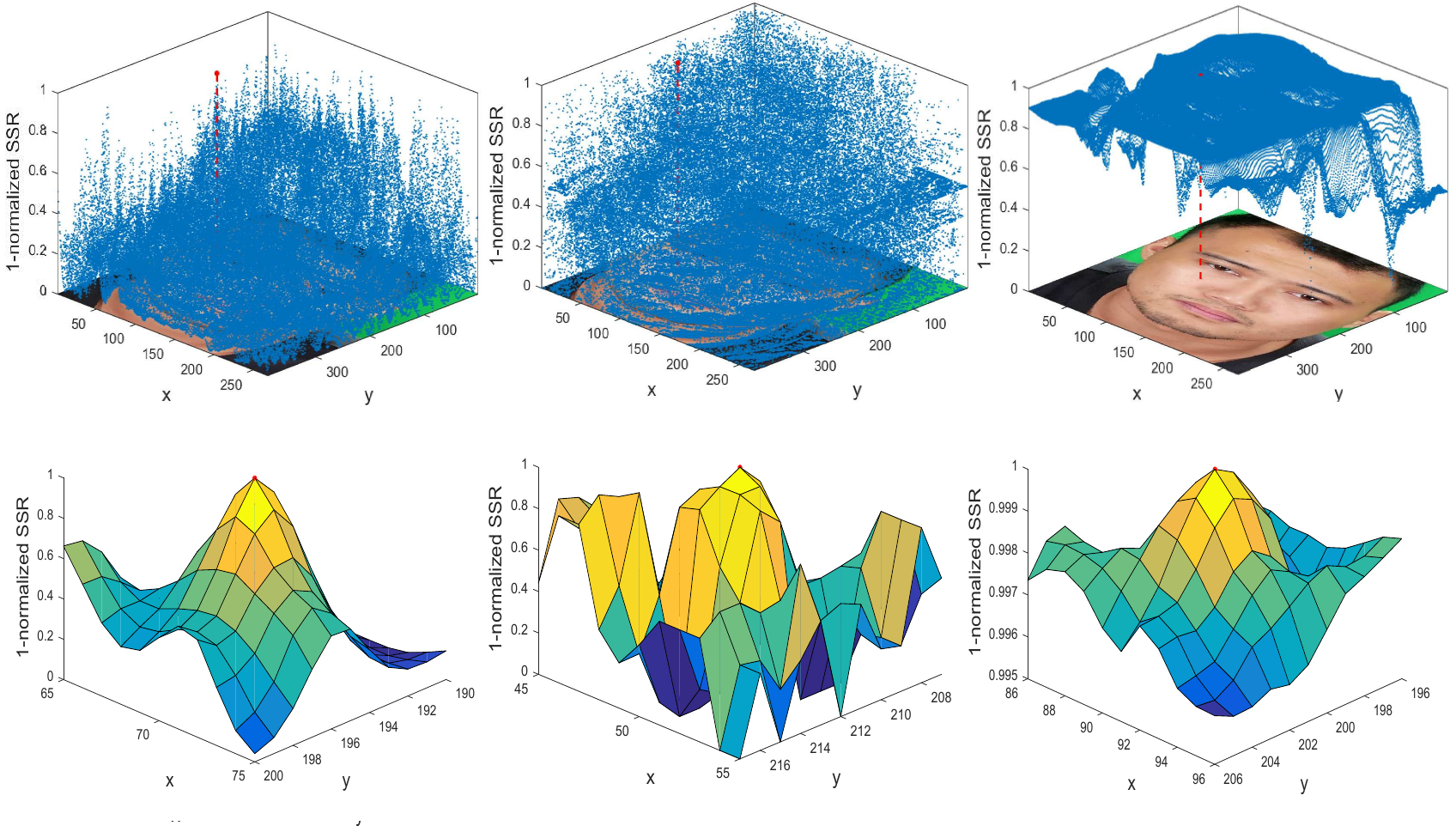}
\caption[SSR as a function of the spatial image coordinates.]{SSR as a function of the spatial image coordinates for the images with the smallest localization errors for tracking face moles under bike conditions for SIFT (left), SURF (middle), and DFE (right) and their corresponding regional enlargements of the optimum. The red dashed lines mark the point in the space with minimum SSR, which is plotted as a red circle. The distances to the ground truth are 0.04, 0.98, and 0.06 pixels, respectively.}
\label{fig:spatial_ssr_best}
\end{figure*}

We looked into the extreme cases for localization performance of SIFT, SURF, and DFE methods.
Figure~\ref{fig:spatial_ssr_best} shows the SSR landscape of the frame with the smallest localization error for tracking the face mole under bike conditions.
We chose this condition because for the best cases the localization error of all methods is below one pixel of distance. 
We can see that the SSR landscape of SIFT and SURF is chaotic and does not have any clear structure or pattern.
This is in contrast to the SSR landscape of our DFE method which has smaller values near the face.
This observation holds independent of the methods localization performance (Appendix \ref{app:spatialSSRlandscape}).
This suggests that our method is better at creating feature descriptors for skin features than the other state-of-the-art methods.
A structured SSR landscape will prevent the algorithm from selecting a false positive which is spatially far from the ground truth.

\subsection{Generalisation to PD data}
\begin{figure*}[tb!]
 \centering
  \includegraphics[width=\textwidth]{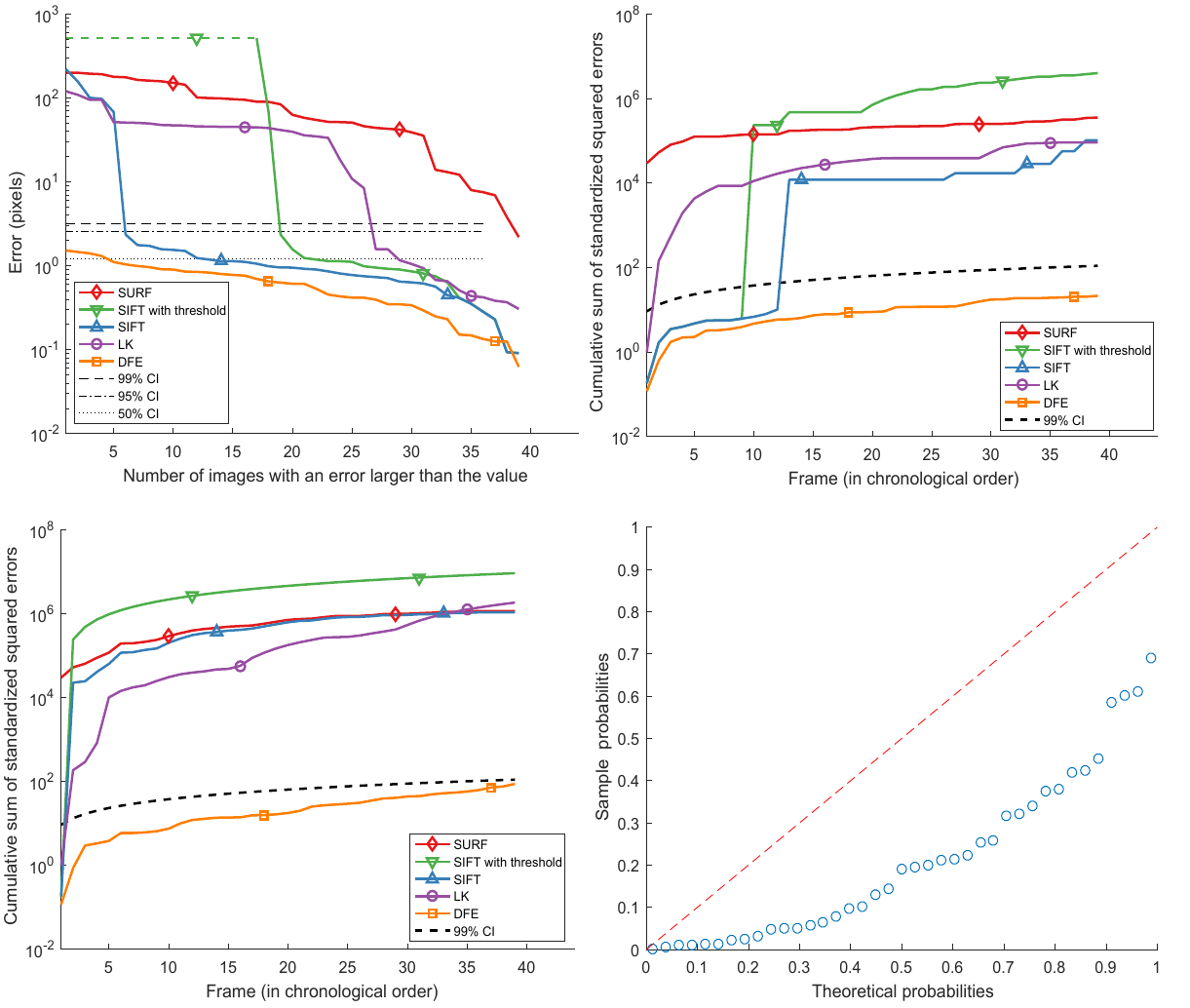}
\caption[Results from the methods for tracking the mole on the hand of the PD patient.]{Results of running the methods for tracking the mole on the hand of the PD patient. Sorted errors for tracking the mole (top-left), cumulative error for tracking using the reference feature from the original image (top-right), cumulative error for tracking using the reference feature from the previous image (bottom-left), and  P-P plot for the standardized squared spatial errors (bottom-right).}
\label{fig:results_pd}
\end{figure*}

\begin{figure*}[tb!]
 \centering
  \includegraphics[width=\textwidth]{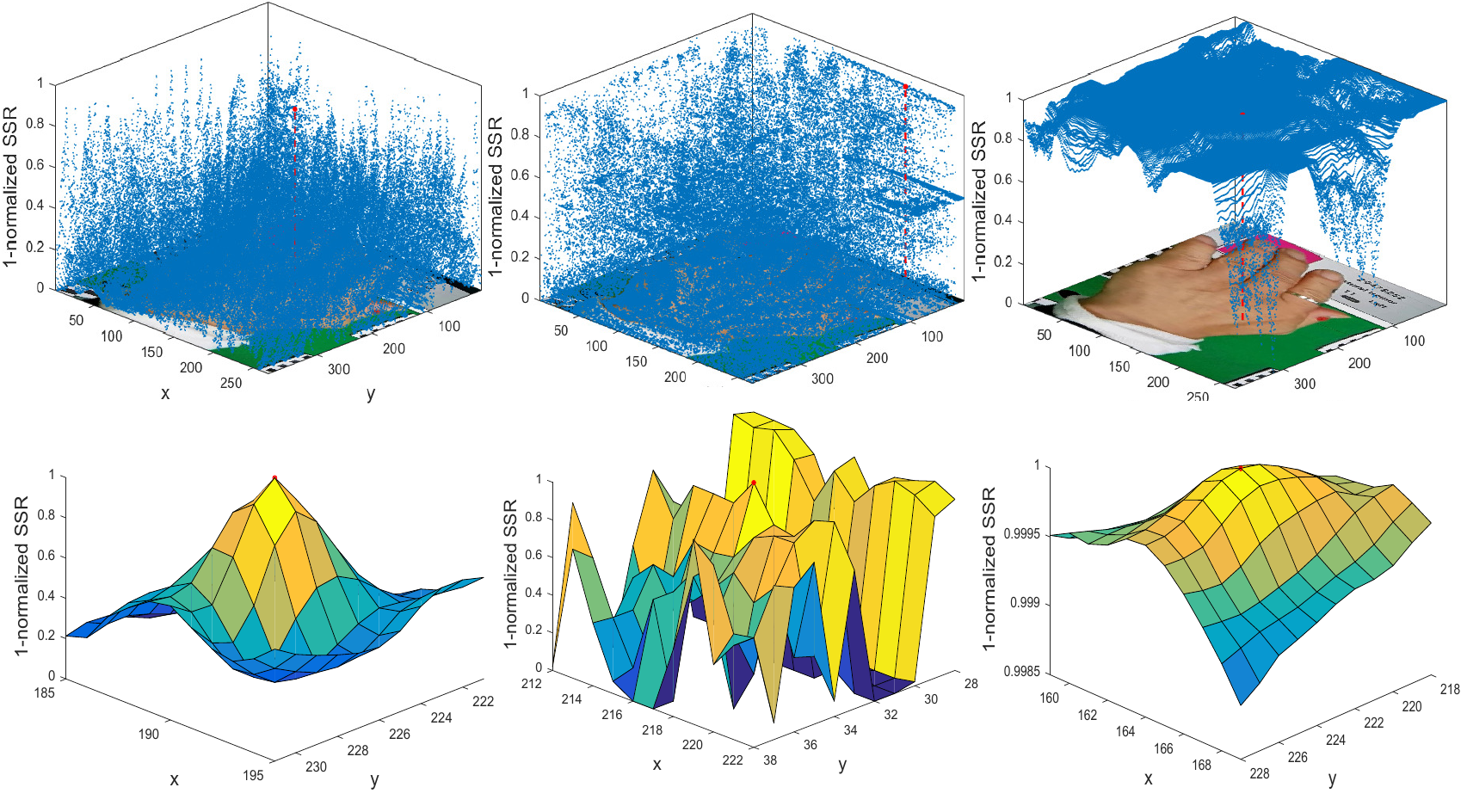}
\caption[SSR as a function of the spatial image coordinates.]{SSR as a function of the spatial image coordinates for the images with the smallest localization errors for tracking the mole on the hand of the PD patient for SIFT (left), SURF (middle), and DFE (right) and their corresponding regional enlargements of the optimum. The red dashed lines mark the point with minimum SSR. The distances to the ground truth are 0.09, 196.12, and 0.06 pixels, respectively.}
\label{fig:spatial_ssr_best_pd}
\end{figure*}

The performance of the methods is qualitatively the same as for the face mole under bike conditions based on a comparison of Figures~\ref{fig:results_pd},~\ref{fig:sorted_matching_errors}, ~\ref{fig:tracking_errors_ofeat}, ~\ref{fig:tracking_errors_pfeat}, and ~\ref{fig:chi2_pp_plots}.
DFE is still the best method for tracking the mole on the hand of the PD patient.
In contrast to Figure~\ref{fig:sorted_matching_errors}, the second-best method is SIFT, followed by LK.
Although the images in this dataset are qualitatively similar and there is little distortion due to motion, both SIFT and LK for some images yield errors exceeding 100 pixels. 
This makes the result qualitatively closer to the face mole under bike conditions than static conditions.
SURF again failed to track the mole. 
For DFE the errors are likely due to human labelling and not the method itself since the sampled errors are lower than the $x=y$ line in the P-P plot.
Figure~\ref{fig:spatial_ssr_best_pd} shows that the SSR landscape of DFE has more structure that the landscapes produced by either SIFT or SURF for the image with smallest error.
The tracking results are also qualitatively similar to the ones shown in Figures ~\ref{fig:tracking_errors_ofeat} and ~\ref{fig:tracking_errors_pfeat}.
These show that DFE is the best tracking method, no matter what tracking scheme is employed, as it was the only method that did not diverge.
It also shows that usage of the reference feature from the original image resulted in a smaller cumulative error.
These results are qualitatively similar to Figures~\ref{fig:tracking_errors_ofeat} and ~\ref{fig:tracking_errors_pfeat}.

\section{Discussion and Conclusions}\label{sec:conclusions}

We present an algorithm that is more accurate and precise, and more importantly, reliable for skin feature matching than the traditional Computer vision methods under the conditions of interest for our ballistocardiography and Parkinson's disease applications.
It is more precise thanks to the global minima containing several nearest neighbors within the range of acceptable predictions without the inherent problem of a large number of false positives that arise from calculating the similarity of two high-dimensional feature vectors in SIFT and SURF.
The distinctiveness of the feature representations can be attributed to three main factors: preservation of spatial information through the convolution operation, learning of complex features through hierarchical feature learning, and the creation of a set of latent variables that is able to capture the variance of the samples.
As demonstrated in our experiments, when the matching method makes an obviously wrong prediction of the location of the skin feature, the error is usually extremely large.
This will cause tracking algorithms to diverge.

To ensure that our testing is exhaustive, yet simple, and that large changes cannot cause trouble, we currently do not search locally for features but rather do a global search for the best match.
Thus, our method is not computationally optimized.
It currently takes approximately four times longer than SIFT to run on the validation data with our hardware.
We could assume that the skin features have approximate registration and perform a local search to speed up the computation in the future.
In addition, the matching scheme could be sped up by implementing an approximate technique for matching neighbors as it has been demonstrated in SIFT.
The search could also be optimized if we took advantage of the convexity of the search space, which we have locally for DFE in all cases.
We have not considered the occlusion of the key point.

In conclusion, there is still much work to be done before our pipeline is useful in the vast majority of situations.
However, the Deep feature encodings have already shown promising results in terms of precision and accuracy for matching features in our applications.
Previous feature tracking and feature matching methods are not good enough for the applications we are interested in.

%
%

\section{Statements and declarations}

\subsection{Acknowledgments}
We thank Chien-Chih Wang for collecting the ballistocardiography data as a part of this Master's thesis work.

\subsection{Funding}
This research was supported by the Ministry of Science and Technology in Taiwan (MOST 108-2218-E-006-046, 109-2224-E-006-003, and 110-2222-E-006-010).

\subsection{Competing interests}
The authors have no financial or non-financial interests that are directly or indirectly related to the work submitted for publication.

\subsection{Ethics approval}
The use of the UTKface and remote ballistocardiography datasets was reviewed and approved by the National Cheng Kung University Human Research Ethics Review Committee under case 108-244.
The use of the Parkinson's Disease postural tremor test dataset was reviewed and approved by the Kaohsiung Medical University Chung-Ho Memorial Hospital Institutional Review Board under the number KMUHIRB-E(I)-20190173. 

\subsection{Data}
Our ethics approval does not allow us to make the validation data publicly available. 
The training of the autoencoder used in the Deep Feature Encodings can be reproduced using the publicly available UTKface dataset.

\subsection{Code availability}
Our method can be easily implemented based on the description in the article and pseudo code in the Appendix \ref{app:implementation}.

\subsection{Authors' contribution}
Author contribution using the CRediT taxonomy: 
Conceptualization: TN; Data curation: JC; Formal analysis: JC; Methodology: JC and TN; Investigation: JC; Software: JC; Verification: JC and TN; Visualization: JC and TN; Writing - original draft preparation: JC; Writing - review and editing: TN; Funding acquisition: TN; Project administration: TN; Resources: TN; Supervision: TN.

\appendix
\section{Appendix}
\subsection{Autoencoders for feature construction}
\begin{figure}[b!]
\centering
\includegraphics[width=\linewidth]{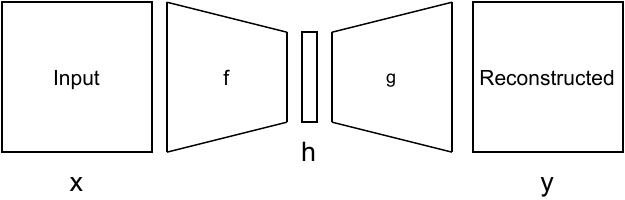}
\caption[Typical architecture of an autoencoder network.]{Typical architecture of an autoencoder network. The input $x$ is compressed into a representation $h$ through $f$ ($h=f(x)$) and then  a reconstruction $y$ is created from $h$ through $g$ ($y=g(h)=g(f(x))$).}\label{fig:autoencoder_structure}
\end{figure}

An autoencoder is a type of neural network that uses an unsupervised training strategy to learn the identity function, see Figure~\ref{fig:autoencoder_structure}.
As such, it can be trained by minimizing the discrepancy or distance between the original input data and its reconstruction.
The identity function may seem like a trivial and meaningless function to learn.
However, if we constrain the latent space to have fewer dimensions than our input, then we force the network to learn the most salient and distinctive features of our data.

Their use for dimensionality reduction was popularized in 2006 by Geoffrey Hinton \citep{Hinton2006}.
Similar to Principal Component Analysis, a key feature of autoencoders is their ability to encode the input data into latent state representations that provide insight into the samples of the data.
Strictly speaking, an autoencoder only requires a single encoder and a single decoder layer.
However, there are several advantages of adding depth to the architecture of these networks, $e.g.$ fewer computations required to represent some functions, fewer data needed to learn some functions \citep{Goodfellow2016}, better compression compared to their shallow or linear counterparts \citep{Hinton2006}.

Convolutional Auto Encoders (CAE), sometimes also referred to as Stacked Convolutional Auto Encoders, are a type of deep autoencoder that uses regional information to reconstruct inputs that are two- or three- dimensional.
In a standard autoencoder, the layers in the encoder and decoder are fully connected.
A CAE maintains the same structure as the standard autoencoder but replaces the fully-connected layers with convolutional layers when downsampling and transpose convolutional layers when upsampling \citep{Dumoulin2018}.
Similarly, as the standard autoencoder, the CAE also learns abstractions from the data even when the CAE is complete, $i.e.$ when the dimensionality of the latent space is as large as the input \citep{Manakov2019}.

Due to their ability to consider spatial information, CAEs have been especially useful when analyzing imaging data.
A deep CAE was used to cluster the embedded features of imaging data of several benchmark datasets \citep{Guo2017}.
The method, coined Deep Convolutional Embedded Clustering, can create features that were used to cluster the MNIST handwritten digits with an accuracy of $0.8897$.
Some other CAE applications include removing artifacts in images, denoising images, and coloring grayscale images.
Here we use a CAE to learn skin features.

\subsection{Statistics of the training dataset} \label{sec:statistics_training_data}

\begin{figure*}[tb!]
\centering
\includegraphics[width=\textwidth]{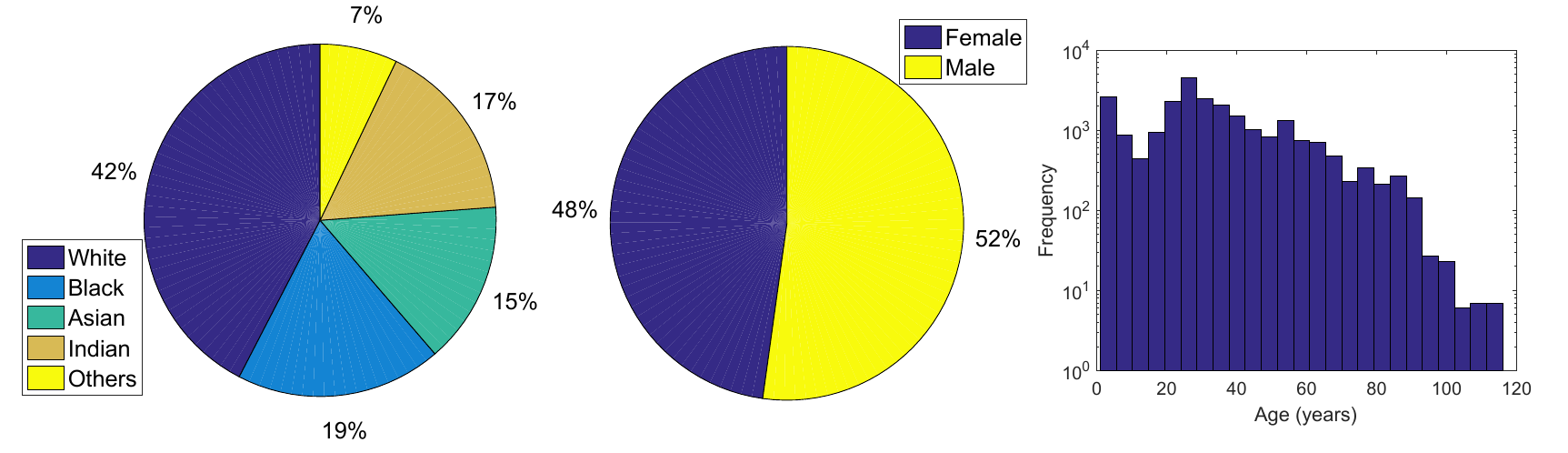}
\caption[Demographic information of the UTKface wild faces dataset.]{Demographic information of race (left), sex (middle), and age (right) of the UTKface wild faces dataset.}
\label{fig:UTKface_demographic_info}
\end{figure*}

\begin{figure*}[tb!]
\centering
\includegraphics[width=\textwidth]{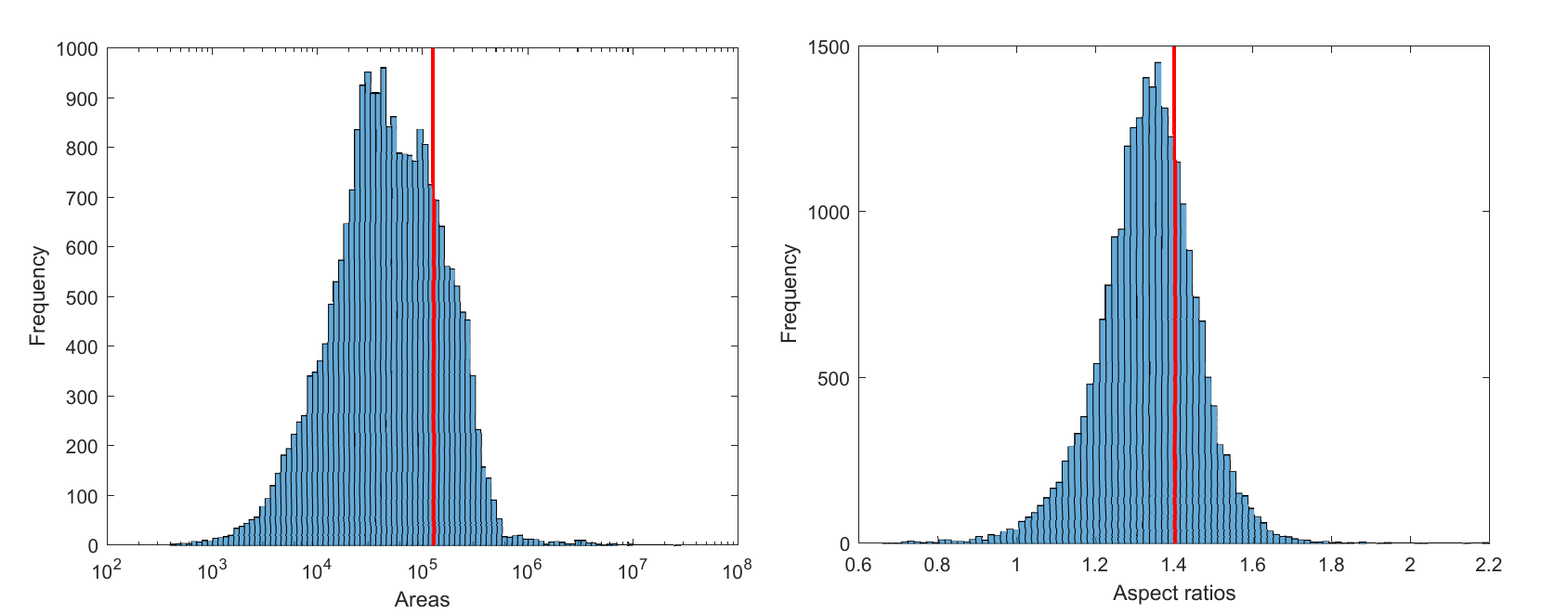}
\caption[Areas and aspect ratios of the cropped faces using ResNet-10 based model from the UTKwild faces dataset.]{Areas (left) and aspect ratios (right) of the cropped faces using the ResNet-10 model applied to the UTKface wild faces dataset. The red lines mark the area and aspect ratio of the resized validation of $1.26 \times 10^5$ pixels$^2$ and 1.4, respectively.}
\label{fig:Resnet10_cropped_faces_UTKwild_validation_marked}
\end{figure*}

UTKface dataset of faces in the wild contains $24,108$ faces gathered from the internet.
The face images are labeled by age, gender, and ethnicity; see Figure~\ref{fig:UTKface_demographic_info}.
The ground truth of age, gender, and race are estimated through the Deep EXpectation (DEX) algorithm \citep{Rothe2018} and verified by a human annotator. 
Each image was cropped to ensure the face dominates it. 
The face was in each image detected using the OpenCV-ResNet-10 based model.
The areas of the cropped faces are shown in Figure~\ref{fig:Resnet10_cropped_faces_UTKwild_validation_marked}.

\subsection{CIELAB color space}\label{sec:cielab}

All the images in the UTKface dataset are coded using the RGB color model.
This color model is the conventional model in which most images are created or displayed.
However, this color model is device-dependent, $i.e.$ given that the images were created using cameras with different white points (reference whites, target whites) the colors would be represented differently.
In addition, the RGB color model does not accurately represent how humans see color.
A large sum of squares differences in a pair of crops might not make them seem very different to our eyes, $e.g.$ we consider light green to be similar to yellow but not to red and we consider yellow to be very different from black but more similar to white.  
Since we want to train a device independent autoencoder to reconstruct image crops, we are interested in a reconstruction that is device independent and minimizes differences in accordance with how our visual system perceives them.  

\begin{figure*}[tb!]
 \centering
  \includegraphics[width=\textwidth]{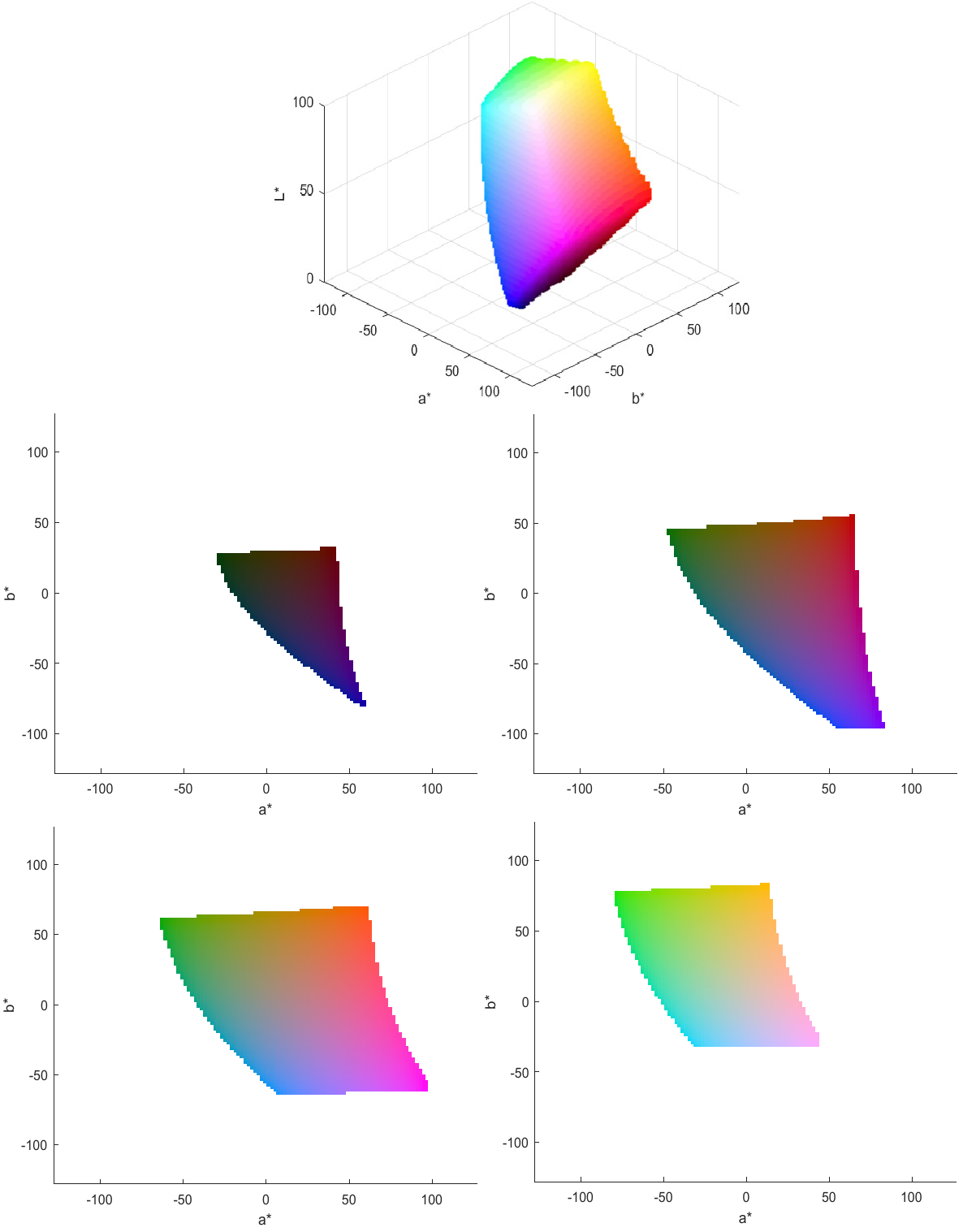}
\caption[The CIELAB color space.]{The CIELab color space. The color space has L* value ranges of $0$ to $100$, a* ranges from $-127$ to $127$, and b* from -$127$ to $127$. The CIELab color space contains the entire RGB gamut (top). The a*-b* plane is shown at $L=20$ (middle-left), $L=40$ (middle-right), $L=60$ (bottom-left) and $L=80$ (bottom-right).}
\label{fig:cielab_color_space}
\end{figure*}

In $1976$ the International Commission on Illumination (Commission internationale de l'\'eclairage in French, CIE) defined the  CIELAB color space (also known as CIE L*a*b*).
It is a device-independent color model that defines colors relative to the white point of the CIEXYZ space from which they were converted.
It represents colors in three axes:
\begin{enumerate}
	\item \textbf{L*} for lightness: Ranging from $0$ as black to $100$ for white.
	\item \textbf{a*} for green-red color channels: Ranging from $-127$ for green to $127$ to red.
	\item \textbf{b*} for blue-yellow color channels: Ranging from $-127$ for blue to $127$ to yellow.
\end{enumerate}
Here the asterisk (*) is pronounced star and is added in order to differentiate the $L^*$, $a^*$, and $b^*$ from Hunter's $L$, $a$, and $b$.
The color gamut of CIELAB includes the gamut of the RBG color model, ensuring no information is ignored at the conversion see, Figure~\ref{fig:cielab_color_space}.

The CIELAB color space aims to be perceptually uniform, $i.e.$ the nonlinear relations for $L*$, $a*$, and $b*$ are intended to mimic the nonlinear response of the eyes when looking at different colors.
As such, the Euclidean distance between two colors is proportional to the color difference perceived by the standard (colorimetric) observer \citep{Colantoni2016}. 
In addition, unlike many color spaces that use cylindrical representations of the hues of colors (HSV, HSL, or HSB), the color representations are non-periodic.
This property is especially good for training machine learning models that minimize the sum of squared residuals as the same hue of color can have drastically different values.
In HSV the red hue ranges from roughly 345 to 15 with a transition to 0 at 360 in the H channel.

CIELAB is copyright and license-free.
It is also mathematically fully defined.
A conversion of a floating point image (with intensity range of $0$ to $1$) in the RGB color space, with a reference white point given by the CIE standard illuminant D65, $i.e.$ common sRGB, to CIELAB color space is given by:
\begin{gather*}
\begin{bmatrix}
X\\
Y\\
Z
\end{bmatrix}=
\begin{bmatrix}
0.412453 & 0.357580 & 0.180423\\
0.212671 & 0.715160 & 0.072169\\
0.019334 & 0.119193 & 0.950227
\end{bmatrix}
\begin{bmatrix}
R\\
G\\
B
\end{bmatrix}
\\
X=X/X_n, \quad \text{where}\quad X_n=0.950456\\
Z=Z/Z_n, \quad \text{where}\quad Z_n=1.088754\\
L=\begin{cases}
116*Y^{1/3}-16 & \text{for}\quad Y > 0.008856\\
903.3*Y & \text{for}\quad Y \leq 0.008856
\end{cases}\\
a=500(f(X)-f(Y))\\
b=200(f(Y)-f(Z)),
\end{gather*} 
where
\begin{gather*}
f(t)=\begin{cases}
t^{1/3} & \text{for}\quad t > 0.008856\\
7.787t+16/116 & \text{for}\quad t \leq 0.008856.
\end{cases}
\end{gather*}

The aforementioned conversion is the one employed in OpenCV under the assumption that the image uses the sRGB profile.
The conversion takes as inputs floating point RGB (ranging $0$ to $1$) and outputs CIELAB values with ranges in \textbf{L*} from $0$ to $100$, \textbf{a*} from $-127$ to $127$, and \textbf{b*} from $-127$ to $127$.
It is typical when a variable has a defined upper and lower bound, to normalize it to values between $0$ and $1$.
This will also allow us to directly compare the outputs of the prediction layer in the convolutional autoencoder which has a Sigmoidal function to constrain the values within this range.
To normalize each of these channels to the $0$ to $1$ range we have implemented a channel-wise min-max normalization technique.
\begin{align*}
x'=\frac{x-min(x)}{max(x)-min(x)}.
\end{align*}
Here $min(x)$ and $max(x)$ are the absolute minimum and maximum of each respective channel in the CIELAB color space.

Since the images in the UTKface dataset were obtained from the internet, it is not clear if they are encoded in standard RGB using white point D65 or not. 
However, we assumed all of these to be in standard RGB as this is considered to be the default color space for the internet \citep{anderson1996proposal}.
The images collected for the remote ballistocardiography dataset were collected using a Panasonic GX85 camera with a default standard RGB color space.
The images from the Parkinson's disease postural tremor test were collected using the main camera of a Samsung S7 phone which also encodes them in standard RGB.

\subsection{Normality tests for relabeling errors in x and y directions}
\label{sec:normality_of_errors}
The errors in both x and y directions were calculated by subtracting the mean of 6 relabeling attempts for each of the 15 images and concatenating them
into a vector with 90 elements.
The labelling of the image was done at pixel level by a human operator repeatedly clicking on the screen to label the feature.
Since the range of the relabelling errors is small, this yields data that is quantized into $n_{quant}$ values which are shown in Table ~\ref{tab:relabeling_errors_n_quant}.

\begin{table}[tb!]
\centering
\caption[Number of states that the relabelling errors in both x and y directions are quantized into.]{Number of states that the relabelling errors in both x and y directions are quantized into ($n_{quant}$).}
\label{tab:relabeling_errors_n_quant}
\begin{tabular}{c c c c c }
\hline
                 & \multicolumn{2}{c}{$n_{quant}$} & \multicolumn{2}{c}{Range} \\
                 & $\delta_x$           &$\delta_y$            & $\delta_x$            & $\delta_y$         \\
\hline
Static face mole & 19            & 21           & 4.5         & 5.5         \\
Static nose tip  & 25            & 30           & 5.5         & 5.7         \\
Bike face mole   & 31            & 26           & 6.0           & 5.5         \\
Bike nose tip    & 31            & 28           & 6.3         & 6.0           \\
PD hand mole     & 26            & 22           & 6.8         & 4.5     \\   
\hline
\end{tabular}
\end{table}

\begin{figure*}[tb!]
 \centering
  \includegraphics[width=0.7\linewidth]{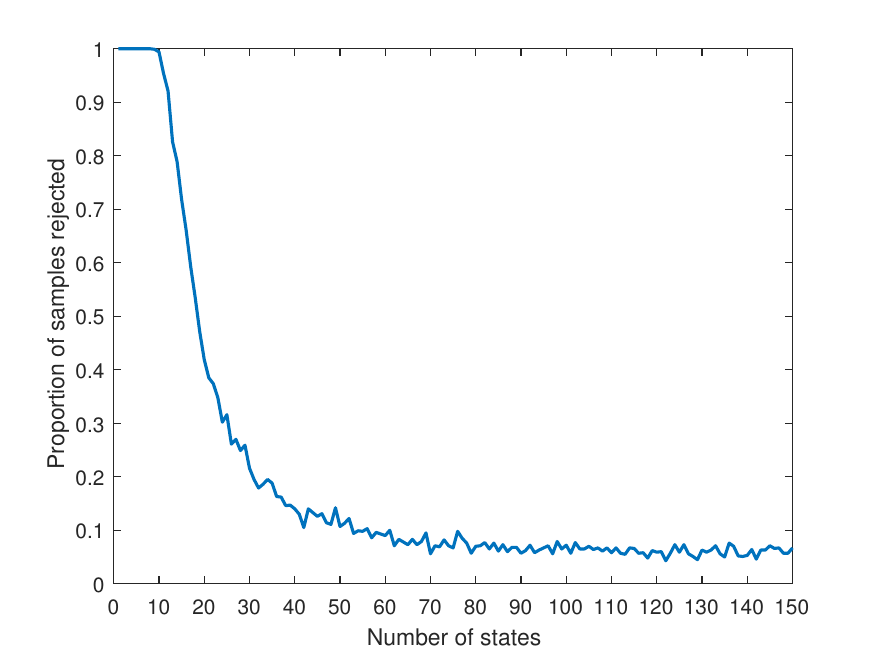}
\caption[Proportion of samples for which the null hypothesis of the data coming from a normal distribution is rejected using the Lilliefors test at significance level 0.05 as a function of the number of states that the values were quantized into. All samples were drawn from a standard normal distribution.]{Proportion of samples for which the null hypothesis of the data coming from a normal distribution is rejected using the Lilliefors test at significance level 0.05 as a function of the number of states that the values were quantized into. All samples were drawn from a standard normal distribution.}
\label{fig:n_quant}
\end{figure*}

Quantization of a normal distribution makes the distribution more similar to a binomial distribution \citep{tarongi2010normality}.
Therefore, when applying normality tests, $e.g.$ Lilliefors tests, the data looks non-normal.
To avoid this, the number of quantization levels must increase as the sample size increases.
The results of a Monte Carlo simulation of sampling 1000 standard normal distributions of 90 values each and applying the Lilliefors test for normality is shown in Figures ~\ref{fig:n_quant}.
Only 90 values were drawn to match the number of relabeling errors.
These were binned into 1 to 150 bins and replaced by the bin center after min max normalisation to 0 to 1. 
For the number of quantization states seen in our data (19-31) the Lilliefors test is misleading.

\begin{figure*}[tb!]
 \centering
  \includegraphics[width=\textwidth]{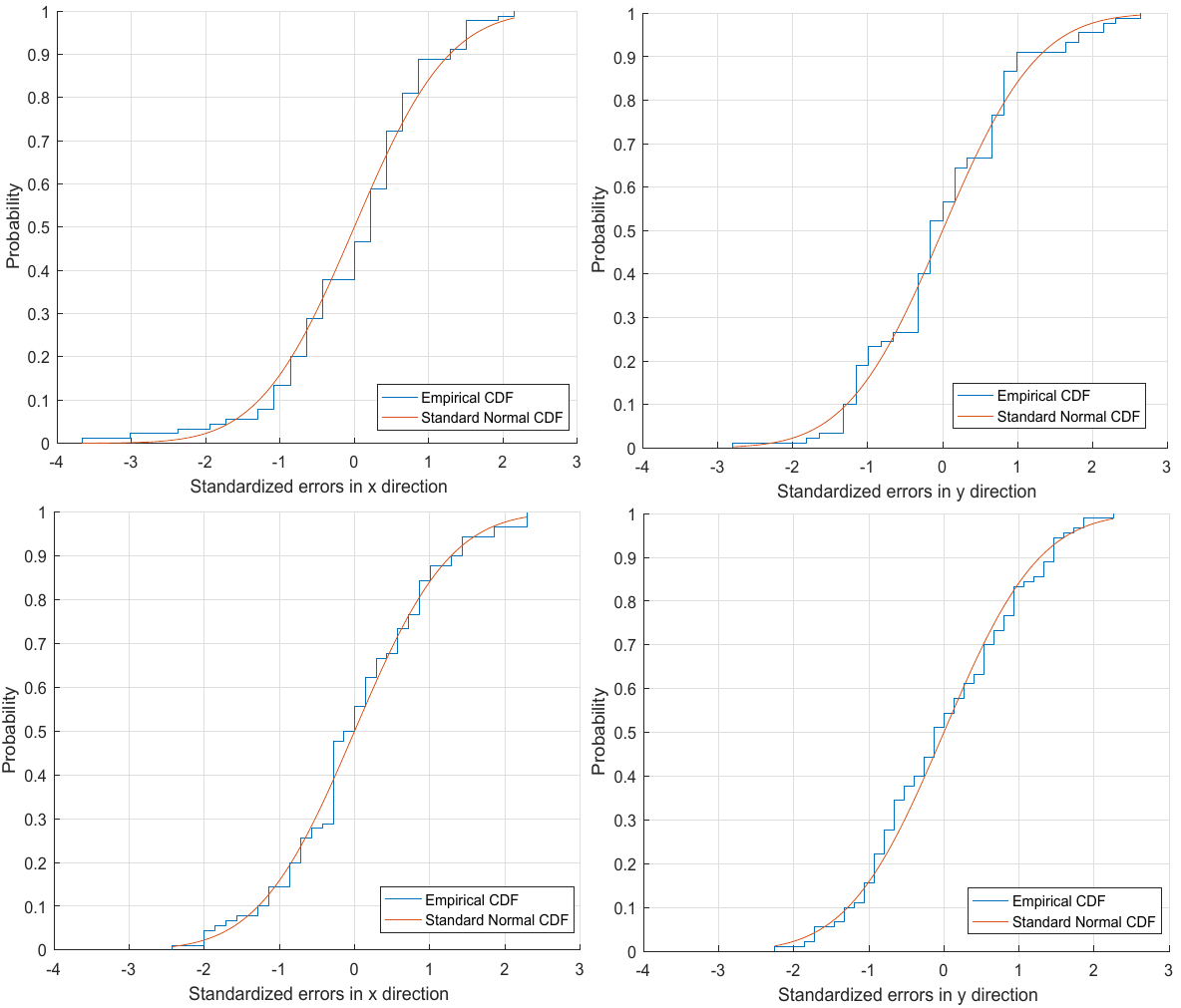}
\caption[CDF of the standardized errors from the videos of static conditions in x direction for face mole (top-left), errors in y direction for face mole (top-right), errors in x directions for nose tip (bottom-left), and errors in y direction for nose tip (bottom-right). The errors were standardized by dividing them by their respective standard deviation. The red line marks the theoretical CDF of a standard normal distribution.]{CDF of the standardized errors from the videos of static conditions in x direction for face mole (top-left), errors in y direction for face mole (top-right), errors in x directions for nose tip (bottom-left), and errors in y direction for nose tip (bottom-right). The quantized errors were standardized by dividing them by their respective standard deviation. The red line marks the theoretical CDF of a standard normal distribution.}
\label{fig:empirical_cdf_dx_dy_static}
\end{figure*}

\begin{figure*}[tb!]
 \centering
  \includegraphics[width=\textwidth]{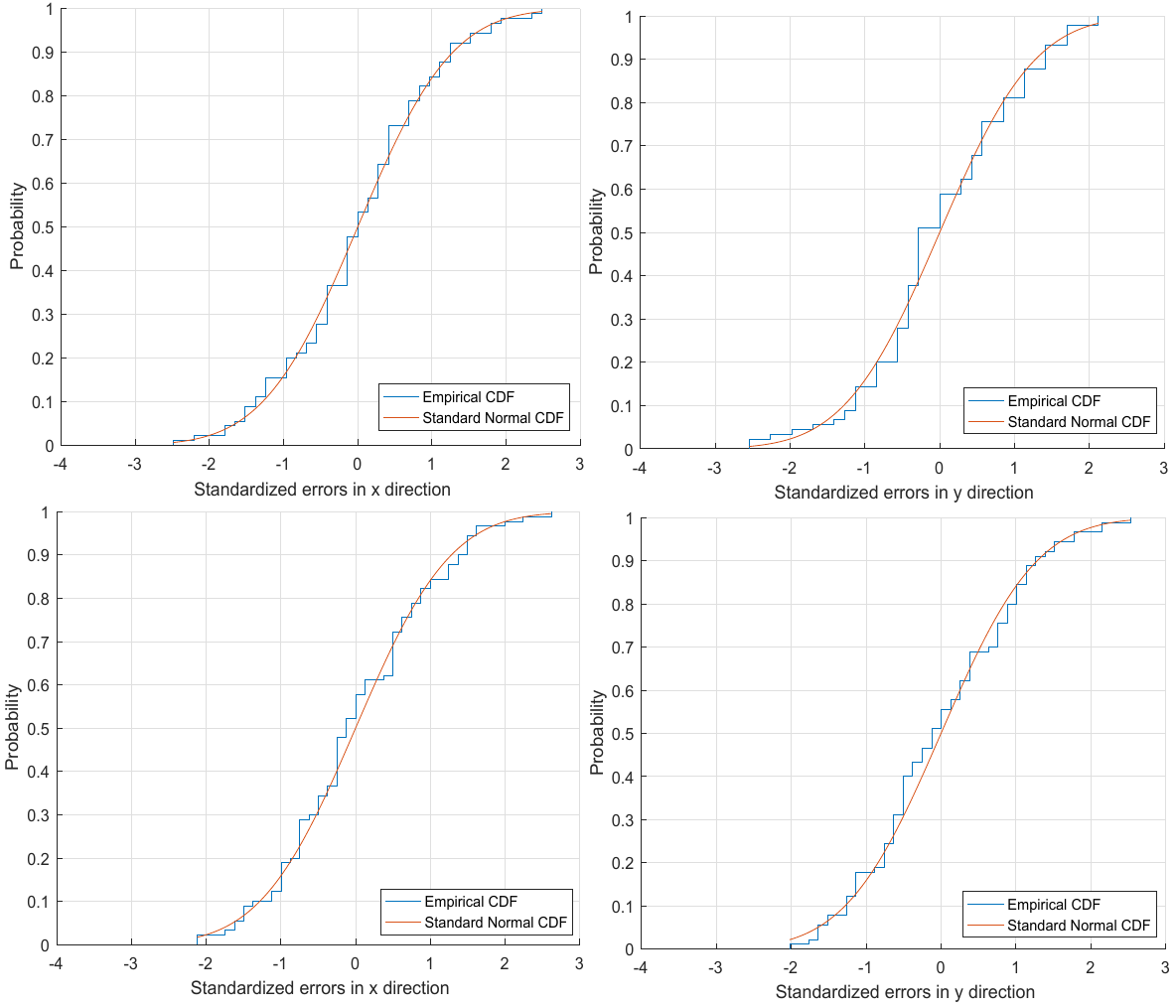}
\caption[CDF of the standardized errors from the videos of bike conditions in x direction for face mole (top-left), errors in y direction for face mole (top-right), errors in x directions for nose tip (bottom-left), and errors in y direction for nose tip (bottom-right). The errors were standardized by dividing them by their respective standard deviation. The red line marks the theoretical CDF of a standard normal distribution.]{CDF of the standardized errors from the videos of bike conditions in x direction for face mole (top-left), errors in y direction for face mole (top-right), errors in x directions for nose tip (bottom-left), and errors in y direction for nose tip (bottom-right). The quantized errors were standardized by dividing them by their respective standard deviation. The red line marks the theoretical CDF of a standard normal distribution.}
\label{fig:empirical_cdf_dx_dy_bike}
\end{figure*}

\begin{figure*}[tb!]
 \centering
  \includegraphics[width=\textwidth]{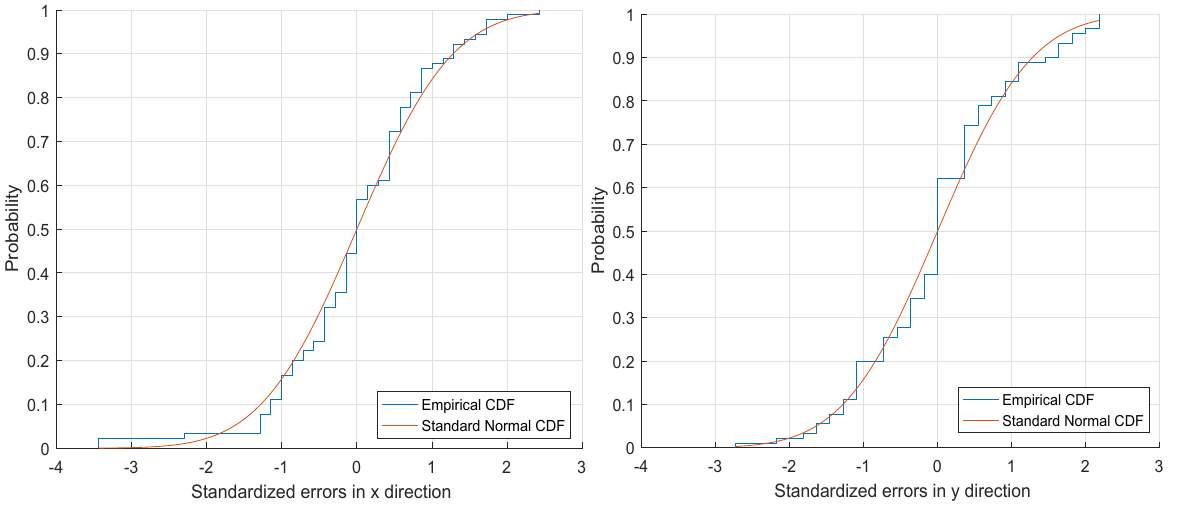}
\caption[CDF of the standardized errors from the videos of PD condition in x direction for hand mole (left) and errors in y direction for hand mole (right). The errors were standardized by dividing them by their respective standard deviation. The red line marks the theoretical CDF of a standard normal distribution.]{CDF of the standardized errors from the videos of PD condition in x direction for hand mole (left) and errors in y direction for hand mole (right). The quantized errors were standardized by dividing them by their respective standard deviation. The red line marks the theoretical CDF of a standard normal distribution.}
\label{fig:empirical_cdf_dx_dy_pd}
\end{figure*}

However, quantization only affects the CDF of a distribution by making it look less smooth and more like a staircase. 
We standardized the errors in x and y directions by their respective standard deviations and compared them against a theoretical standard normal distribution ($\mu=0$ and $\sigma=1$) in  Figures~\ref{fig:empirical_cdf_dx_dy_static},~\ref{fig:empirical_cdf_dx_dy_bike},and~\ref{fig:empirical_cdf_dx_dy_pd}.
Using this method, we can determine that the manual labelling errors are normally distributed.

\begin{figure*}[tb!]
\centering
\includegraphics[width=0.7\linewidth]{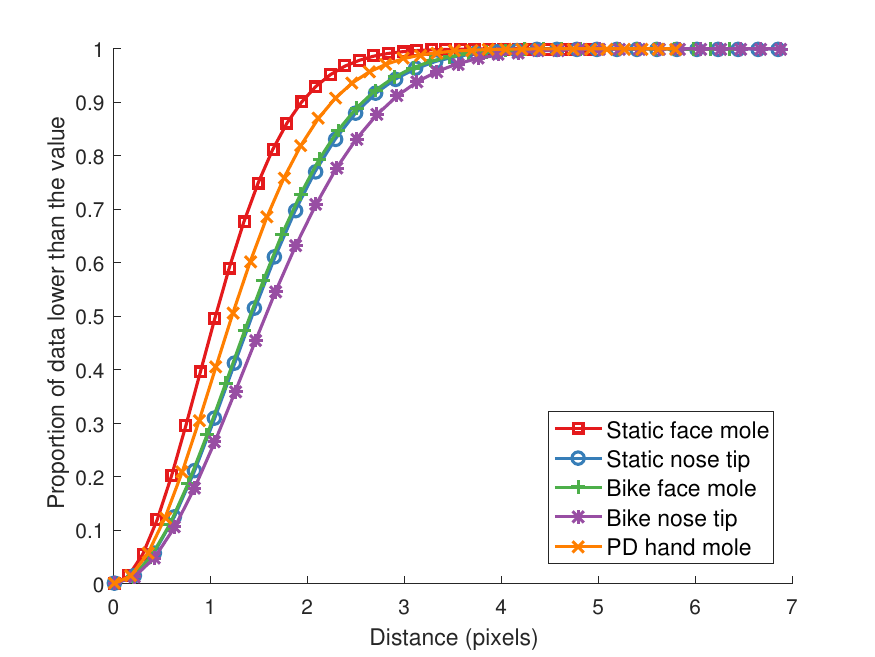}
\caption[Simulated cumulative distribution functions for all the body part and motion conditions.]{Simulated cumulative distribution functions for each of the body part and motion conditions.}\label{fig:chi_square_cdf_all_conditions}
\end{figure*}

Given that the errors in x and y directions are normal, the sum of their squares is Chi square distributed
\begin{align*}
\delta x^2 +\delta y^2 \sim \chi^2.\
\end{align*}
The Cumulative Density Function (CDF) of each of the five conditions, is shown in Figure~\ref{fig:chi_square_cdf_all_conditions}, generated by the $normal$ function of the $random$ number routines of $numpy$ version 1.19.3 using Python 3.6.
We use the CDF of the Chi-square distribution to determine the likelihood that an error stems from the manual labelling and establish error thresholds at the common $0.01$, $0.05$, and $0.5$ significance levels ($\alpha$).
The human errors fall below these thresholds with probability $0.99$, $0.95$, and $0.5$, respectively.
The respective thresholds for all the three confidence levels can be seen in Table~\ref{tab:error_thresh_cl}.

\begin{table}
\centering
\caption[Error thresholds (in pixels) for the three significance levels.]{Error thresholds (in pixels) for the three significance levels.}\label{tab:error_thresh_cl}
\begin{tabular}{c c c c}
\hline
 &$\alpha=0.50$ &$\alpha=0.05$&$\alpha=0.01$\\
\hline
Static face mole& 1.049& 2.222& 2.789\\
Static nose tip&  1.425&  2.968&  3.683\\
Bike face mole& 1.404& 2.922& 3.623\\
Bike nose tip& 1.564& 3.246& 4.028\\
PD hand mole& 1.219& 2.571& 3.225\\
\hline
\end{tabular}
\end{table}

The test statistics described in Chi square analysis are described in Table~\ref{tab:test_statistics} with the respective 99\% confidence threshold to reject the null hypothesis of the errors coming from human labelling.

 \begin{table*}[tb!]
 \centering
\caption[Test statistics for determining whether the errors are due to human errors.]{Test statistics for rejecting the null hypothesis of whether the errors are due to human errors. Statistics that fall below the threshold are in bold.}
\label{tab:test_statistics}
\begin{tabular}{C{2.5cm} C{2cm} C{2cm} C{2cm} C{2cm} C{2cm} }
\hline
&SIFT&SURF& LK& DFE& Threshold \\
\hline
Static face mole& 6.8887e+05 & 5.0428e+06 & \textbf{507.8146}& \textbf{467.2694}&597.9500 \\
Static nose tip& 1.2527e+03&  7.4170e+05 & 1.0507e+03 & 1.3334e+03& 597.9500\\
Bike face mole& 2.3114e+05 &  9.2964e+05 & 6.9858e+04& \textbf{123.9379} & 401.4090\\
Bike nose tip& 7.0651e+05 &  1.4756e+06 & 4.8984e+04 & 1.3331e+03 & 401.4090\\
PD hand mole &1.0317e+05  & 3.5759e+05  & 9.2540e+04  & \textbf{21.4238}& 109.9581\\
 \hline
 \end{tabular}
 \end{table*}

\subsection{Calculation of SIFT descriptors}
\begin{figure*}[tb!]
\centering
\includegraphics[width=0.85\textwidth]{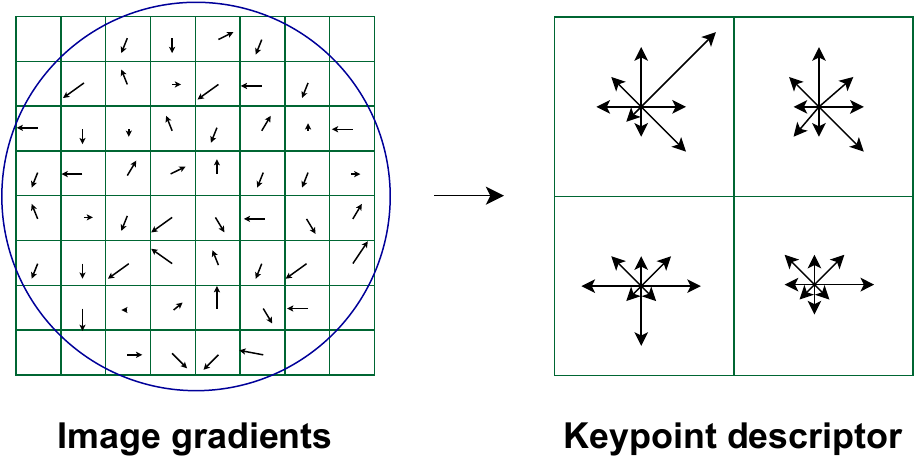}
\caption[Calculation of the SIFT descriptor through the histogram of the magnitudes and gradients of neighboring pixels.]{Calculation of the SIFT descriptor through the histogram of the magnitudes and gradients of neighboring pixels. The neighboring pixels are smoothed using a Gaussian filter (blue circle). The magnitudes and gradients are then accumulated and represented as histograms for each of the sub-regions. The length of each arrow in the sub-sampled space corresponds to the sum of the gradient magnitudes near that direction within that region. This figure shows a $2\times 2$ descriptor array computed from an $8\times 8$ set of samples, whereas the original publication \citep{Lowe2004} use a $4\times 4$ descriptor array computed from a $16\times 16$ sample array.}\label{fig:sift_descriptor}
\end{figure*}

Figure~\ref{fig:sift_descriptor} shows how SIFT descriptor are calculated from histograms of the magnitudes of the gradients of neighboring pixels.
The magnitude of gradients around a $16 \times 16$ area around the keypoint are weighted using a Gaussian kernel.
Histograms of the weighted magnitudes of each $4 \times 4$ area in the $16 \times 16$ area are created; resulting in $4 \times 4$ histograms.
The bins in each 8-binned histogram represent the orientation of the gradients and the height of the bars represent the magnitude of that orientation. 
These histograms are concatenated into a vector of $4 \times 4 \times 8$ dimensions which represents the keypoint descriptor.

\subsection{Approximations of Laplacian of Gaussians using box filters}

\begin{figure*}[tb!]
\centering
\includegraphics[width=\textwidth]{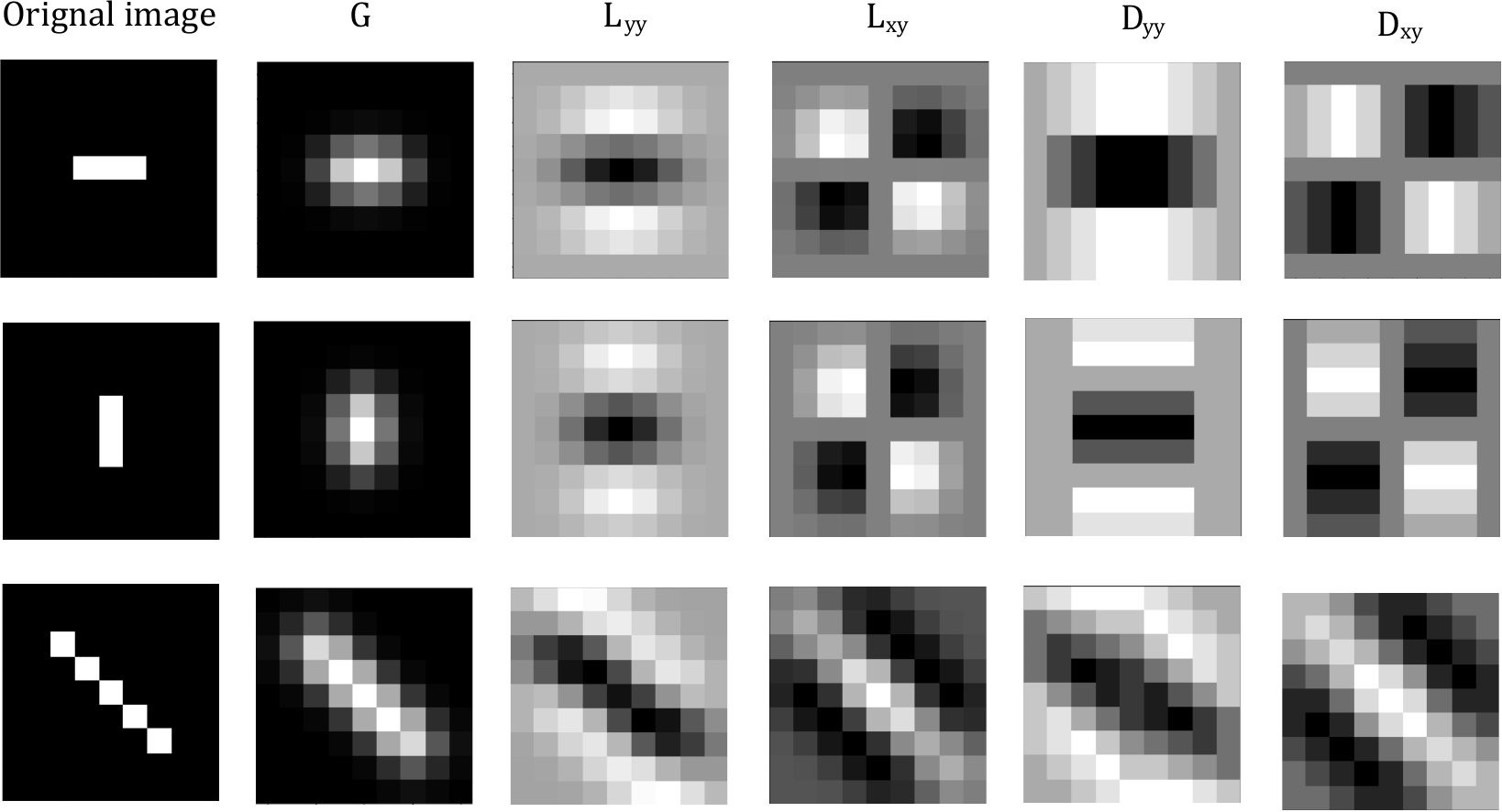}
\caption[Examples of approximation of second derivatives using box filters.]{Examples of approximation of second derivatives using box filters. $L_{yy}$ is the second derivatives in $y$ direction of the Gaussian convolved image $G$, $L_{xy}$ is the second derivative first in $x$ direction and then in $y$ direction of the Gaussian convolved image $G$, $D_{yy}$ is the approximation of $L_{yy}$ using a box vertical filter, and $D_{xy}$ is the approximation of $L_{xy}$ using a box diagonal filter in.}\label{fig:log_approximations_box}
\end{figure*}

SURF approximates the Laplacian of Gaussian using box filters.
Figure~\ref{fig:log_approximations_box} shows several examples of these approximations.
Box filters is a cruder approximation of Gaussian second derivatives than the Difference of Gaussians used in SIFT \citep{Lowe2004}.
However, the calculation of these box filter approximations is computationally cheap with the use of integral images, independently of their size.

\subsection{Implementation of SIFT, SURF, and LK}\label{app:implementation}

We used SIFT, SURF, and LK as implemented in version 3.4.2.17 of OpenCV.
For SIFT and SURF we used the \texttt{compute} function to calculate the feature descriptors of the given coordinate of the skin feature of interest in the reference frame.
For LK we used the \texttt{calcOpticalFlowPyrLK} function to predict the location of the skin feature in the current frame.
Details of our feature matching between the reference point and its closest match in the current frame are provided in Algorithm~\ref{algo:siftsurf} for SIFT and SURF and Algorithm~\ref{algo:dfe} for DFE.
A detailed explanation of the methods implemented in the functions \texttt{calculate\_quadratic\_surface\_params()} and \texttt{get\_subpixel\_level\_pred()} is given in section Subpixel level prediction in the manuscript.
Location of the skin feature of interest in the current frame using LK is described in Algorithm~\ref{algo:lk}. 
Note that we use Python notation for clarity of certain steps and assume the usage of OpenCV and Python functions.
The default parameters specific to each method are described in their respective sections in the manuscript.

\begin{algorithm*}[tb!]
\caption{Feature matching using SIFT and SURF to locate the skin feature. }\label{algo:siftsurf}
\begin{algorithmic}[1]
\Require ref\textsubscript{point}, img\textsubscript{reference}, img\textsubscript{current}
\Ensure pred \Comment{Location of best match}
\State size\textsubscript{window} = (31,31), stride = 1 \Comment{Initialize parameters}
\State positions = [ ], curvatures = [ ], candidates = [ ]
\State descriptor = cv2.xfeatures2d.method\_create() \Comment{Create SIFT/SURF object by replacing ``method" in the function name by SIFT/SURF}
\State ref\textsubscript{descriptor} = descriptor.compute(img\textsubscript{{reference}}, ref\textsubscript{{point}}) \Comment{Get reference descriptor}  
\For{(i \textless int((img\textsubscript{current}.shape[0]-size\textsubscript{{window}}[0])/stride)} \Comment{Generate list of possible positions}  
	\For{(j \textless int((img\textsubscript{current}.shape[1]-size\textsubscript{{window}}[1])/stride)}
		\State x = int(j+ceil(size\textsubscript{window}[1]/2))
		\State y = int(i+ceil(size\textsubscript{window}[0]/2))
		\State positions.append([y,x]) \Comment{Get all possible positions where a $31 \times 31$ window can be centered}
	\EndFor
\EndFor
\State descriptors = descriptor.compute(img\textsubscript{current},positions)
\State ssrs = sum((descriptors-ref\textsubscript{descriptor})\textsuperscript{2},axis=1) \Comment{Row-wise calculation of SSRs}
\For{error\textsubscript{unique} in unique(sort(ssrs))} \Comment{Filter out points by their curvature}
	\For{candidate in  where(ssrs == error\textsubscript{unique})} 
		\State c = calculate\_quadratic\_surface\_params(positions[candidate], img\textsubscript{current})  \Comment{Get the 6 parameters of the surface  fitted to the $3\times 3$ neighborhood around the candidate point. See section Subixel level prediction.}
		\State D = 4c[5]c[4]-c[3]\textsuperscript{2}
		\If{D\textgreater 0 and c[4] \textgreater 0} \Comment{Check if the fitted surface has a local minimum}
			\State{curvatures.append(D)}
			\State{candidates.append(candidate)}
		\EndIf
	\EndFor
	\If{bool(curvatures)}
		\State break \Comment{Stop if a point fulfills the local minimum criteria}
	\EndIf
\EndFor
\State pred\textsubscript{pixel} = positions[candidates[argmax(curvatures)]] \Comment{Predicted position with pixel accuracy}
\State pred = get\_subpixel\_level\_pred(pred\textsubscript{pixel},img\textsubscript{{current}}) \Comment{Predicted position with subpixel accuracy.  See section Subixel level prediction.}
\end{algorithmic}
\end{algorithm*}

\begin{algorithm*}[tb!]
\caption{Feature matching using DFE to locate the skin feature. }\label{algo:dfe}
\begin{algorithmic}[1]
\Require ref\textsubscript{point}, img\textsubscript{reference}, img\textsubscript{current}, encoder
\Ensure pred \Comment{Location of best match}
\State size\textsubscript{window} = (31,31), stride = 1 \Comment{Initialize parameters}
\State ref\textsubscript{crop} = img\textsubscript{reference}[ref\textsubscript{point}[0]-size\textsubscript{window}[0]:ref\textsubscript{point}[0]+size\textsubscript{window}[0], ref\textsubscript{point}[1]-size\textsubscript{window}[1]:ref\textsubscript{point}[1]+size\textsubscript{window}[1], :] \Comment{Create reference crop}
\State ref\textsubscript{descriptor} = encoder.predict(img\textsubscript{reference},ref\textsubscript{point}) \Comment{Get reference descriptor}  
\State positions = [ ], crops = [ ], curvatures = [ ], candidates = [ ]
\For{(i\textless int((img\textsubscript{current}.shape[0]-size\textsubscript{window}[0])/stride)}
	\For{(j\textless int((img\textsubscript{current}.shape[1]-size\textsubscript{window}[1])/stride)}
		\State x = int(j+ceil(size\textsubscript{window}[1]/2))
		\State y = int(i+ceil(size\textsubscript{window}[0]/2))
		\State crops.append(img\textsubscript{current}[i:i+size\textsubscript{window}[0], j:j+size\textsubscript{window}[1], :])
		\State positions.append([y,x]) \Comment{Generate list of possible positions}
	\EndFor
\EndFor
\State descriptors = encoder.predict(crops)
\State ssrs = sum((descriptors-ref\textsubscript{descriptor})\textsuperscript{2},axis=1) \Comment{Row-wise calculation of SSRs}\\
\For{error\textsubscript{unique} in unique(sort(ssrs))} \Comment{Filter out points by their curvature}
	\For{candidate in  where(ssrs == error\textsubscript{unique})} 
		\State c = calculate\_quadratic\_surface\_params(positions[candidate], img\textsubscript{current})  \Comment{Get the 6 parameters of the fitted surface around the $3\times 3$ neighborhood around the candidate point. See section Subixel level prediction.}
		\State D = 4c[5]c[4]-c[3]\textsuperscript{2}
		\If{D\textgreater 0 and c[4] \textgreater 0} \Comment{Check if the fitted surface has a local minimum}
			\State{curvatures.append(D)}
			\State{candidates.append(candidate)}
		\EndIf
	\EndFor
	\If{bool(curvatures)}
		\State break \Comment{Stop if a point fulfills the local minimum criteria}
	\EndIf
\EndFor
\State pred\textsubscript{pixel} = positions[candidates[argmax(curvatures)]] \Comment{Predicted position with pixel accuracy}
\State pred = get\_subpixel\_level\_pred(pred\textsubscript{pixel},img\textsubscript{{current}}) \Comment{Predicted position with subpixel accuracy.  See section Subixel level prediction.}
\end{algorithmic}
\end{algorithm*}

\begin{algorithm*}[tb!]
\caption{Location of the skin feature in the current frame using LK.}\label{algo:lk}
\begin{algorithmic}[1]
\Require ref\textsubscript{point}, img\textsubscript{reference}, img\textsubscript{current}

\Ensure pred \Comment{Location of best match}
\State lucas\_kanade\_params = dict(winSize  = (10,10), maxLevel = 4, criteria = (cv2.TERM\_CRITERIA\_EPS $\mid$ cv2.TERM\_CRITERIA\_COUNT, 10, 0.03))  \Comment{Required parameters for LK.}
\State pred = cv2.calcOpticalFlowPyrLK(img\textsubscript{reference}, img\textsubscript{current}, ref\textsubscript{point}, None, **lucas\_kanade\_params)

\end{algorithmic}
\end{algorithm*}

\subsection{Weighted errors}

We also weigh the errors $e$, $i.e.$ distances between the prediction and the ground truth, of the validation set by their probability based on the CDF of the simulated Chi square distribution. 
The weighted error $e_{weighted}$ is defined as 
\begin{align*}
e_{weighted}=\frac{e}{1-F_{\chi^2}(e)},
\end{align*}
where $F_{\chi^2}(e)$ is the CDF of the Chi square distribution evaluated at $e$.
This weighting scheme penalizes errors that are inherent of the methods and not likely due to human mislabeling.
The weighted errors for each condition can be seen in Table~\ref{tab:weighted_errors}.

\begin{table*}[tb!]
\centering
\caption[Weighted mean errors.]{Weighted mean errors -- the original error weighted by the reciprocal of the significance level of the simulated Chi square distribution. The best results are highlighted in bold.}
\label{tab:weighted_errors}
\begin{tabular}{c c c c c c }
\hline
 &\multirow{2}{*}{SIFT} & SIFT with& \multirow{2}{*}{SURF} &\multirow{2}{*}{LK}& \multirow{2}{*}{DFE}\\
 & & threshold & & & \\
\hline
Static face mole& 1.63$\times 10^7$ & 1.87$\times 10^8$& 1.72$\times 10^8$ & 1.02$\times 10^2$ & \textbf{6.14$\times 10^1$}\\
Static nose tip& 6.05$\times 10^4$ & 3.14$\times 10^8$& 1.63$\times 10^8$ & \textbf{3.75$\times 10^4$}& 6.35$\times 10^4$\\
Bike face mole& 1.56$\times 10^7$ & 1.76$\times 10^8$ & 1.36$\times 10^8$ & 3.83$\times 10^6$& \textbf{2.34$\times 10^1$} \\
Bike nose tip& 6.29$\times 10^7$ & 3.45$\times 10^8$ & 2.57$\times 10^8$ & 9.40$\times 10^6$ & \textbf{1.11$\times 10^5$}\\
\hline
\end{tabular}
\end{table*}

\subsection{Maximum errors}
The maximum errors for feature matching, $i.e.$ tracking the original reference feature in the first frame, are found in Table~\ref{tab:max_errors_ofeat}.
DFE achieves the lowest maximum error for all conditions except for tracking the face mole under static conditions.
For this condition, the method with the lowest maximum error is LK followed by SIFT.
Table~\ref{tab:max_errors_pfeat} shows the maximum errors for tracking the prediction of the previous frame.
For this tracking scheme, DFE was the method with the lowest maximum error for all conditions.
This is because DFE was the only method that never diverged.

\begin{table*}[tb!]
\centering
\caption[Max errors.]{Maximum errors for feature matching. The best results are highlighted in bold.}
\label{tab:max_errors_ofeat}
\begin{tabular}{c c c c c c }
\hline
 &\multirow{2}{*}{SIFT} & SIFT with& \multirow{2}{*}{SURF} &\multirow{2}{*}{LK}& \multirow{2}{*}{DFE}\\
 & & threshold& & & \\
\hline
Static face mole& 210.2243 & 516.1395 & 265.0907  &  2.2168  &  \textbf{2.1347}\\
Static nose tip &5.6971 &  516.1395 & 139.7204  &  \textbf{5.1310}  &  5.7030\\
Bike face mole& 218.5698 & 516.1395 & 207.2922 & 254.7781  &  \textbf{2.0818} \\
Bike nose tip& 79.6503 & 516.1395 & 206.8976  & 76.3600  &  \textbf{6.4795}\\
PD hand mole& 219.8465 & 516.1395 & 200.1399 & 120.3210  &  \textbf{1.5241}\\
\hline
\end{tabular}
\end{table*}

\begin{table*}[tb!]
\centering
\small
\caption[Max errors.]{Maximum errors for tracking the prediction from the previous frame. The best results are highlighted in bold.}
\label{tab:max_errors_pfeat}
\begin{tabular}{c c c c c c }
\hline
 &\multirow{2}{*}{SIFT} & SIFT with& \multirow{2}{*}{SURF} &LK& DFE\\
 & & threshold& & & \\
\hline
Static face mole& 266.1209 & 516.1395 & 277.9227 & 263.1993  &  \textbf{5.4279}\\
Static nose tip &250.1239 & 516.1395 & 265.6182 & 287.4956  &  \textbf{9.5427}\\
Bike face mole& 282.1509 & 516.1395 & 319.0986 & 346.0556  &  \textbf{4.8943} \\
Bike nose tip& 304.4242 & 516.1395 & 325.6847 & 293.7723  &  \textbf{5.1006}\\
PD hand mole& 231.5862 & 516.1395 & 276.3275 & 354.3394  &  \textbf{3.2052}\\
\hline
\end{tabular}
\end{table*}

\subsection{Tracking errors}
A sudden increase in the tracking error shown in Figure 7 occurs for frames 43 to 46 for DFE for tracking the nose tip under bike conditions.
This is the only condition for which we can see this sudden increase in the tracking error for DFE.
A sudden increase in the error of LK, another method that performed well in most circumstances, also appears in the same frames.
A slightly smaller increase is seen in Figure 8 for the same frames for DFE.
This suggests these frames are of interest as they are the first frames for which we can see a clear drop in performance. 
Thus we decided to investigate this drop.

\begin{figure*}[tb!]
 \centering
  \includegraphics[width=\textwidth]{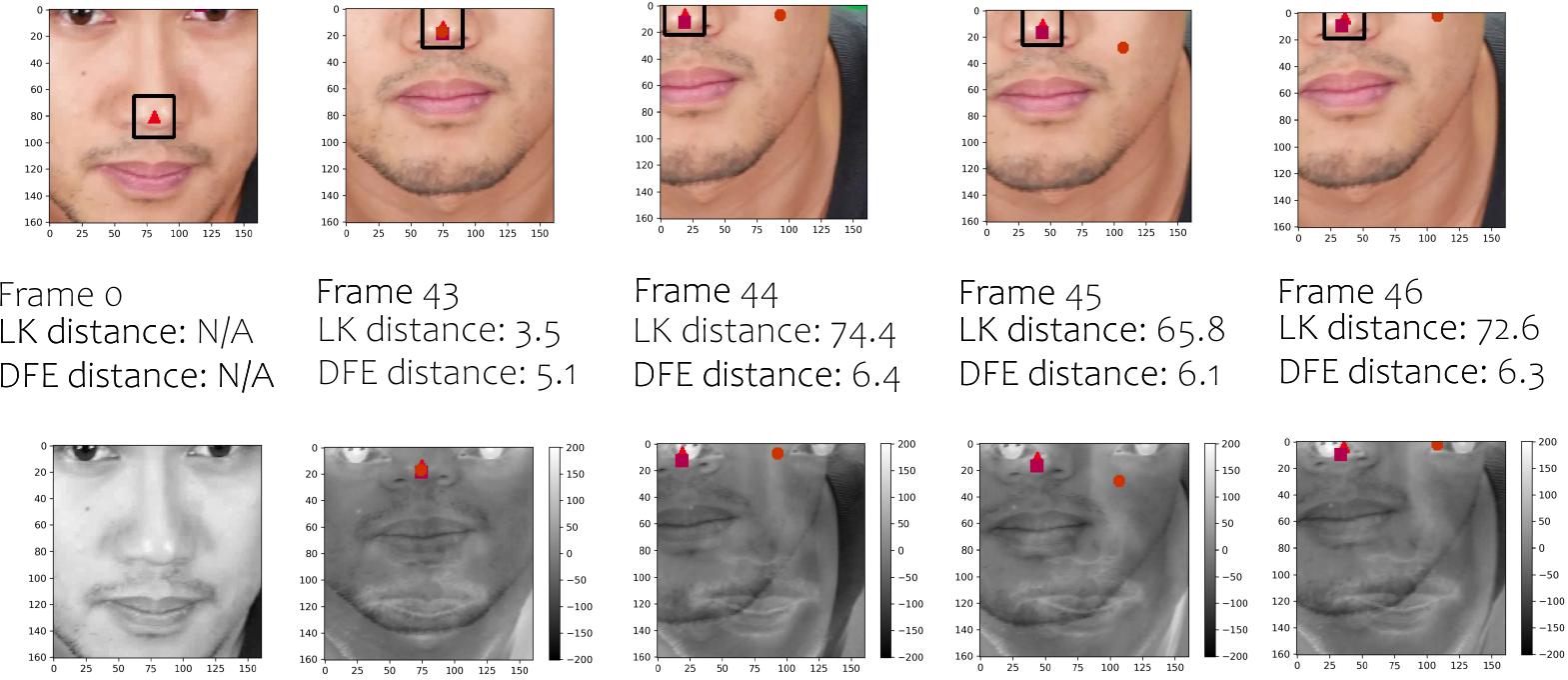}
\caption[Original frame and the first frames with large error for tracking the nose tip under bike conditions. Each frame is accompanied by a $160 \times 160$ difference map of corresponding image and the reference crop. The ground truth is marked by a red triangle, the LK prediction is marked by an orange circle and the DFE prediction is marked by a purple square.]{Original frame and the first frames with large error for tracking the nose tip under bike conditions. Each frame is accompanied by a $160 \times 160$ difference map of corresponding image and the reference crop. The ground truth is marked by a red triangle, the LK prediction is marked by an orange circle and the DFE prediction is marked by a purple square.}
\label{fig:divergent_error_frames}
\end{figure*}

We plotted frames 43 to 46 (in chronological order) along side the original frame in Figure~\ref{fig:divergent_error_frames}.
We also plotted the enlarged $160 \times 160$ window around the reference crop and the difference between the reference crop and other windows of the same dimensions in the frames 43 to 46 in grayscale.
These are similar to the observation window of LK with the parameters used in our experiments, $i.e.$ window size of $10 
\times 10$ and a $L_m$ of 4.

\begin{figure*}[tb!]
 \centering
 \includegraphics[width=0.7\linewidth]{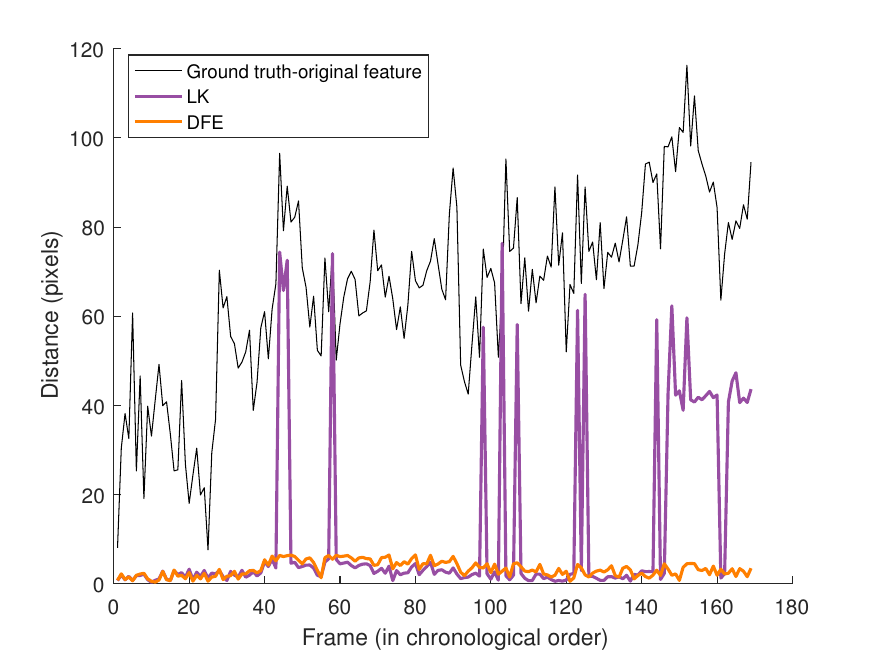}
\caption[Distance between the ground truth and the original reference image and the distances of LK and DFE methods for tracking the nose tip under bike conditions]{Distance between the ground truth and the original reference image and the distances of LK and DFE methods for tracking the nose tip under bike conditions. The correlation coefficient for the errors of LK and DFE are 0.52 and 0.33, respectively.}
\label{fig:correlation_large_movement_error_lk_dfe}
\end{figure*}

Figure~\ref{fig:correlation_large_movement_error_lk_dfe} shows the distance between the ground truth and the original reference image and the distances of LK and DFE methods for tracking the nose tip under bike condition temporally.
We believe that the large error from LK is due to the ground truth having a very large displacement from the original feature.
This displacement causes the new feature to be deformed and thus making the optical flow equations hard to solve.
As mentioned previously, an assumption from the LK method is that the motion is small \citep{lucas1981iterative}.
The error in LK has a correlation coefficient of 0.52 with the distance of the ground truth to the original reference feature.
Determining whether a correlation is strong is subjective but a correlation coefficient above 0.5 is often considered moderate to strong \citep{akoglu2018user}.
Thus we believe this to in part explain the large errors.

\subsection{Nearest neighbors to second nearest neighbor distance ratios}\label{app:nearestneighorsratios}

In SIFT a set of keypoints from one image is matched to a set of keypoints in another image by comparing the L2 norm of their residuals \citep{Lowe2004}.
Merely using tthe global minimum of the L2 norm produces a lot of false positive matches for the high-dimensional descriptors used in SIFT.
Lowe resorted to matching keypoints only if the ratio of the descriptor distances, $i.e.$ square root of the SSR, between the nearest neighbor and second nearest neighbor is lower than $0.8$. 
The threshold of $0.8$ is an experimental result after comparing the PDF of correct matches versus the PDF of incorrect matches for different closest to second closest nearest neighbor distances on a database of $40,000$ keypoints.
For their object recognition implementation, the threshold eliminated 90\% of false matches while discarding less than 5\% of the correct matches.
This strategy works well because matches need to have the closest neighbor significantly closer than the closest incorrect match to achieve a reliable matching, hence making the algorithm more robust.
For this reason we decided to also investigate the nearest and second nearest neighbor distances.

\begin{figure*}[tb!]
 \centering
  \includegraphics[width=\textwidth]{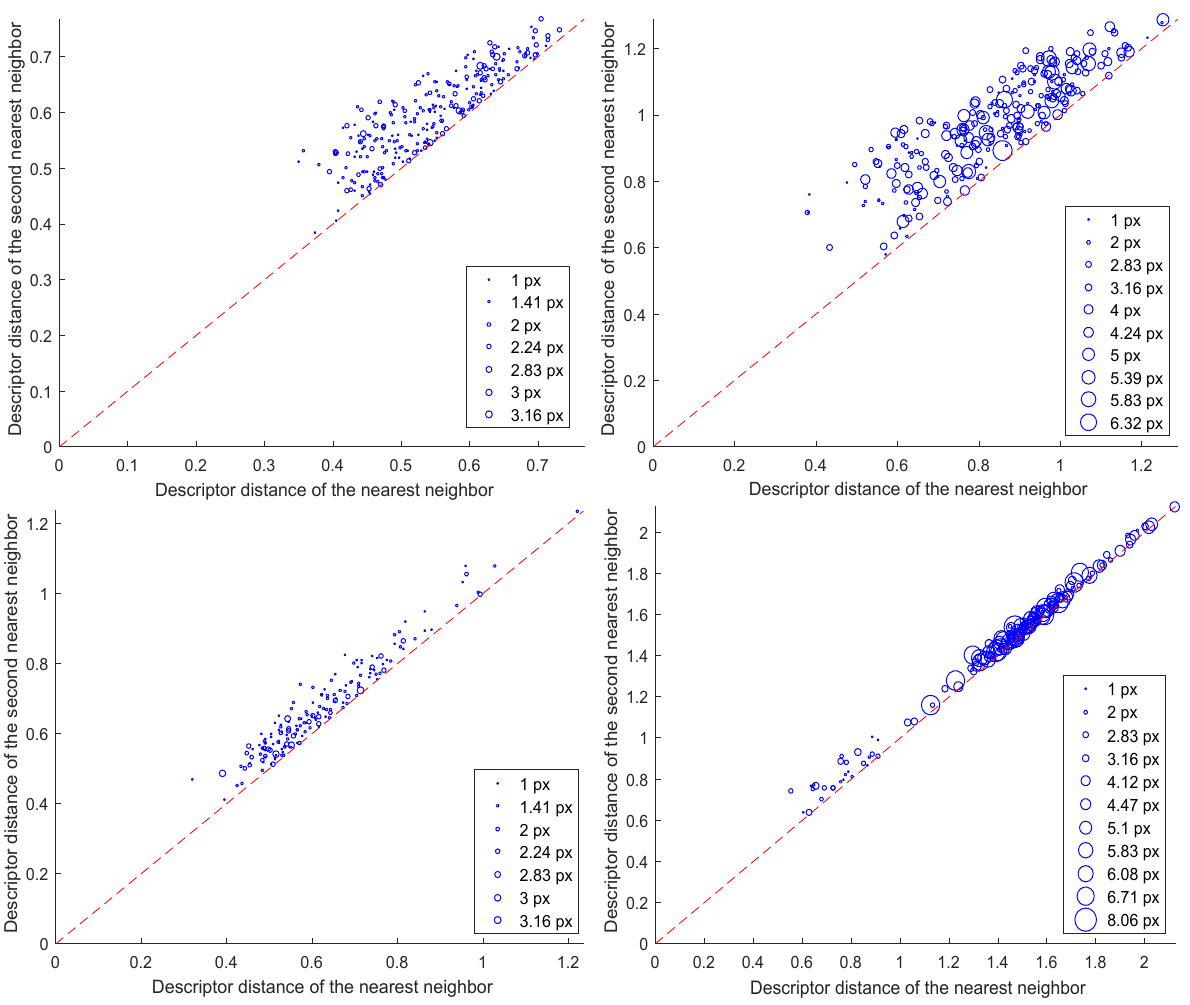}
\caption[Distance of the second nearest neighbor as a function of the distance of the nearest neighbor for the DFE method.]{Distance of the second nearest neighbor as a function of the distance of the nearest neighbor for matching the face mole under static conditions (top-left), nose tip under static conditions (top-right), face mole under bike conditions (bottom-left) \& nose tip under bike conditions (bottom-right). The descriptor distance of the nearest neighbors refers to the distance of the high dimensional representations between the reference point and the closest matching features in the image. The size of the markers is proportional to the largest spatial distance of the two neighbors to the ground truth. The red dashed line marks the $x=y$ line.}
\label{fig:nn_to_2nn}
\end{figure*}

Figure~\ref{fig:nn_to_2nn} show the spatial distance to the ground truth of the nearest and second nearest neighbors as a function of their descriptor distances in the 128-dimensional latent feature space with respect to the reference descriptor for our DFE method, $i.e.$ the square root of the SSR.
In 56\% of the images, the spatial distance between the ground truth and the second nearest neighbor is larger than that of the nearest neighbor.
In 90\% of the images, the largest spatial distance between any of the first two neighbors and the ground truth is smaller than 4.4 pixels.
The largest spatial distance among the first two neighbors, shown for tracking the nose tip in bike conditions, is just $8$ pixels.
The largest distance of the first two neighbors is larger for the nose tip than for the face mole, reflecting the fact that it is a less distinct feature.
For the static conditions, our method can achieve a lower nearest neighbor to second nearest neighbor ratio than for the bike conditions.
For the bike conditions, a large portion of the ratios is close to one.
This is not surprising, as for the bike conditions the deformation is larger and thus the point is harder to recognize across frames.

\begin{figure*}[tb!]
 \centering
  \includegraphics[width=\textwidth]{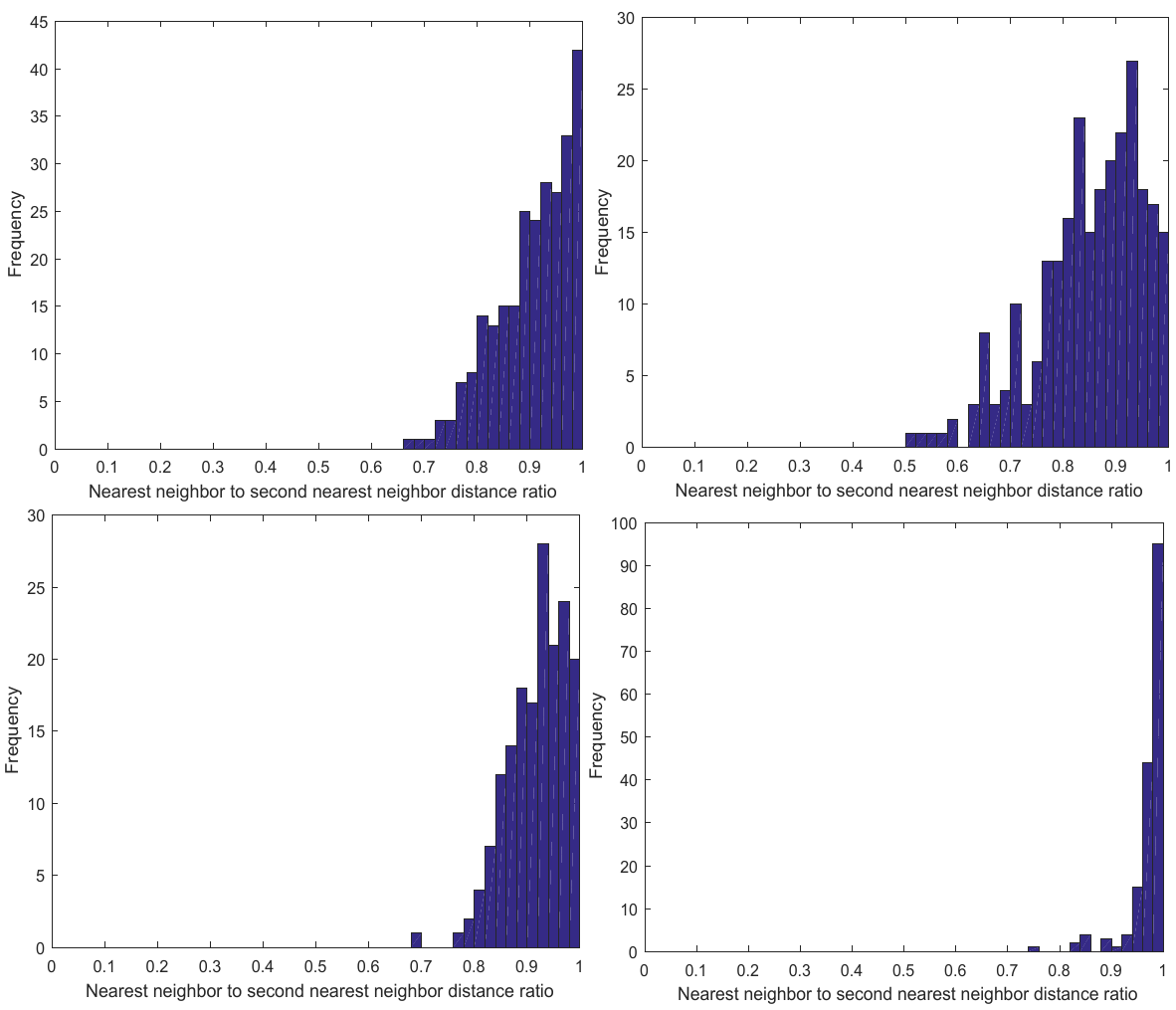}
\caption[Distribution of the ratios of the nearest to the second nearest neighbor distances for the Deep feature encodings method.]{Distribution of the ratios of the nearest to the second nearest neighbor distances for matching the face mole under static conditions (top-left), nose tip under static conditions (top-right), face mole under bike conditions (bottom-left) \& nose tip under bike conditions (bottom-right). The mean ratios are 0.9065, 0.8504, 0.9174, and 0.9718; respectively.}
\label{fig:nn_to_2nn_ratio_hists}
\end{figure*}

The distributions of the ratios of the nearest to second nearest neighbor distances for DFE are in Figure~\ref{fig:nn_to_2nn_ratio_hists}.
The smallest average mean ratio is for tracking the nose tip under static conditions and the largest mean ratio is for the nose tip under bike conditions.

Interpretation of the ratio of the nearest to second nearest neighbor distances is complicated by the fact that a large ratio can occur for different reasons.
In the case of a highly distinctive feature, characterised by a small spatial change leading to a large change in the feature descriptor, a large ratio would occur if the center of the feature occur midway between two pixels.
The SSR at both these pixels would then be significantly larger than the SSR of the center, but similar and seen as a ratio near one.
In the case of a second skin feature that locally is very similar to the one of interest but at a different location, the ratio would also be large.
Since we only evaluate the SSR with pixel level accuracy when calculating these ratios, we cannot distinguish this case from a case of having two similar skin features at different location in the image with similar SSR.
In the former case a high ratio is good, while in the latter case it is problematic since noise could cause the SSR of the wrong skin feature to be lower than the one of interest and thus a large tracking error. 
In the case of a not so distinctive feature, characterised by a small spatial change leading to a small change in the feature descriptor, the ratio of all points surrounding the center point of the feature will be large.
When tracking this feature, the large ratio increases the robustness because a small spatial error results from matching with one of the neighbouring pixels in some images. 
At the same time, the feature cannot be precisely located due to the low curvature of the local SSR landscape.
Taken together these cases highlight the fact that sometimes a high ratio and sometimes a low ratio is desired.
For DFE, we have observed local convexity of the SSR landscape around the skin features we have validated it on.
Thus a moderate nearest neighbor to second nearest neighbor ratio is desired because moving just one pixel from the best match should give a fair match and a similar SSR at the same time a distinction to the best match is desired.
This makes the prediction of the location of the reference feature, which at pixel-level is a result of a feature matching task work better using our subpixel level method.

\subsection{Nearest neighbors within an acceptable distance threshold}
\begin{figure*}[tb!]
 \centering
  \includegraphics[width=\textwidth]{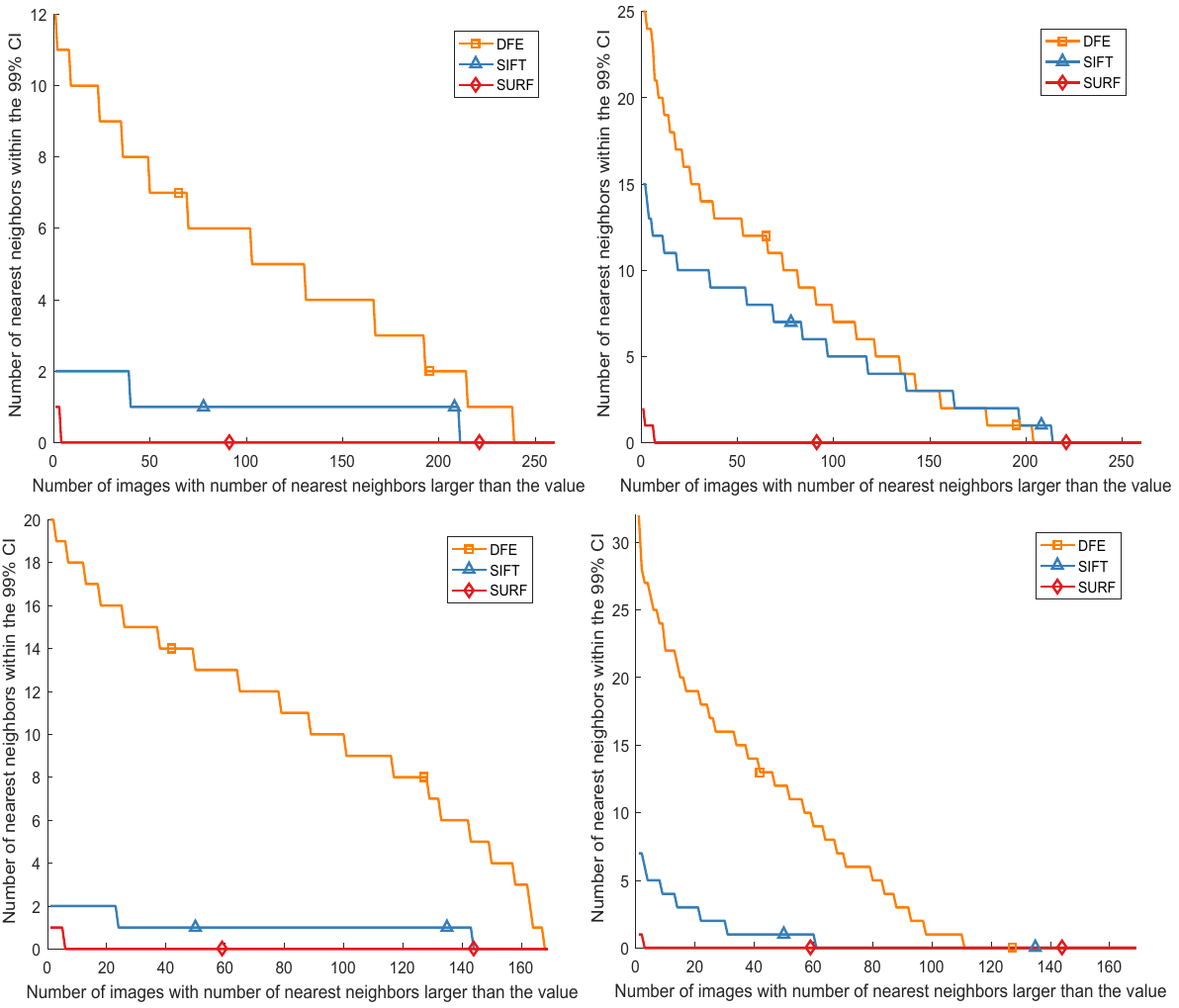}
\caption[Number of nearest neighbors within the 99\% CI]{Number of nearest neighbors within the 99\% CI for tracking the face mole under static conditions (top-left), nose tip under static conditions (top-right), face mole under bike conditions (bottom-left), and nose tip under bike conditions (bottom-right). The distances of the nearest neighbors are calculated at a pixel level resolution.}
\label{fig:sorted_acceptable_nn}
\end{figure*}

We also investigated the sensitivity of the methods to the selection of nearest neighbors within the distance threshold defined by the 99\% CI of the simulated Chi-square distribution.
We refer here to the spatial distance (in pixels) between the ground truth and the nearest neighbor. 
For each frame, we investigated the nearest neighbor rank, sorting by the distance of the neighbors to the ground truth point in ascending order, of the first point to fall outside of the threshold for SIFT, SURF, and DFE.
Figure~\ref{fig:sorted_acceptable_nn} shows the number of nearest neighbors that fall within this spatial distance threshold for all conditions.
Our method is superior to SIFT and SURF in all conditions, having more nearest neighbors close to the ground truth.
This implies that even when our method localizes a point incorrectly, it still typically selects a point in the neighborhood of the ground truth.
This implies that DFE is more precise since the neighbors are closer to each other.
From Figure 6, we know that DFE typically is more accurate, since the error to ground truth is the smallest.

\subsection{Spatial SSR landscape}\label{app:spatialSSRlandscape}
\begin{figure*}[tb!]
 \centering
  \includegraphics[width=\textwidth]{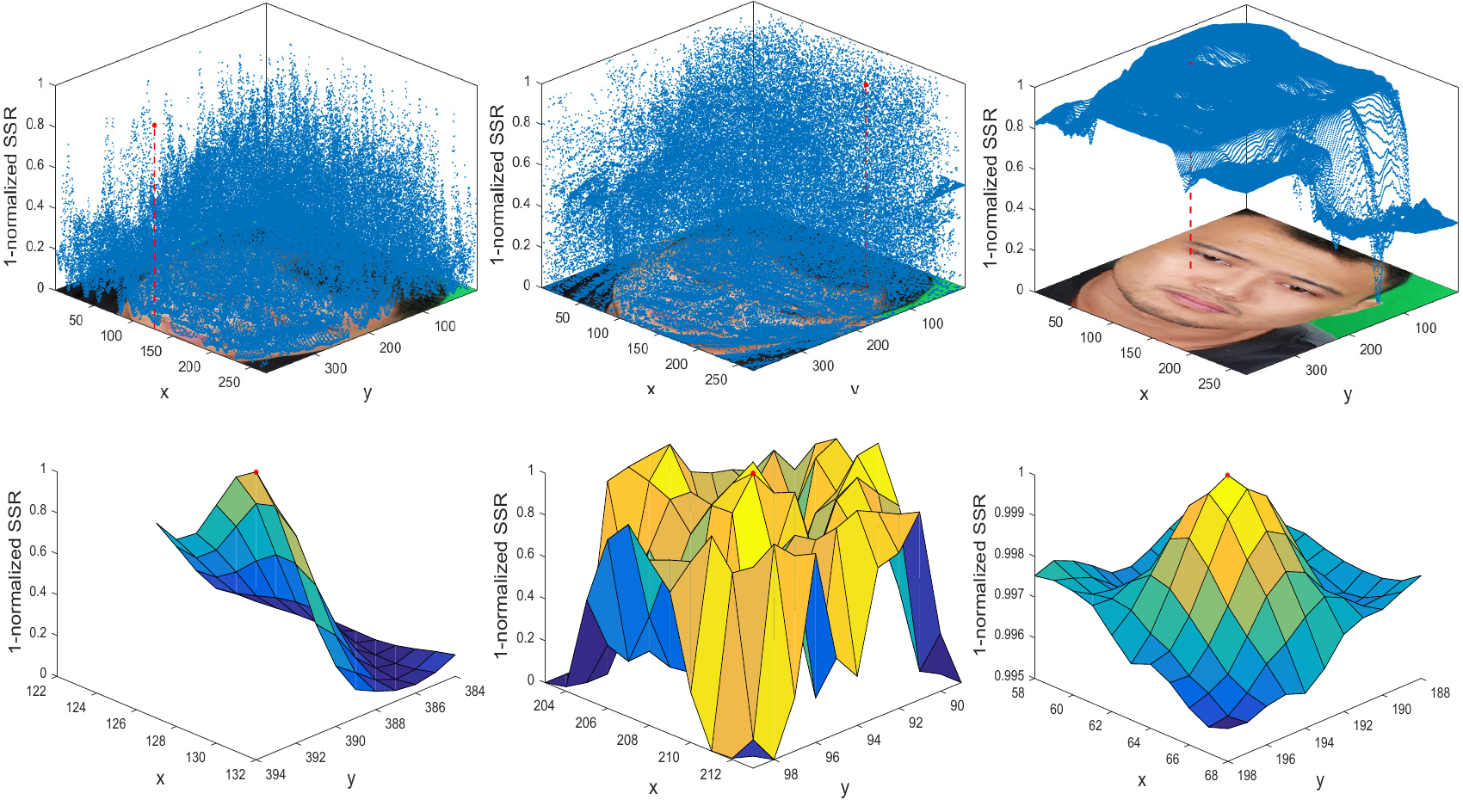}
\caption[SSR as a function of the spatial image coordinates.]{SSR of the frames with the $15$th largest SSR value for tracking the face mole under bike conditions as a function of the spatial image coordinates for SIFT (left), SURF (middle), and DFE (right) and their corresponding regional enlargements of the optimum. The red dashed lines mark the point with minimum SSR. The distances to the ground truth are 95.65, 212.41, and 1.73 pixels, respectively. Both SIFT and SURF perform poorly while DFE performs well.}
\label{fig:example_spatial_ssr}
\end{figure*}

\begin{figure*}[tb!]
 \centering
  \includegraphics[width=\textwidth]{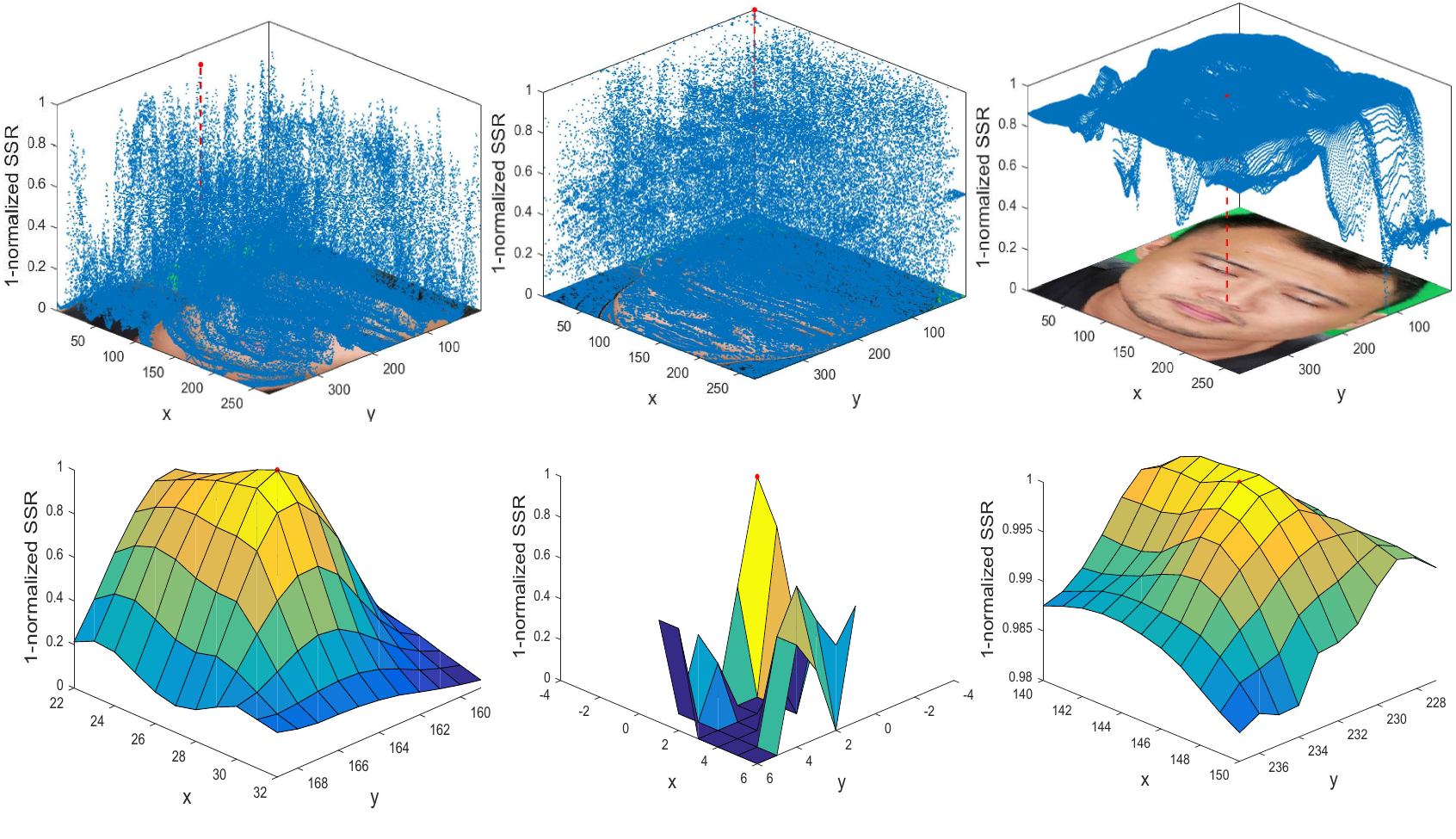}
\caption[SSR as a function of the spatial image coordinates.]{SSR as a function of the spatial image coordinates for the images with the largest localization errors for nose tips under bike conditions for SIFT (left), SURF (middle), and DFE (right) and the regional enlargements of the optimum. The red dashed lines mark the point in the space with minimum SSR, which is plotted as a red circle. The distances to the ground truth are 179.65, 355.45, and 6.48 pixels, respectively.}
\label{fig:spatial_ssr_worst}
\end{figure*}

\begin{figure*}[tb!]
 \centering
  \includegraphics[width=\textwidth]{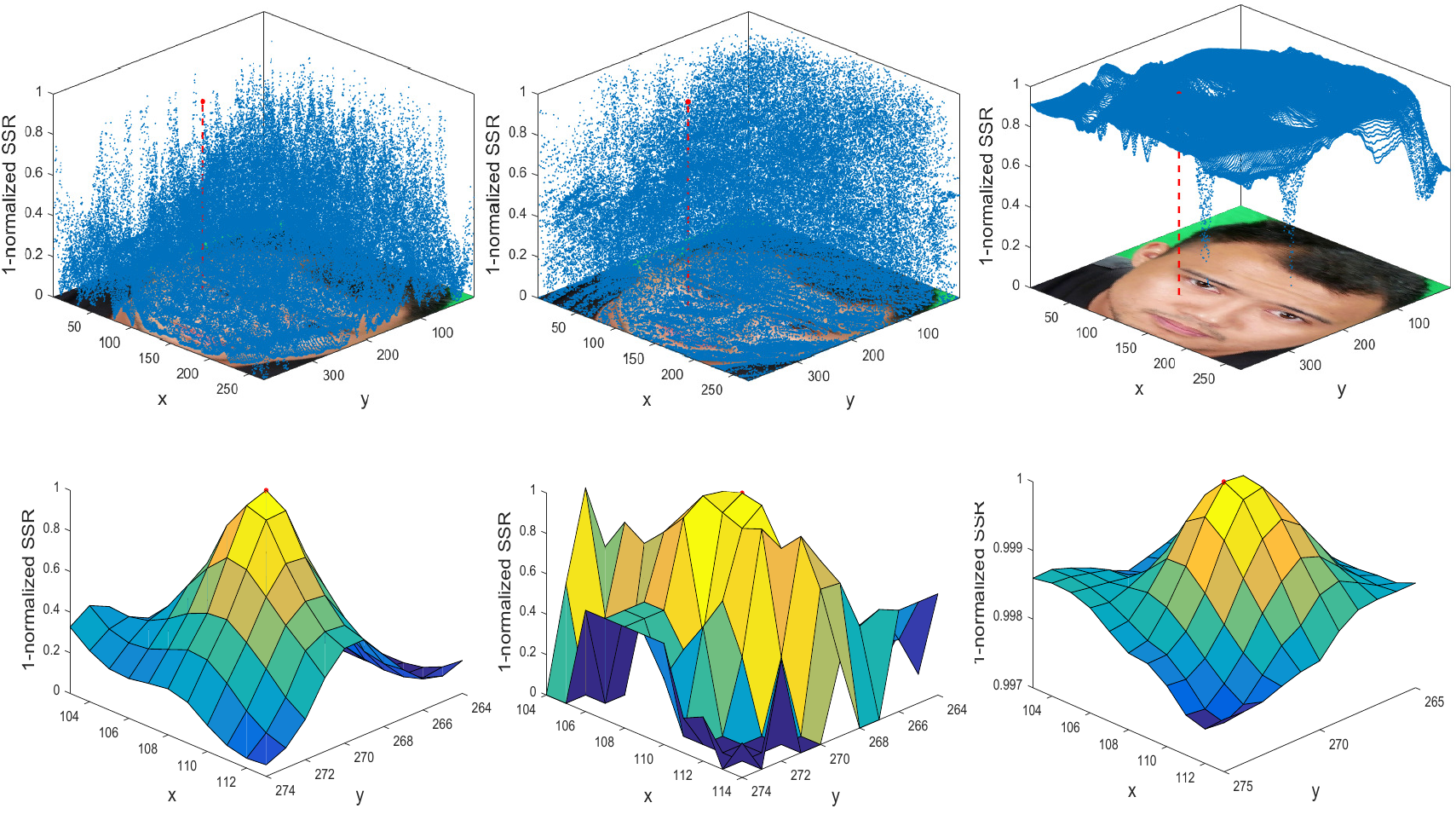}
\caption[SSR as a function of the spatial image coordinates.]{SSR as a function of the spatial image coordinates for the single image with the smallest sum of localization errors across SIFT (left), SURF (middle), and DFE (right) for tracking the face mole under bike conditions and their corresponding regional enlargements. The red dashed lines mark the point with minimum SSR. The distances to the ground truth are 0.89, 1.00, and 0.89 pixels, respectively.}
\label{fig:spatial_ssr_single}
\end{figure*}

Similarly to Figures 12,~\ref{fig:spatial_ssr_worst}, and ~\ref{fig:spatial_ssr_single}; we in Figure~\ref{fig:example_spatial_ssr} observe that the SSR landscape of our method contains an organized structure, contrary to SIFT and SURF, which appear to have a random structure.
We plotted the SSR as a function of spatial conditions of the image for the 15th largest error for tracking the face mole under bike conditions. 
We chose this image because it is a case in which DFE performs well, achieving the lowest error of 1.73 pixels, while SIFT and SURF perform poorly.
It is hence an example where accurate feature matching is possible.

\begin{figure*}[tb!]
 \centering
  \includegraphics[width=\textwidth]{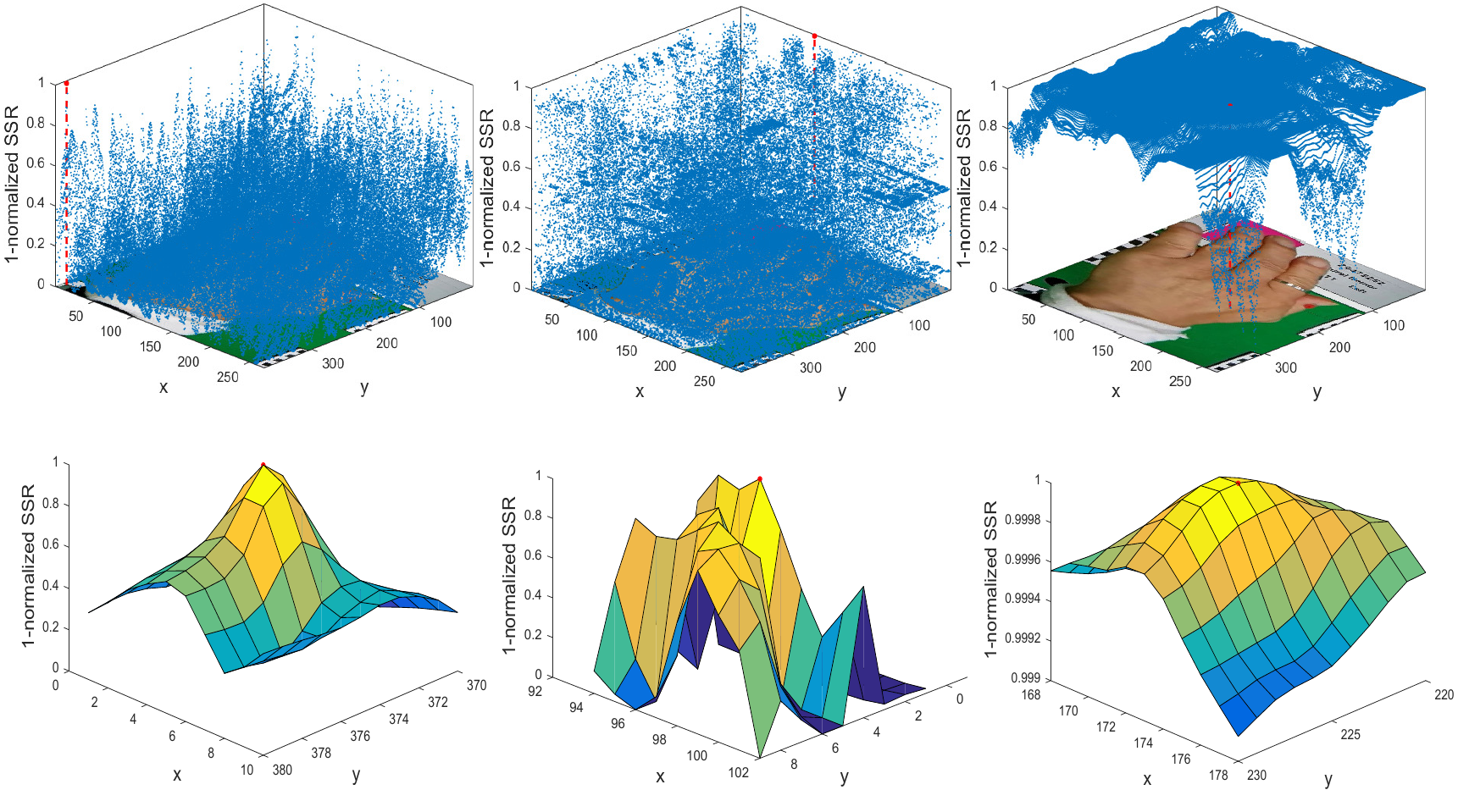}
\caption[SSR as a function of the spatial image coordinates.]{SSR as a function of the spatial image coordinates for the images with the largest localization errors for tracking the moles in the hand of the PD patient for SIFT (left), SURF (middle), and DFE (right) and their corresponding regional enlargement of the optimum. The red dashed lines mark the point with minimum SSR. The distances to the ground truth are 219.85, 242.34, and 1.52 pixels, respectively.}
\label{fig:spatial_ssr_worst_pd}
\end{figure*}

\begin{figure*}[tb!]
 \centering
  \includegraphics[width=\textwidth]{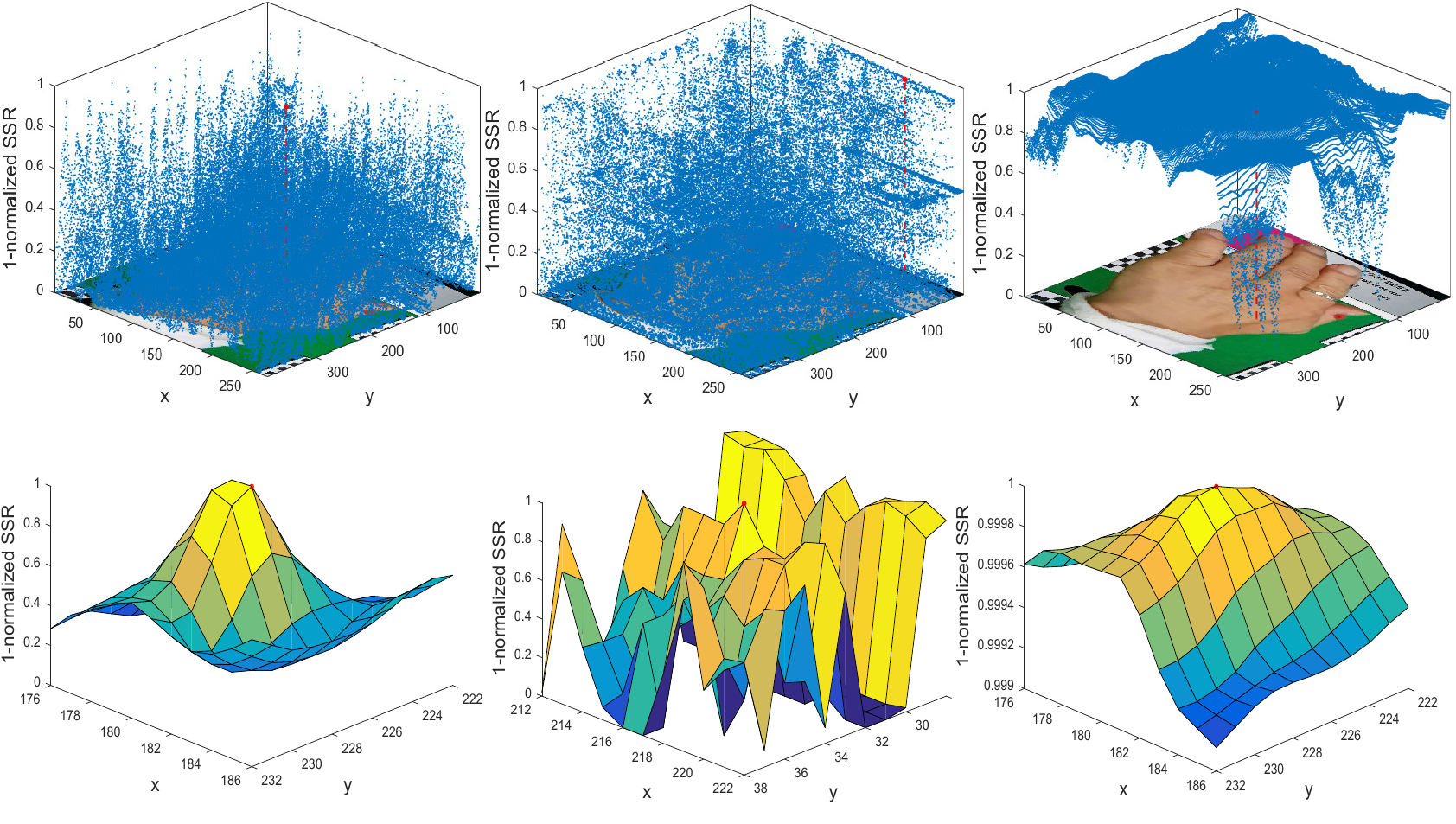}
\caption[SSR as a function of the spatial image coordinates.]{SSR as a function of the spatial image coordinates for the single image with the smallest sum of localization errors for tracking the mole on the hand of the PD patient for SIFT (left), SURF (middle), and DFE (right) and their corresponding regional enlargements of the optimum. The red dashed lines mark the point with minimum SSR. The distances to the ground truth are 1.55, 196.12, and 1.41 pixels, respectively.}
\label{fig:spatial_ssr_single_pd}
\end{figure*}

In addition, we also calculated the SSR landscape of the largest errors and smallest sum of localization errors in Figures~\ref{fig:spatial_ssr_worst} and ~\ref{fig:spatial_ssr_single}, respectively.
For the worst cases the localization error is the worst globally among all body part-motion combinations. 
The cumulative error for the image with smallest sum of localization errors is $2.786$ pixels for tracking the face mole under bike conditions.
Similarly to Figure 12 and Figure~\ref{fig:spatial_ssr_worst}, only the DFE show an organized structure.
SIFT and SURF have landscapes that seem random and irregular.
The generalisations to the PD data are seen in Figures~\ref{fig:spatial_ssr_worst_pd} and ~\ref{fig:spatial_ssr_single_pd} for the largest errors and smallest sum of localization errors, respectively.


\subsection{Algorithm errors}
\begin{figure*}[tb!]
 \centering
  \includegraphics[width=\textwidth]{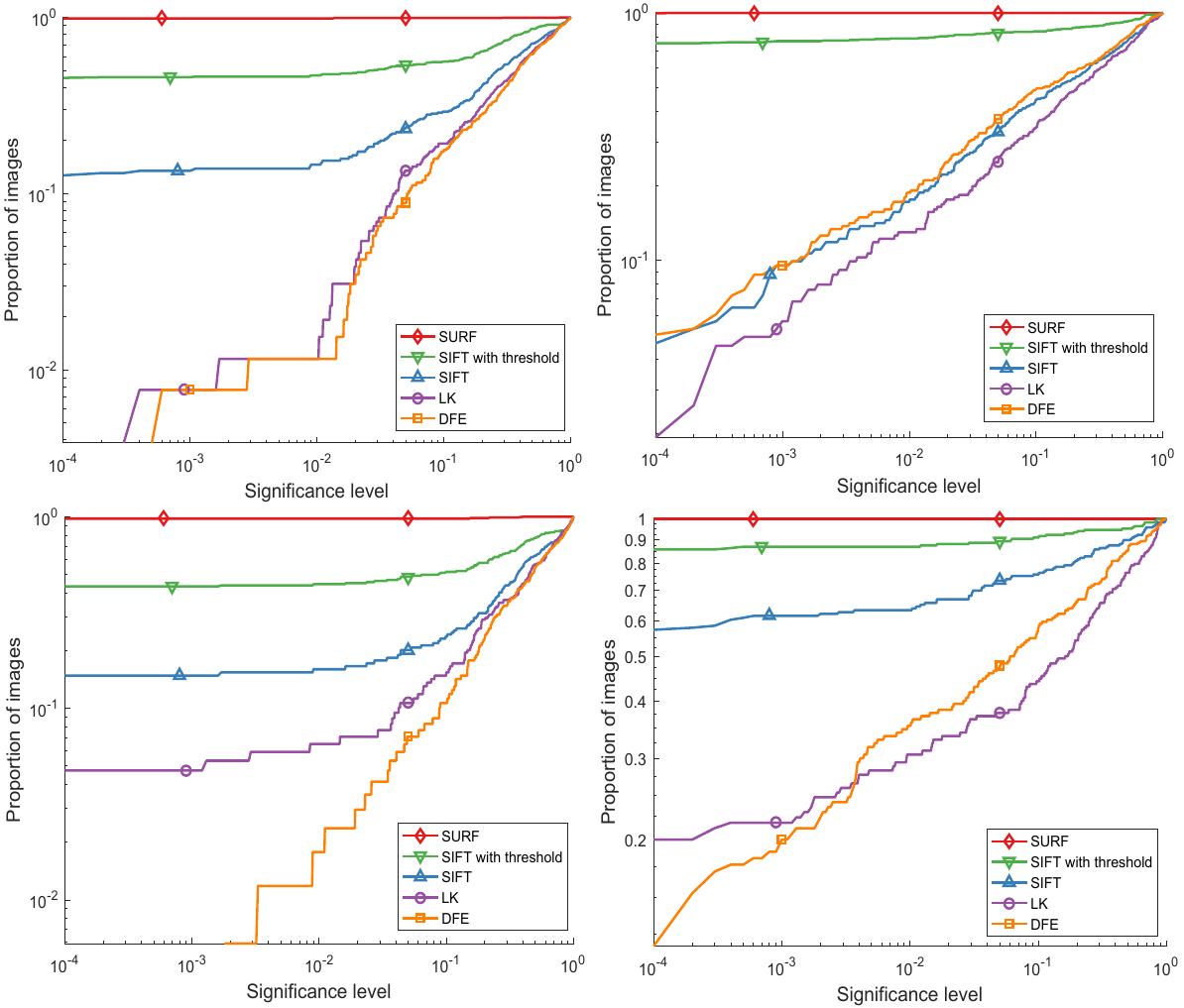}
\caption[Proportion of errors above the threshold as a function of significance level.]{Proportion of images for which the hypothesis that the error is due to human labelling of the ground truth can be rejected at the significance level for tracking the face mole under static conditions (top-left), nose tip under static conditions (top-right), face mole under bike conditions (bottom-left), and nose tip under bike conditions (bottom-right).}
\label{fig:proportion_errors_significance_level}
\end{figure*}

The proportion of images for which the hypothesis that the error is due to human labelling of the ground truth can be rejected as a function of significance level reveals when the error is likely due to the algorithm.
Figure~\ref{fig:proportion_errors_significance_level} shows the proportion of images in each of the four validation cases below each possible significance threshold established by our Chi square analysis.
The error in a large proportion of the images is unlikely to be due to human labelling of the ground truth, especially for SIFT and SURF, while it is likely to be due to human labelling for DFE.
A low line at all points in these figures is a hallmark of the good performance of a method.

\clearpage
\bibliography{../../LibraryAllReferences} 


\end{document}